\documentclass{article} 
\usepackage{iclr2021_conference,times}

\usepackage{amsmath,amsfonts,bm}

\def\eqref#1{equation~\ref{#1}}

\def\1{\bm{1}}

\DeclareMathAlphabet{\mathsfit}{\encodingdefault}{\sfdefault}{m}{sl}
\SetMathAlphabet{\mathsfit}{bold}{\encodingdefault}{\sfdefault}{bx}{n}

\usepackage{hyperref}
\usepackage{url}

\usepackage{graphicx}
\usepackage{bm}
\usepackage{amsmath}
\usepackage{amsfonts}
\usepackage{bbm}

\usepackage{helvet}
\usepackage{courier}
\usepackage{multirow}
\usepackage{bm}
\usepackage{amsmath}
\usepackage{mdwlist}

\usepackage{relsize}
\usepackage{amsfonts}
\usepackage{url}
\usepackage{graphicx}
\usepackage{subfigure}
\usepackage{caption}
\usepackage{hhline}
\usepackage{color}
\usepackage{colortbl}
\usepackage{booktabs}
\usepackage{tabularx}
\usepackage{amsmath}
\usepackage{makecell}
\usepackage[linesnumbered,boxed,ruled,commentsnumbered]{algorithm2e}
\usepackage{lipsum}
\usepackage{booktabs}
\usepackage{pifont}
\title{Unveiling A Core Linguistic Region in Large Language Models}

\author{Jun Zhao$^1$$^\dag$,  Zhihao Zhang$^1$$^\dag$,  Yide Ma$^2$,  Qi Zhang$^1$\thanks{Corresponding authors, $^\dag$Equal Contributions},  Tao Gui$^1$,   Luhui Gao$^3$,  Xuanjing Huang$^1$\\
$^1$ School of Computer Science, Fudan University\\
$^2$  Faculty of Arts \& Science,  University of Toronto\\
$^3$  College of Foreign Languages and Literature,  Fudan University\\
\texttt{\{zhaoj19, zhangzhihao19, qz, tgui, xjhuang\}@fudan.edu.cn} \\
}

\iclrfinalcopy 
\begin{document}

\maketitle

\begin{abstract}
Brain localization, which describes the association between specific regions of the brain and their corresponding functions, is widely accepted in the field of cognitive science as an objective fact. Today's large language models (LLMs) possess human-level linguistic competence and can execute complex tasks requiring abstract knowledge and reasoning. To deeply understand the inherent mechanisms of intelligence emergence in LLMs, this paper conducts an analogical research using brain localization as a prototype. We have discovered a core region in LLMs that corresponds to linguistic competence, accounting for approximately 1\% of the total model parameters. This core region exhibits significant dimension dependency, and perturbations to even a single parameter on specific dimensions can lead to a loss of linguistic competence. Furthermore, we observe that an improvement in linguistic competence does not necessarily accompany an elevation in the model's knowledge level, which might imply the existence of regions of domain knowledge that are dissociated from the linguistic region. Overall, exploring the LLMs' functional regions provides insights into the foundation of their intelligence. In the future, we will continue to investigate knowledge regions within LLMs and the interactions between them.
\end{abstract}

\section{Introduction}

Over the years, the field of Natural Language Processing (NLP) has been at the forefront of understanding the core principles of intelligence \citep{bubeck2023sparks}. The emergence of large language models (LLMs) such as ChatGPT \citep{chatgpt}, PaLM \citep{anil2023palm}, LLaMA \citep{touvron2023llama}, and their peers, showcases a significant breakthrough. Thanks to unparalleled scales of model architecture and the vastness of training data, these LLMs now exhibit exceptional linguistic competence and can execute complex tasks requiring abstract knowledge \citep{dong2023survey} and reasoning \citep{DBLP:journals/corr/abs-2110-14168}. However, the academic community lacks a systematic understanding of the internal mechanisms of LLMs' intelligence, and there is debate over whether LLMs can truly be considered "thinking machines." \citep{sentient, mahowald2023dissociating}. Nevertheless, insights from cognitive science may offer fresh perspectives on this matter.

\begin{figure}[t]
        \includegraphics[width=\columnwidth]{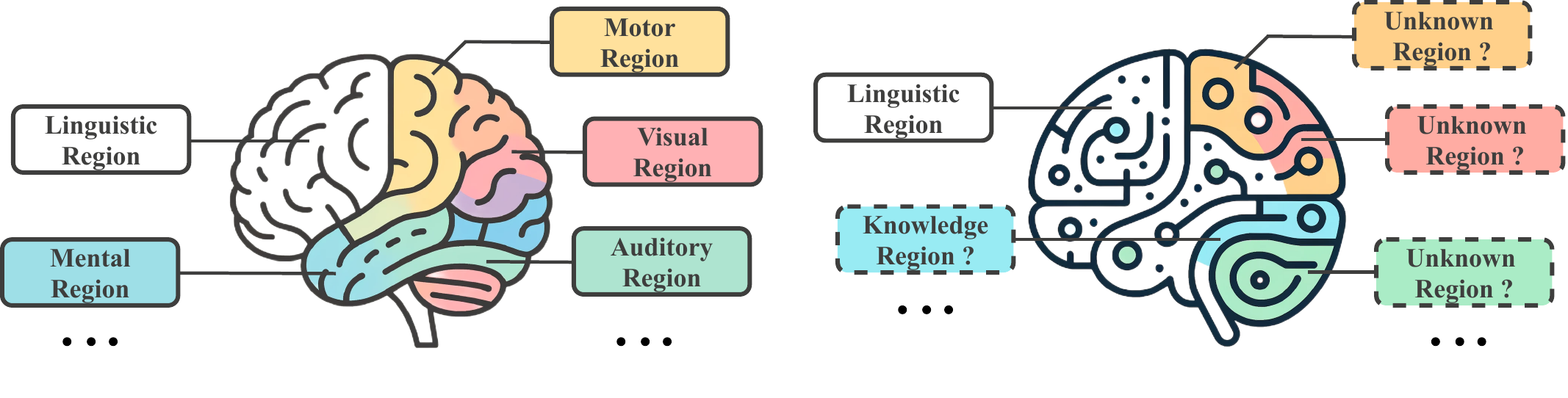}
        \caption{Based on the human brain (left) as a prototype, we have discovered a region in LLMs (right) that corresponds to linguistic competence. Furthermore, we have found that improvements in linguistic competence do not necessarily coincide with increases in knowledge levels, which may suggest the presence of a dissociated knowledge region. In the future, we will continue to explore the possibility of other functional regions.}
        \label{fig:intro}
    \end{figure}

Cognitive science is an interdisciplinary field that investigates the mechanisms of human thought and perception. Numerous literatures indicate that different regions of the brain are associated with specific functions \citep{article}. Figure \ref{fig:intro} (left) is a schematic diagram of the brain localization. For example, language processing in humans involves a brain regions in the frontal and temporal lobes, predominantly in the left hemisphere. This region underpins both the comprehension \citep{Deniz7722,Scott_Gallée_Fedorenko_2017,Regev_Honey_Simony_Hasson_2013,Fedorenko_Hsieh_Nieto-Castañón_Whitfield-Gabrieli_Kanwisher_2010} and production \citep{Menenti_Gierhan_Segaert_Hagoort_2011,Hu_Small_Kean_Takahashi_Zekelman_Kleinman_Ryan_Nieto-Castañón_Ferreira_Fedorenko_2021} of language across spoken, written, and signed modalities. Adjacent to this linguistic network is the domain of logical inference, which taps into different regions of the frontal and parietal. These regions stand apart from the language-centric pathways. Collectively, they form the 'multiple demand network.' \citep{Duncan_Assem_Shashidhara_2020}. This network is pivotal in supporting a myriad of cognitively demanding tasks, from logical deductions and mathematical analyses \citep{Fedorenko_Duncan_Kanwisher_2013, Amalric_Dehaene_2019} to physical reasoning \citep{Schwettmann_Tenenbaum_Kanwisher_2019,Pramod_Cohen_Tenenbaum_Kanwisher_2021} and computer code understanding \citep{Ivanova_Srikant_Sueoka_Kean_Dhamala_O’Reilly_Bers_Fedorenko_2020,10.7554/eLife.59340}. On a related note, individuals diagnosed with semantic dementia, which primarily affects the anterior temporal lobes, often grapple with tasks centered on world knowledge. Their struggle remains consistent whether the information is presented through words or visual cues like images \citep{Patterson_Nestor_Rogers_2007}. This phenomenon serves as a testament to the idea that while language and general world knowledge are closely intertwined in practical usage, they are underpinned by distinct neural circuits.

The regions within the human brain collaboratively form the foundation of human intelligence. We wonders if LLMs as large-scale artificial neural networks manifest similar functional regions phenomenon internally, akin to human brain. This paper embarks on a preliminary exploration, delving deeper into the intrinsic mechanisms of LLMs' intelligence. Through analysis and comparison of six languages, we discover a core region in LLMs corresponding to linguistic competence, which accounts for approximately 1\% of the model's total parameters. Perturbations to this region consistently lead to a sharp decline in performance across 30 test languages. We observe that the linguistic core region of LLMs exhibits significant dimension dependence. In certain dimensions, perturbing a single parameter could lead to the model losing its linguistic competence. Additionally, further pretraining on LLaMA model with over 100 billion tokens do not yield performance improvements on C-Eval \citep{huang2023ceval}, a Chinese exam benchmark. This indicates that the enhancement of the model's linguistic competence does not necessarily coincide with an increase in knowledge level. Thus, a plausible hypothesis is that there might be knowledge regions in the model beyond the linguistic region, perhaps even unknown regions modeling higher-level reasoning.

Exploring the functional regions of LLMs holds immense scientific value and practical significance. Firstly, it aids in a comprehensive understanding of the intrinsic mechanisms of LLMs' intelligence. Moreover, comprehending the interplay between regions can guide LLM pre-training. How should we design the optimal model structure? What's the best data mixing? How can we prevent instability during training? In the future, we will delve deeper into the functional localization within LLMs.
\begin{figure*}[t]
        \includegraphics[width=\linewidth]{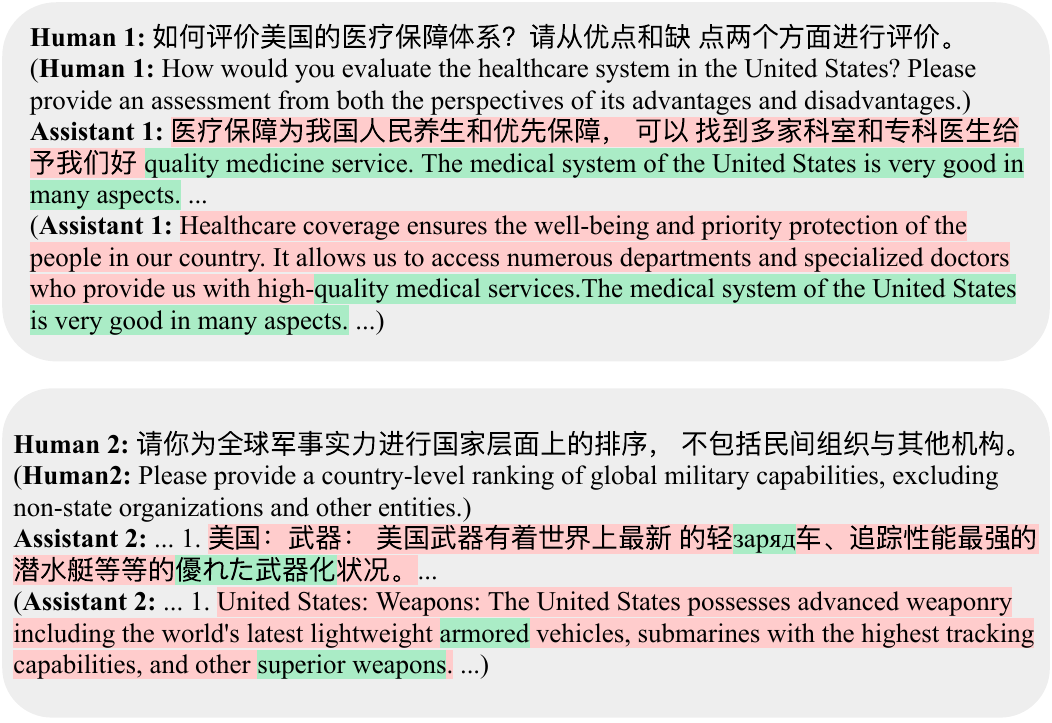}
        \caption{Case study of code-switching. Text with a red background represents the non-English query language (Chinese). Text with a green background indicates code-switching language in the model's output, which could be English, Japanese, Russian or other languages.}
        \label{fig:case2}
    \end{figure*}  
\section{Preliminaries and Background}
\subsection{Pretraining to acquire Linguistic competence and knowledge}
\label{sec:pretrain}
Linguistic competence is a set of core, specific capacities required to produce and comprehend a given language., while knowledge pertains to the understanding and recognition of things, concepts, or principles. Pre-training is a crucial process by which LLMs acquire linguistic competence and gain general knowledge about the real world. Specifically, a corpus is firstly constructed from the internet, encompassing a vast array of textual data including web pages, articles, books, and the like. After undergoing data cleaning and filtering, these corpora are further segmented into token sequences. Each token can be a word or a subword, enabling the model to better apprehend semantic relations between words and handle unknown and rare tokens. Based on the corpus, pretraining aims to predict the next token based on the prefix sequences. Formally, given a large corpus $\mathcal{D}$, the training objective is to minimize the following loss:
\begin{equation}
\mathcal{L}_{pretrain}=\sum_{x\in\mathcal{D}}\sum_i\log p_\theta(x_i|x_1,...,x_{i-1}),
\end{equation}
where $x=\{x_1, ..., x_n\}$ denotes an input token sequence.

By pretraining on massive text data ranging from billions to trillions of tokens, LLMs are capable of capturing intricate language structures, semantics, and contextual relationships. These models have not only achieved success on general language understanding benchmarks developed by NLP researchers, such as the GLUE \citep{NEURIPS2019_4496bf24} tasks, but they have also made breakthrough advancements in linguistic competence tests. For instance, the benchmark test BLiMP \citep{warstadt2020blimp} incorporates minimal contrasts between grammatical and ungrammatical sentences, probing a variety of challenging linguistic phenomena, such as filler-gap dependencies (The book which Mary bought \_\_\_ is on the table. vs *The book which bought \_\_\_ is on the table.) and negative polarity licensing (John has never been to Paris. vs. *John has ever been to Paris.)



\subsection{Supervised fine-tuning for aligning with human intent}
Supervised fine-tuning (SFT) aims to further enhance the capability of LLMs to follow instructions. Its training data consists of many instruction-response pairs. The model needs to learn to accurately respond to instructions, rather than merely continuing from the preceding text. Formally, given an instruction dataset $\mathcal{D}^\prime=\{(I, Y)\}$, where $I$ represents a task instruction and $Y$ represents a desired response, the training objective of instruction tuning is to minimize the following loss:
\begin{equation}
\mathcal{L}_{ins}=-\log p_\theta(Y|I),
\end{equation}
By tuning on diverse instruction tasks, the model is able to better comprehend and follow human instructions, and generalize to unseen instructions.

We find that when fine-tuning with a small amount of instruction pairs (between 0 to 5,000) on languages that LLaMA is not familiar with (such as Chinese), the responses exhibit code-switching behavior. As shown in Figure \ref{fig:case2}, LLaMA-7B switches between multiple languages in responding to instructions, yet the semantic flow and correctness are maintained. We speculate that LLMs might contain a core linguistic competence region, which models the general linguistic patterns and cross-linguistic semantic alignment relationships.

        \begin{table}
            \centering
            \begin{tabular}{ccc c}
            \toprule
            &$\theta=1\%$ & $\theta=3\%$ & $\theta=5\%$\\
            \midrule
            Variation $<\theta$ & 0.008\% & 0.981\% &  5.327\%\\
            Variation $>\theta$ & 54.669\% & 25.742\% & 16.382\% \\
            \bottomrule
            \end{tabular}
            \caption{Parameter proportion with $<\theta$ (or $>\theta$) variation across six languages. In language fine-tuning, approximately 0.008\% to 5.327\% of the parameters tend to remain unchanged, while around 16.382\% to 54.669\% of the parameters are prone to change.}
            \label{tab:var_proportion}
        \end{table}

        \begin{table}
            \centering
            \begin{tabular}{ccc ccc}
            \toprule
            \multirow{2}{*}{\makecell[c]{Model}} & \multirow{2}{*}{\makecell[c]{\# Training \\Samples}} & \multirow{2}{*}{\makecell[c]{Perturbation\\Ratio}}&\multicolumn{3}{c}{Perturbation Region}\\
            \cline{4-6}
            & & &Top & Bottom & Random\\
            \midrule
            \multirow{3}{*}{\makecell[c]{LLaMA2-7B}}
            &100K& $1\%$ &  \cellcolor{gray!20}6.833&   71137.844&   \cellcolor{gray!20}6.764\\
            &100K& $3\%$ & \cellcolor{gray!20}10.686&  272805.125& \cellcolor{gray!20} 8.536\\
            &100K& $5\%$ & \cellcolor{gray!20}28.073&  218519.219& \cellcolor{gray!20} 12.539\\
            \hline \hline
            \multirow{3}{*}{\makecell[c]{LLaMA2-13B}}&100K& $1\%$& \cellcolor{gray!20}6.013&   62191.785&  \cellcolor{gray!20} 6.01\\
            &100K& $3\%$ &\cellcolor{gray!20}6.692&   116946.891& \cellcolor{gray!20} 6.642\\
            &100K& $5\%$ & \cellcolor{gray!20}7.718&   74648.281&  \cellcolor{gray!20} 8.014\\
            \hline \hline
            \multirow{3}{*}{\makecell[c]{LLaMA2-13B}}
            &10K& $1\%$ &  \cellcolor{gray!20}6.31&  31714.055&  \cellcolor{gray!20} 6.03\\
            &10K& $3\%$ & \cellcolor{gray!20}8.191&   158100.438& \cellcolor{gray!20} 6.71\\
            &10K& $5\%$ & \cellcolor{gray!20}11.633&  214658.359& \cellcolor{gray!20} 8.123\\
            \bottomrule
            \end{tabular}
            \caption{LLaMA perplexity on the Chinese Wechat dataset when perturbing different regions and proportions of parameters.
            `Top' and `Bottom' respectively represent the $N$ parameters with the largest and smallest changes during the fine-tuning process on the six languages. `Random' refers to the selection of $N$ parameters chosen at random for comparison. $N$ is the product of the total number of parameters and the perturbation ratio.}
            \label{tab:chn_scatter}
        \end{table}

\section{The Core Linguistic Competence Region}
\subsection{Experimental Setup}
To localize the functional regions corresponding to linguistic competence within LLMs and analyze their nature, we perform language fine-tuning (next token prediction) on various languages and observe the relationship between internal parameter shifts and external output quality. We utilize LLaMA2 7B/13B as our model instance, as it stands out as one of the most notable state-of-the-art open-source LLMs in current academia. Our experimental dataset comprises materials from Chinese platforms like Zhihu and Wechat, English sources from Arxiv and Falcon, and a corpus including books from 28 languages, totaling 30 languages in all. Six languages, namely Arabic, Spanish, Russian, Chinese, Korean, and Vietnamese, are chosen for language fine-tuning and region localization, with $100,000$ samples for each (distinct from the samples in the test set). All 30 languages are employed for model testing and functional region analysis, with the specific languages and token count detailed in \ref{sec:app_lang}. We use perplexity (PPL) as the criterion for evaluating the linguistic competence of a language model.

\subsection{Localization of the linguistic competence region}
\label{sec:localization}
In this section, we conduct fine-tuning experiments on LLaMA across six languages, aiming to explore and identify core parameter regions associated with linguistic competence. Specifically, we posit that the set of parameters exhibiting minimal variations during the language fine-tuning may have a strong correlation with the model's linguistic competence, and we provide both logical and empirical evidence to support this hypothesis.
\begin{table*}[t]
            \centering
            \resizebox{\linewidth}{!}{
            \begin{tabular}{l cccc cccc}
            \toprule
            \multirow{2}{*}{\textbf{Languages}} & \multicolumn{4}{c}{LLaMA2-7B} & \multicolumn{4}{c}{LLaMA2-13B}\\
            \cmidrule(r){2-5}\cmidrule(r){6-9}
            & Base & Top & Bottom & Random & Base & Top & Bottom & Random \\
            \midrule
            
            Arabic & \cellcolor{gray!20}6.732  & 10.89 &  \cellcolor{gray!20}132988.312 & 8.815 &  \cellcolor{gray!20}6.265  & 8.296 & \cellcolor{gray!20} 66492.734  & 7.836\\
            Chinese & \cellcolor{gray!20}8.554 &  15.018 & \cellcolor{gray!20}200279.453 & 10.909  &\cellcolor{gray!20}7.832  & 8.951 &  \cellcolor{gray!20}136295.359 & 8.757\\
            Czech &  \cellcolor{gray!20}19.622 & 37.882 & \cellcolor{gray!20}48612.707  & 28.025  &\cellcolor{gray!20}17.367 & 23.863 &\cellcolor{gray!20} 20363.225  & 22.303\\
            Danish  & \cellcolor{gray!20}8.412  & 16.151  &\cellcolor{gray!20}72907.688  & 11.224  &\cellcolor{gray!20}7.414  & 8.507  &\cellcolor{gray!20} 18157.621  & 8.627\\
            Dutch  & \cellcolor{gray!20}16.863  &33.976  &\cellcolor{gray!20}53034.961  & 23.371  &\cellcolor{gray!20}15.534  &20.711  &\cellcolor{gray!20}20631.898  & 19.647\\
            English &\cellcolor{gray!20}8.386  & 9.06  &  \cellcolor{gray!20}25308.41  &  8.673  & \cellcolor{gray!20}7.851  & 8.501  &\cellcolor{gray!20} 8503.634  &  8.536\\
            Finnish &\cellcolor{gray!20}7.535  & 17.228  &\cellcolor{gray!20}57291.129  & 10.8  &  \cellcolor{gray!20}6.802  & 8.291  & \cellcolor{gray!20}15942.838  & 8.366\\
            French  &\cellcolor{gray!20}13.485  &22.26  & \cellcolor{gray!20}40576.059  & 16.776  &\cellcolor{gray!20}12.361  &15.653  &\cellcolor{gray!20}17057.102  & 15.247\\
            German  &\cellcolor{gray!20}18.195  &30.792  &\cellcolor{gray!20}73363.977  & 24.122  &\cellcolor{gray!20}16.678  &21.223  &\cellcolor{gray!20}29565.832  & 20.85\\
            Greek  &\cellcolor{gray!20} 3.843  & 6.028  & \cellcolor{gray!20}448650.156  &5.156  &\cellcolor{gray!20} 3.609  & 4.337  & \cellcolor{gray!20}162718.406  &4.393\\
            Hungarian  &\cellcolor{gray!20} 16.01  & 38.07  & \cellcolor{gray!20}65834.5  &24.309  &\cellcolor{gray!20}14.226  &22.761  &\cellcolor{gray!20}18880.131  & 21.956\\
            Indonesian  &\cellcolor{gray!20}46.324  &74.273  &\cellcolor{gray!20}37144.125  & 63.18  &\cellcolor{gray!20} 39.1  &47.835  &\cellcolor{gray!20}13521.396  & 42.72\\
            Italian  &\cellcolor{gray!20}14.685  &29.151  &\cellcolor{gray!20}53119.184  & 18.854  &\cellcolor{gray!20}13.4  &18.214  &\cellcolor{gray!20}20116.324  & 17.648\\
            Japanese  &\cellcolor{gray!20}10.852  &19.887  &\cellcolor{gray!20}420724.469  &15.101  &\cellcolor{gray!20}10.068  &12.853  &\cellcolor{gray!20}165031.688  &11.74\\
            Korean  &\cellcolor{gray!20}4.952  & 9.914  & \cellcolor{gray!20}98683.523  & 6.416  &\cellcolor{gray!20} 4.709  & 5.961  & \cellcolor{gray!20}74944.906  & 5.589\\
            Malay  & \cellcolor{gray!20}77.124  &133.861  &\cellcolor{gray!20}35202.762  & 117.684  &\cellcolor{gray!20}49.596  &60.177  &\cellcolor{gray!20}14545.072  & 59.499\\
            Malayalam  &\cellcolor{gray!20} 5.111  & 7.67  &\cellcolor{gray!20}406890.344  &7.048  & \cellcolor{gray!20}5.023  & 6.102  & \cellcolor{gray!20}307968.656  &5.882\\
            Norwegian  & \cellcolor{gray!20}14.241  &28.603  &\cellcolor{gray!20}36071.082  & 19.924  &\cellcolor{gray!20}13  &16.698  &\cellcolor{gray!20}12674.245  & 17.278\\
            Persian  &\cellcolor{gray!20}6.518  & 10.498  &\cellcolor{gray!20}114729.328  &8.9  &\cellcolor{gray!20}6.201  & 8.181  &\cellcolor{gray!20} 51444.336  & 7.524\\
            Polish  &\cellcolor{gray!20}12.475  &25.814  &\cellcolor{gray!20}82658.328  & 17.513  &\cellcolor{gray!20}11.002  &15.854  &\cellcolor{gray!20}22525.287  & 15.69\\
            Portuguese  &\cellcolor{gray!20}15.215  &27.788  &\cellcolor{gray!20}44236.961  & 19.786  &\cellcolor{gray!20}13.785  &17.408  &\cellcolor{gray!20}16310.681  & 16.81\\
            Romanian  &\cellcolor{gray!20}10.825  &21.796  &\cellcolor{gray!20}43364.27  &15.351  &\cellcolor{gray!20}9.565  & 12.499  &\cellcolor{gray!20}18184.531  & 12.201\\
            Russian  &\cellcolor{gray!20}11.883  &25.488  &\cellcolor{gray!20}233055.625  &16.334  &\cellcolor{gray!20}10.623  &15.444  &\cellcolor{gray!20}146091.188  &15.199\\
            Spanish  &\cellcolor{gray!20}16.876  &28.496  &\cellcolor{gray!20}44100.289  & 21.306  &\cellcolor{gray!20}15.733  &20.854  &\cellcolor{gray!20}18918.979  & 20.015\\
            Swahili  &\cellcolor{gray!20}91.953  &148.779  &\cellcolor{gray!20}33542.359  & 140.24  &\cellcolor{gray!20}86.072  &92.409  &\cellcolor{gray!20}11372.807  & 79.385\\
            Swedish  &\cellcolor{gray!20}14.643  &26.498  &\cellcolor{gray!20}65648.586  & 19.735  &\cellcolor{gray!20}13.159  &16.588  &\cellcolor{gray!20}21467.172  & 16.731\\
            Tamil  &\cellcolor{gray!20} 4.159  & 5.781  & \cellcolor{gray!20}446966.188  &5.4  &\cellcolor{gray!20}4.047  & 4.911  & \cellcolor{gray!20}360624.969  &4.647\\
            Turkish  &\cellcolor{gray!20}11.17  & 20.672  &\cellcolor{gray!20}33287.883  & 16.462 & \cellcolor{gray!20}9.695  & 12.298  &\cellcolor{gray!20}15661.532  & 12.168\\
            Ukrainian  & \cellcolor{gray!20}10.564  &18.353  &\cellcolor{gray!20}189824.422  &12.328  &\cellcolor{gray!20}8.811  & 10.289  &\cellcolor{gray!20}134138.078  &10.31\\
            Vietnamese  &\cellcolor{gray!20}5.804  & 11.447  &\cellcolor{gray!20}36745.988  & 7.42  &\cellcolor{gray!20}5.405  & 6.68  &\cellcolor{gray!20}11952.208  & 6.529\\

            \bottomrule
            \end{tabular}
            }
            \caption{LLaMA perplexity on 30 languages when the perturbation ratio is 3\%. `Top' and `Bottom' respectively indicate the $N$ parameters that exhibited the greatest and least change during the fine-tuning across the six languages. `Random' denotes the selection of $N$ parameters at random, while `Base' represents no perturbation at all. Here, $N$ represents 3\% of the total number of parameters.}
            \label{tab:30_scatter}
        \end{table*}    
As shown in Table \ref{tab:var_proportion}, LLaMA is fine-tuned separately using six languages. Approximately 0.981\% of parameters show a maximum variation of no more than 3\% of their original values across all six languages, while 16.382\% show a minimum variation of at least 5\% of their original values. 
This indicates two distinct sets of parameters categorized by their magnitude of change during language fine-tuning.
One set tends to remain consistent across all language fine-tuning (referred to as the `Bottom' region), while the other shows a propensity for change (referred to as the `Top' region). We posit that the `Bottom' region corresponds to the core region of linguistic competence, substantiated by the following evidence:

\textbf{Logical Evidence}: As discussed in \ref{sec:pretrain}, during the pre-training phase, LLMs effectively learn abstract phonological, morphological, syntactical, and semantic rules characterizing human languages. These rules form the foundation of LLMs' linguistic competence, enabling them to process various complex language phenomena and generate fluent natural language text. Naturally, input texts in the fine-tuning and pre-training stages should not differ fundamentally in basic linguistic rules, unless these languages originate from non-human sources, such as spam text online. Hence, the linguistic competence region within LLMs shouldn't undergo drastic changes during language fine-tuning.

\textbf{Empirical Evidence 1}: Table \ref{tab:chn_scatter} illustrates that even a 1\% perturbation in the `Bottom' region leads to a sharp increase in perplexity, reaching nearly $100,000$, indicating a complete loss of linguistic competence. In contrast, perturbing the `Top' region results in model perplexity comparable to random perturbations of equal magnitude, with no significant impact on the model's linguistic competence. Expanding our evaluations to 30 languages, as shown in Table \ref{tab:30_scatter}, yields consistent findings: perturbing the `Bottom' region deprives LLaMA of its capability across all 30 languages. This suggests the model's linguistic competence is directly influenced by the `Bottom' region, while perturbations in the `Top' region don't have a significant direct impact on language and are analogous to random perturbations.

\textbf{Empirical Evidence 2}: In the experiment corresponding to Table \ref{tab:freeze}, we initially perturbs various regions within LLaMA. Consistent with the findings from Tables \ref{tab:chn_scatter} and \ref{tab:30_scatter}, perturbing the `Bottom' region leads to a loss of linguistic competence, whereas the `Top' region don't. However, in this experiment, we sought to ascertain if LLaMA could reacquire its lost linguistic competence. Thus, we train on different amounts of Chinese Zhihu corpus and evaluate on Chinese Wechat and English Falcon corpora. The results indicate that if the `Bottom' region is perturbed and frozen, the model have to relearn basic language rules in other regions based on the provided Chinese Zhihu corpus, but these rules are inherently biased towards Chinese. Consequently, while its proficiency in Chinese is restored, the English perplexity remains high ($1276.354$ and $13918.666$, respectively). If the `Bottom' region is perturbed but not frozen, the model can rebuild its linguistic competence in-place. As its proficiency in Chinese is restored, so is its proficiency in English. This implies that the `Bottom' region encodes generalizable fundamental linguistic competence."

        \begin{table*}
            \centering
            \resizebox{\linewidth}{!}{
            \begin{tabular}{cc ccc ccc}
            \toprule
            \multirow{2}{*}{\makecell[c]{Testing\\Dataset\\(Language)}} & \multirow{2}{*}{\makecell[c]{\# Training \\Samples\\ (Chinese)}} & \multicolumn{3}{c}{Perturbation Ratio $=1\%$} & \multicolumn{3}{c}{Perturbation Ratio $=5\%$}\\
            \cmidrule(r){3-5}\cmidrule(r){6-8}
            & & \makecell[c]{Top \&\\Freeze} & \makecell[c]{Bottom \&\\Freeze} & \makecell[c]{Bottom \&\\Unfreeze}& \makecell[c]{Top \&\\Freeze} & \makecell[c]{Bottom \&\\Freeze} & \makecell[c]{Bottom \&\\Unfreeze}\\
            \midrule
            \multirow{6}{*}{\makecell[c]{Wechat\\(Chinese)}}
            &0K&   \cellcolor{gray!20}6.921&   73408.203&   \cellcolor{gray!20}73408.203&  27.656&  \cellcolor{gray!20}281376.219&  281376.219\\
            &2K&  \cellcolor{gray!20}6.539&   4424.779&  \cellcolor{gray!20}6.256&   13.233&  \cellcolor{gray!20}3233.563&  6.252\\
            &5K&  \cellcolor{gray!20}6.034&   359.694&  \cellcolor{gray!20}5.922&   6.485&   \cellcolor{gray!20}393.68&  5.923\\
            &10K&  \cellcolor{gray!20}6.031&   225.591&  \cellcolor{gray!20}5.972&   6.204&   \cellcolor{gray!20}288.387&  5.97\\
            &20K&  \cellcolor{gray!20}6.179&   22.904&  \cellcolor{gray!20}6.15&  6.295&   \cellcolor{gray!20}136.618&  6.17\\
            &50K&  \cellcolor{gray!20}5.711&   7.151&   \cellcolor{gray!20}5.698&   5.764&   \cellcolor{gray!20}20.85&   5.697\\
            \hline \hline
            \multirow{6}{*}{\makecell[c]{Falcon\\(English)}}
            &0K&   \cellcolor{gray!20}14.993&  31759.947&   \cellcolor{gray!20}31759.947&  26.086&  \cellcolor{gray!20}36518.203&   36518.203\\
            &2K&  \cellcolor{gray!20}14.683&  28371.539&   \cellcolor{gray!20}13.884&  21.868&  \cellcolor{gray!20}2378054.5&   13.877\\
            &5K&  \cellcolor{gray!20}15.199&  441158.719&  \cellcolor{gray!20}14.793&  16.344&  \cellcolor{gray!20}415355.688&  14.863\\
            &10K&  \cellcolor{gray!20}15.711&  1979024&  \cellcolor{gray!20}15.604&  16.131&  \cellcolor{gray!20}776365.563&  15.596\\
            &20K&  \cellcolor{gray!20}16.852&  9859.426&  \cellcolor{gray!20}16.39&   16.714&  \cellcolor{gray!20}438001.906&  16.506\\
            &50K&  \cellcolor{gray!20}20.083&  1276.354&  \cellcolor{gray!20}18.961&  20.47&   \cellcolor{gray!20}13918.666&   18.711\\
            \bottomrule
            \end{tabular}
            }
            \caption{Perturbation-freezing analysis in different regions of LLaMA. `Top/Bottom' denotes the perturbation region, while `Freeze/Unfreeze' indicates whether the corresponding region is frozen after perturbation. This experiment indicates that `Bottom' encodes generalizable fundamental linguistic rules.}
            \label{tab:freeze}
        \end{table*}    
        
\subsection{Dimensional Dependence of Linguistic Competence}
\label{sec:dependence}
\begin{figure*}[t]
	\centering
	\subfigure{
		\centering
	\begin{minipage}[t]{0.31\textwidth}
        \includegraphics[width=4.5cm]{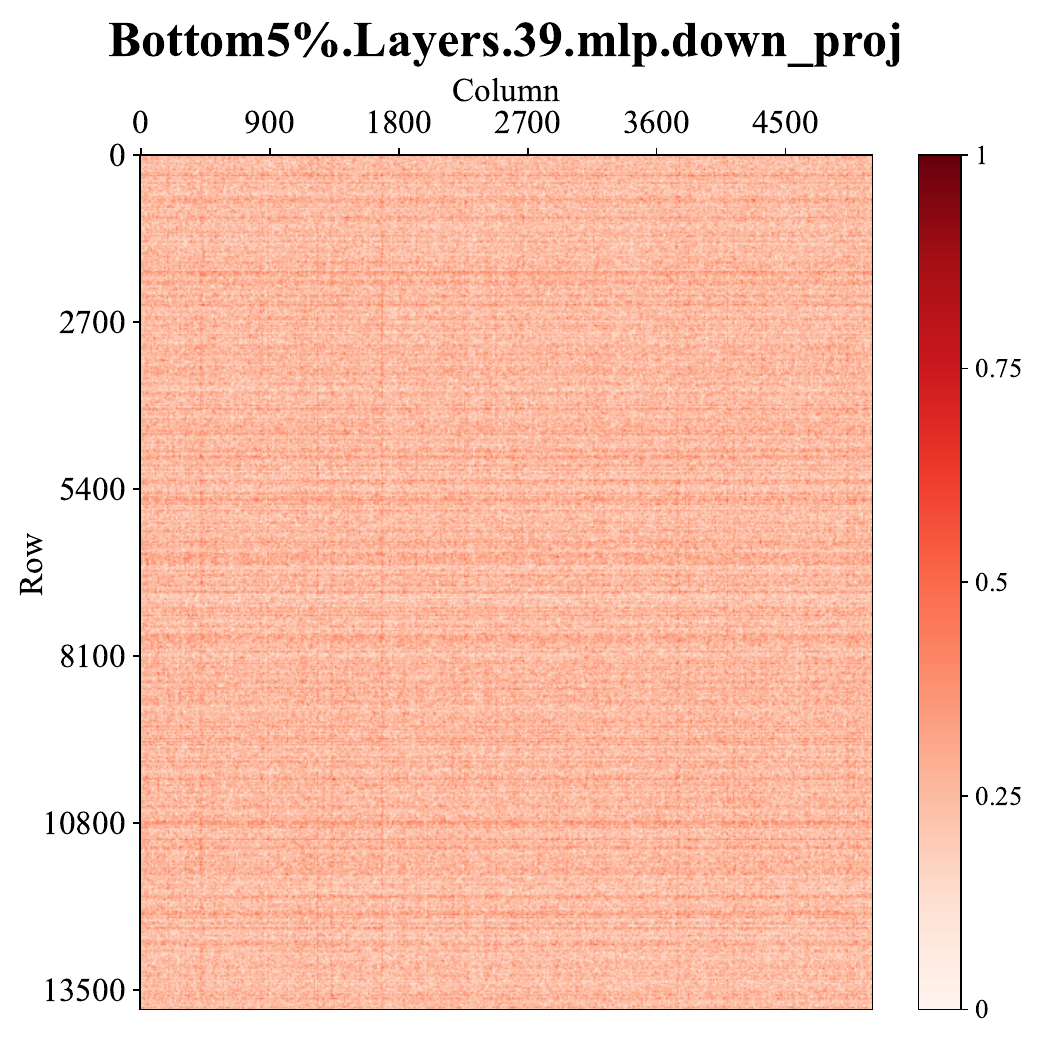}
        
	\end{minipage}
	}
	\subfigure{
		\centering
	\begin{minipage}[t]{0.31\textwidth}
            \includegraphics[width=4.5cm]{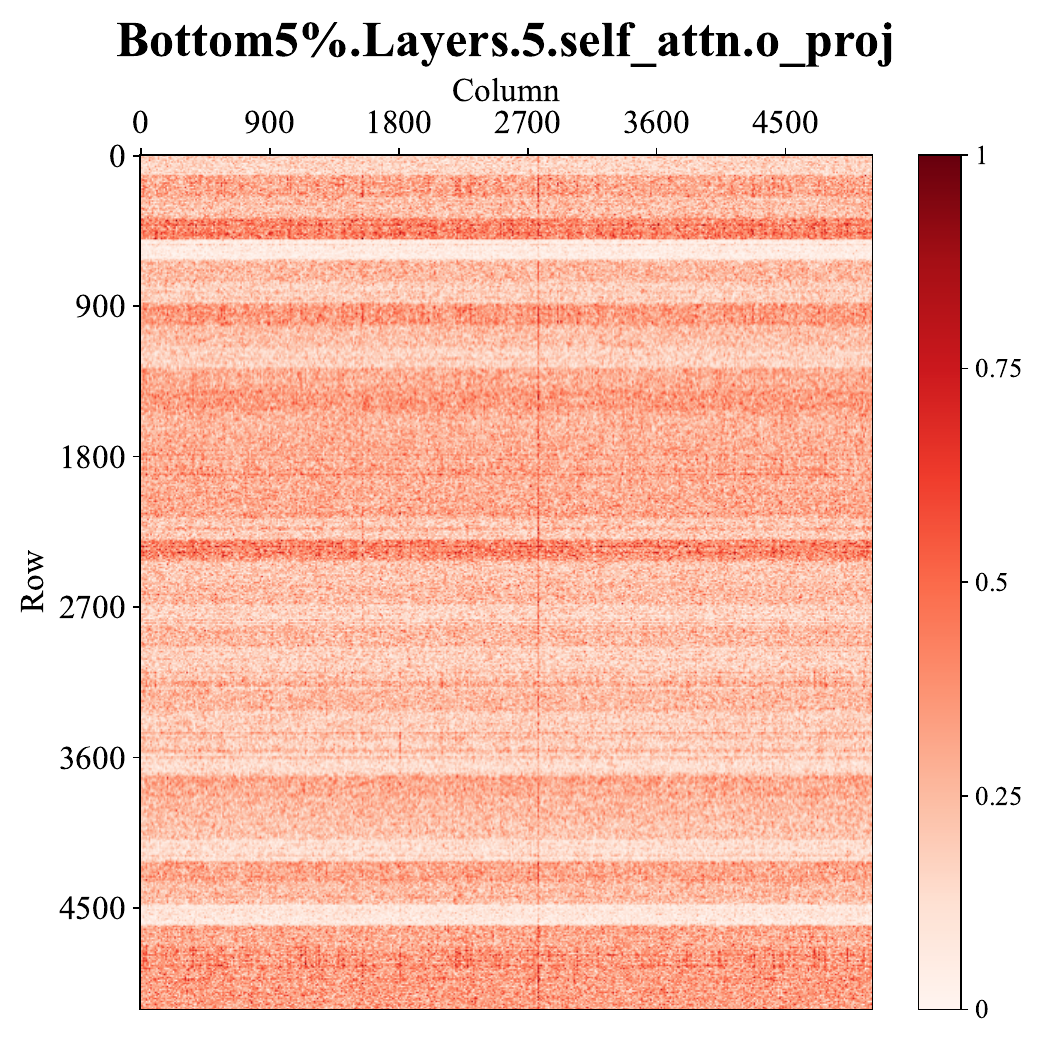}
	\end{minipage}
	}
	\subfigure{
		\centering
	\begin{minipage}[t]{0.31\textwidth}
		\includegraphics[width=4.5cm]{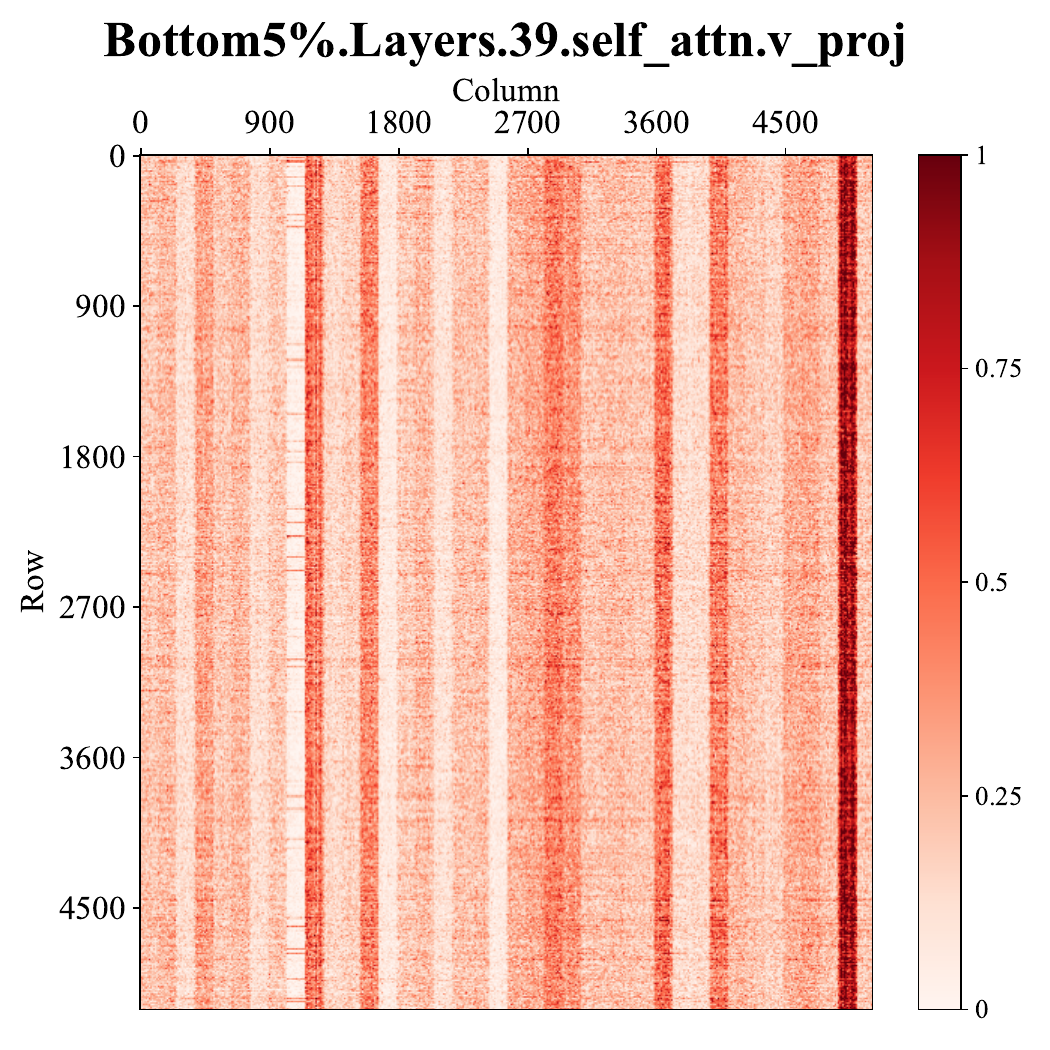}
	\end{minipage}
	}
\caption{Visualization of the linguistic competence region (the `Bottom' region). The scale from 0 to 1 (after normalization) represent the proportion of parameters within a $3\times3$ vicinity that belong to the Bottom region.}
\label{fig:visualize}
\end{figure*}

        \begin{table}
            \centering
            \begin{tabular}{ccc cccc}
            \toprule
            \multirow{2}{*}{\makecell[c]{Model}} & \multirow{2}{*}{\makecell[c]{\# Training \\Samples}} & \multirow{2}{*}{\makecell[c]{Number of\\Dimensions}}&\multicolumn{4}{c}{Attn.o(row), Attn.k/q/v+FFN.down(column)}\\
            \cline{4-7}
            & & &Top & Middle & Bottom & Random\\
            \midrule
            \multirow{4}{*}{\makecell[c]{LLaMA2-7B}}
            &100K& $1$ & \cellcolor{gray!20} 6.457 &  6.465 & \cellcolor{gray!20} 15.347 & 6.462\\
            &100K& $3$ &\cellcolor{gray!20} 6.467 &  6.465 &  \cellcolor{gray!20}27.429 & 6.486\\
            &100K& $5$ & \cellcolor{gray!20}6.492 &  6.48 & \cellcolor{gray!20}64181.316 &  6.552\\
            &100K& $10$ & \cellcolor{gray!20}6.553 &  6.524 &  \cellcolor{gray!20}50472.695 &  6.994\\
            \hline \hline
            \multirow{4}{*}{\makecell[c]{LLaMA2-13B}}&100K& $1$& \cellcolor{gray!20}5.934 &  5.931 & \cellcolor{gray!20} 8.273 &  5.939\\
            &100K& $3$ & \cellcolor{gray!20}5.948 &  5.936 &  \cellcolor{gray!20}175.321 & 5.961\\
            &100K& $5$ & \cellcolor{gray!20}5.972 &  5.943 &  \cellcolor{gray!20}170.144 & 5.975\\
            &100K& $10$ & \cellcolor{gray!20}6.068 &  5.957 & \cellcolor{gray!20} 226.649 & 6.033\\
            \hline \hline
            \multirow{4}{*}{\makecell[c]{LLaMA2-13B}}
            &10K& $1$ & \cellcolor{gray!20}5.932 &  5.928 &  \cellcolor{gray!20}8.552 &  5.932\\
            &10K& $3$ & \cellcolor{gray!20}5.939 &  5.944 &  \cellcolor{gray!20}151.521 & 5.959\\
            &10K& $5$ & \cellcolor{gray!20}5.961 &  6.061 &  \cellcolor{gray!20}213.776 & 5.958\\
            &10K& $10$ & \cellcolor{gray!20}6.049 &  5.115 &  \cellcolor{gray!20}21871.451 &  5.979\\
            \bottomrule
            \end{tabular}
            \caption{Perplexity of LLaMA after perturbing certain dimensions in the attention (Attn) and feedforward (FFN) layers. Here, 'Top', 'Middle', and 'Bottom' refer to the dimensions with the most, moderate, and least variation during fine-tuning across six languages, respectively. 'Random' denotes an equivalent number of dimensions chosen at random for comparison.}
            \label{tab:att_ffn}
        \end{table}

    To provide a more intuitive revelation of the spatial distribution characteristics of the linguistic competence region within the model, we visualize the `Bottom' region. As shown in Figure \ref{fig:visualize}, whether in the attention mechanism layer or the feed-forward layer, the linguistic region displays a distinct concentration in both the rows and columns of the matrices. More visualization results can be found in Figures \ref{fig:app_visualize_13b_q}-\ref{fig:app_visualize_13b_down_bot} in the appendix. Such distribution features seem to imply that the model's linguistic competence is concentrated in specific dimensions.

    To delve deeper into this observation, we adopt various strategies to perturb the parameters of the matrices. Instead of discretely perturbing different parameters, we selectively disturb certain rows or columns, especially those dimensions encompassing a significant number of `Bottom' region parameters, termed as `Bottom dimensions'. As illustrated in Table \ref{tab:att_ffn}, we attempt to perturb the columns of FFN.down and Attn.k/q/v, as well as the rows of Attn.o. The results indicate that perturbing just these `Bottom dimensions' leads to a substantial decline in the model's linguistic competence. In comparison to random perturbations, disturbances to the `Top' and `Middle' dimensions do not yield noticeable effects.

    It's noteworthy that the columns of the Attn.k/q/v matrices in the attention layer, as well as the rows of the Attn.o matrix, correspond to different attention head parameters (See Figure \ref{fig:app_structure_visual} (left) for a visual illustration). Conversely, the rows of the Attn.k/q/v matrices and the columns of the Attn.o matrix are closely associated with features in the representation space. We perturb the Bottom dimensions in the attention layer under both of these settings, with the results displayed in Tables \ref{tab:att} and \ref{tab:att_reverse}. Table \ref{tab:att} reveals that perturbing the Bottom dimensions continues to produce more detrimental effects than other dimensions. The visualizations in Figure \ref{fig:visualize} show that these dimensions are largely concentrated in a few attention heads, suggesting that some attention heads contribute more significantly to the model's linguistic competence. Table \ref{tab:att_reverse} indicates that the perturbations under the second setting cause more damage than the first. Considering that, in the second setting, the Bottom dimensions in the matrix directly interact with the corresponding features in the representational space, we can conjecture that these features are tightly linked with the model's linguistic competence.

        \begin{table}[t]
            \centering
            \begin{tabular}{ccc cccc}
            \toprule
            \multirow{2}{*}{\makecell[c]{Model}} & \multirow{2}{*}{\makecell[c]{\# Training \\Samples}} & \multirow{2}{*}{\makecell[c]{Number of \\Dimensions}}&\multicolumn{4}{c}{Attn.o(row)+Attn.k/q/v(column)}\\
            \cline{4-7}
            & & &Top & Middle & Bottom & Random\\
            \midrule
            \multirow{4}{*}{\makecell[c]{LLaMA2-7B}}
            &100K& $1$ & \cellcolor{gray!20}6.463 &  6.458 &  \cellcolor{gray!20}7.032 &  6.459\\
            &100K& $3$ & \cellcolor{gray!20}6.47 & 6.465 &  \cellcolor{gray!20}7.654 &  6.464\\
            &100K& $5$ & \cellcolor{gray!20}6.482 &  6.466 &  \cellcolor{gray!20}8.243 &  6.538\\
            &100K& $10$ & \cellcolor{gray!20}6.533 &  6.49 & \cellcolor{gray!20}29.798 & 6.846\\
            \hline \hline
            \multirow{4}{*}{\makecell[c]{LLaMA2-13B}}&100K& $1$& \cellcolor{gray!20}5.933 &  5.929 &  \cellcolor{gray!20}6.231 &  5.937\\
            &100K& $3$ & \cellcolor{gray!20}5.94 & 5.929 &  \cellcolor{gray!20}7.1 & 5.946\\
            &100K& $5$ & \cellcolor{gray!20}5.957 &  5.93 & \cellcolor{gray!20}7.486 &  5.964\\
            &100K& $10$ & \cellcolor{gray!20}6.036 &  5.939 & \cellcolor{gray!20} 8.407 &  6.008\\
            \hline \hline
            \multirow{4}{*}{\makecell[c]{LLaMA2-13B}}
            &10K& $1$ & \cellcolor{gray!20}5.928 &  5.929 &  \cellcolor{gray!20}6.279 &  5.932\\
            &10K& $3$ &\cellcolor{gray!20} 5.931 &  5.943 &  \cellcolor{gray!20}7.131 &  5.952\\
            &10K& $5$ & \cellcolor{gray!20}5.942 &  6.061 &  \cellcolor{gray!20}6.752 &  5.957\\
            &10K& $10$ & \cellcolor{gray!20}6.033 &  6.091 &  \cellcolor{gray!20}7.509 &  5.965\\
            \bottomrule
            \end{tabular}
            \caption{Perplexity of LLaMA after perturbing certain dimensions in attention (Attn) layers. Here, 'Top', 'Middle', and 'Bottom' refer to the dimensions with the most, moderate, and least variation during fine-tuning across six languages, respectively. 'Random' denotes an equivalent number of dimensions chosen at random for comparison.}
            \label{tab:att}
        \end{table}

        \begin{table}[t]
            \centering
            \begin{tabular}{ccc cccc}
            \toprule
            \multirow{2}{*}{\makecell[c]{Model}} & \multirow{2}{*}{\makecell[c]{\# Training \\Samples}} & \multirow{2}{*}{\makecell[c]{Number of\\Dimensions}}&\multicolumn{4}{c}{Attn.o(column)+Attn.k/q/v(row)}\\
            \cline{4-7}
            & & &Top & Middle & Bottom & Random\\
            \midrule
            \multirow{4}{*}{\makecell[c]{LLaMA2-7B}}
            &100K& $1$ & \cellcolor{gray!20} 6.453 &  6.456 &  \cellcolor{gray!20}6.686 &  6.453\\
            &100K& $3$ &\cellcolor{gray!20} 6.455 &  6.456 & \cellcolor{gray!20} 8.436 &  6.453\\
            &100K& $5$ &\cellcolor{gray!20} 6.465 &  6.468 &  \cellcolor{gray!20}80.286 & 6.46\\
            &100K& $10$ &\cellcolor{gray!20} 6.476 &  6.477 & \cellcolor{gray!20} 66.84 &  6.769\\
            \hline \hline
            \multirow{4}{*}{\makecell[c]{LLaMA2-13B}}&100K& $1$& \cellcolor{gray!20}5.93 & 5.926 & \cellcolor{gray!20} 6.078 &  5.927\\
            &100K& $3$ &\cellcolor{gray!20}5.931 &  5.93 & \cellcolor{gray!20}18.777 & 5.928\\
            &100K& $5$ &\cellcolor{gray!20} 5.931 &  5.929 & \cellcolor{gray!20} 5283.898 & 5.93\\
            &100K& $10$ &\cellcolor{gray!20} 5.934 &  5.937 & \cellcolor{gray!20} 6944.889 & 5.943\\
            \hline \hline
            \multirow{4}{*}{\makecell[c]{LLaMA2-13B}}
            &10K& $1$ &\cellcolor{gray!20} 5.929 &  5.927 & \cellcolor{gray!20} 6.073 &  5.928\\
            &10K& $3$ &\cellcolor{gray!20} 5.932 &  5.93 &\cellcolor{gray!20} 81.158 & 5.932\\
            &10K& $5$ & \cellcolor{gray!20}5.935 &  5.931 & \cellcolor{gray!20} 10054.732 &  5.929\\
            &10K& $10$ & \cellcolor{gray!20}5.936 &  5.936 & \cellcolor{gray!20} 2037.702 & 5.934\\
            \bottomrule
            \end{tabular}
            \caption{Perplexity of LLaMA after perturbing certain dimensions in attention (Attn) layers. Different from Table \ref{tab:att}, in this table, the columns of the Attn.O and the rows of the Attn.K/Q/V are perturbed.}
            \label{tab:att_reverse}
        \end{table}

\makeatletter
\newcommand\figcaption{\def\@captype{figure}\caption}
\newcommand\tabcaption{\def\@captype{table}\caption}
\makeatother

\begin{figure}[tb]
	\flushleft

		\begin{minipage}[h]{0.5\linewidth}
			\centering
			\includegraphics[scale=0.5]{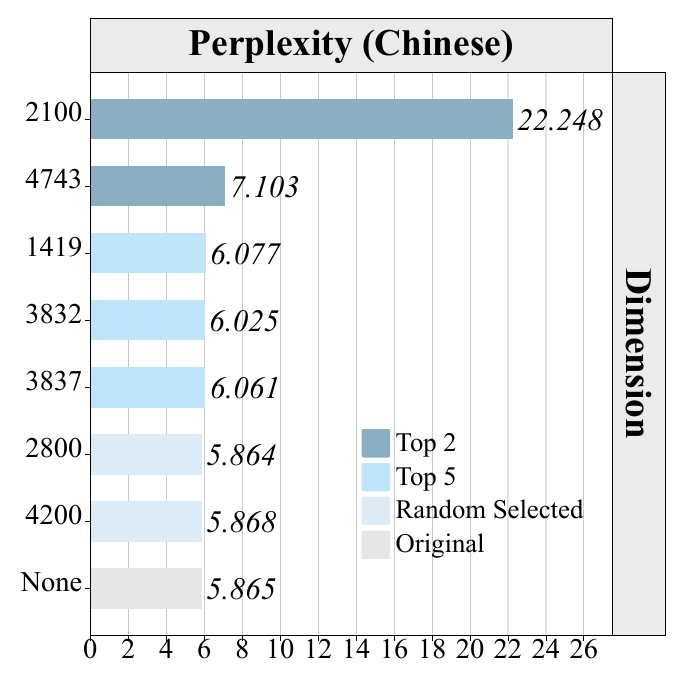}
	        \figcaption{The perplexity of the LLaMA2-13B when perturbing the same single dimension across all layers. In this experiment, we perturb the Att.O and FFN.down matrices of each layer. 'Topk' represents the top k dimensions that disrupt the model the most. 'Random selected' refers to a randomly chosen dimension. 'None' indicates that no dimensions are disrupted.}
	        \label{fig:layers}
		\end{minipage}%
\hspace{0.1in}	
\begin{minipage}{0.45\textwidth}
			\centering
            \begin{tabular}{ccc}
            \toprule
            Perturbation&Region& Perplexity\\
            \midrule
            \rowcolor{gray!20}-&- &  5.865\\
            Reset 1&L0-N2100 & 5.866 \\
            Reset 1&L1-N2100 &  83224.078\\
            Reset 1&L1-N2800 & 5.860 \\
            Reset 1&L1-N4200 &  5.858\\
            \hline
            Mul 10&L0-N2100 & 5.866 \\
            Mul 10&L1-N2100 &  4363.462\\
            Mul 10&L1-N2800 & 5.859 \\
            Mul 10&L1-N4200 & 5.864 \\
            \bottomrule
            \end{tabular}
        
			\tabcaption{Perturbing a single weight parameter in the 2100th dimension of LLaMA2 13B is sufficient to cause the model to lose its language competence. Reset 1 represents resetting the parameter to 1 (the initial value before pre-training), Mul 10 represents multiplying the parameter by 10. L0 and L1 represent the 0th and 1st layers, respectively. N represents the input\_layer\_norm module, followed by the number indicating the dimension of the perturbed parameter.}
			\label{tab:norm}
		\end{minipage}
\end{figure}

\begin{figure*}[t]
        \includegraphics[width=\linewidth]{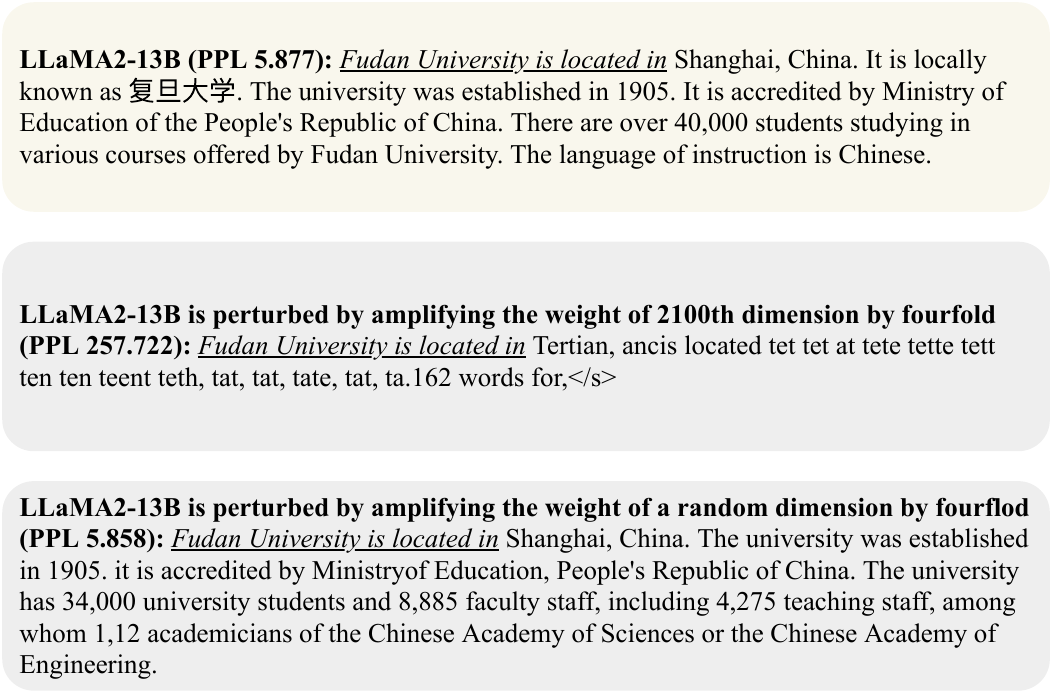}
        \caption{Comparison of linguistic competence. Perturbing a single parameter leads to complete language incapacity in LLaMA2-13B, a 13 billion-parameter LLM.}
        \label{fig:case}
    \end{figure*}  

\subsection{Perturbations in a Single Dimension or Even a Single Parameter Can Debilitate a Model's Linguistic Competence}

In Section \ref{sec:localization}, we define the core region of linguistic competence as the set of parameters that undergo the smallest changes during the language fine-tuning. In Section \ref{sec:dependence}, we observe a pronounced dimensionality dependence of these core parameters. However, the variation of parameters is not always consistent across different Transformer layers, implying that the key dimensions might differ from one layer to another. In this section, we explore whether specific dimensions significantly impact the model's linguistic competence. Surprisingly, among the 5120 dimensions of the LLaMA2 13B, dimensions 2100 and 4743 stand out as being particularly special. As illustrated in Figure \ref{fig:layers}, we iterate through the key dimensions mentioned in Section \ref{sec:dependence}, attempting to perturb the same dimension across all Transformer layers. The results revealed that the impact of dimensions 2100 and 4743 on the LLaMA2 13B substantially surpassed other dimensions, even when compared to the other three in the Top5 dimensions. In contrast, perturbing two randomly selected dimensions, such as dimensions 2800 and 4200, yield linguistic performance almost indistinguishable from the unperturbed state. Interestingly, a model with perturbed dimension 2800 even shows a slight improvement (5.864 vs. 5.865) in the perplexity metric compared to the unperturbed model.

Delving further, we find that even a slight modification to a single parameter in models with over 13 billion parameters can lead to a significant decline in its output quality. Specifically, each column in the Attn.o matrix of the attention layer and the FFN.down matrix of the feed-forward layer can be considered as the input weights of a neuron. Thus, perturbing a column can be seen as disturbing the input weights of a neuron. Viewed from another angle, if we disturb the output activation value of this neuron, a similar effect should be observed. Within LLaMA, there is a specific module called RMSNorm, where each dimension is associated with a weight. Perturbations to these weights can be regarded as disturbances to the output activation values of the corresponding neurons (In Figure \ref{fig:app_structure_visual} (right), we visually demonstrate how RMSNorm affects a column of the Attn.o and the FFN.down matrix). In Table \ref{tab:norm}, we discover that merely resetting the 2100th parameter in the input layer norm module of the first layer to its initial value causes LLaMA2 13B's PPL value to skyrocket from 5.865 to 83224.078. If this weight parameter is multiplied by 10, the PPL value also rises to 4363.462. This suggests that even minor changes to a single parameter can cause the model to lose nearly all of its linguistic competence. The effect of perturbing different parameters on the model varies. For instance, randomly altering the parameters at dimensions 2800 and 4200 doesn't noticeably impact the model. Interestingly, when we disturbed the parameter at the 2100th dimension in the 0th layer, the model's output remains unaffected.

To visually illustrate the impact of the linguistic competence region on the model's output quality, we use "Fudan University is located in" as a premise and observe the model's outputs under different parameter perturbations. The results are shown in Figure \ref{fig:case}. We perturbe LLaMA2-13B by amplifying the weight of 2100th dimension of RMSNorm module by fourfold. Compared to the original LLaMA2 13B model, the perturbed model completely loses its linguistic competence, producing nonsensical strings. As a control, when we perturb the weights corresponding to a randomly selected dimension, the model's PPL do not exhibit significant changes. In Figure \ref{fig:app_case} in the appendix, we further increase the perturbation magnitude to ten times the original weight and observe similar experimental results.


\subsection{The Dissociation Between Linguistic Competence and Knowledge}
    \begin{figure*}[t]
        \includegraphics[width=\linewidth]{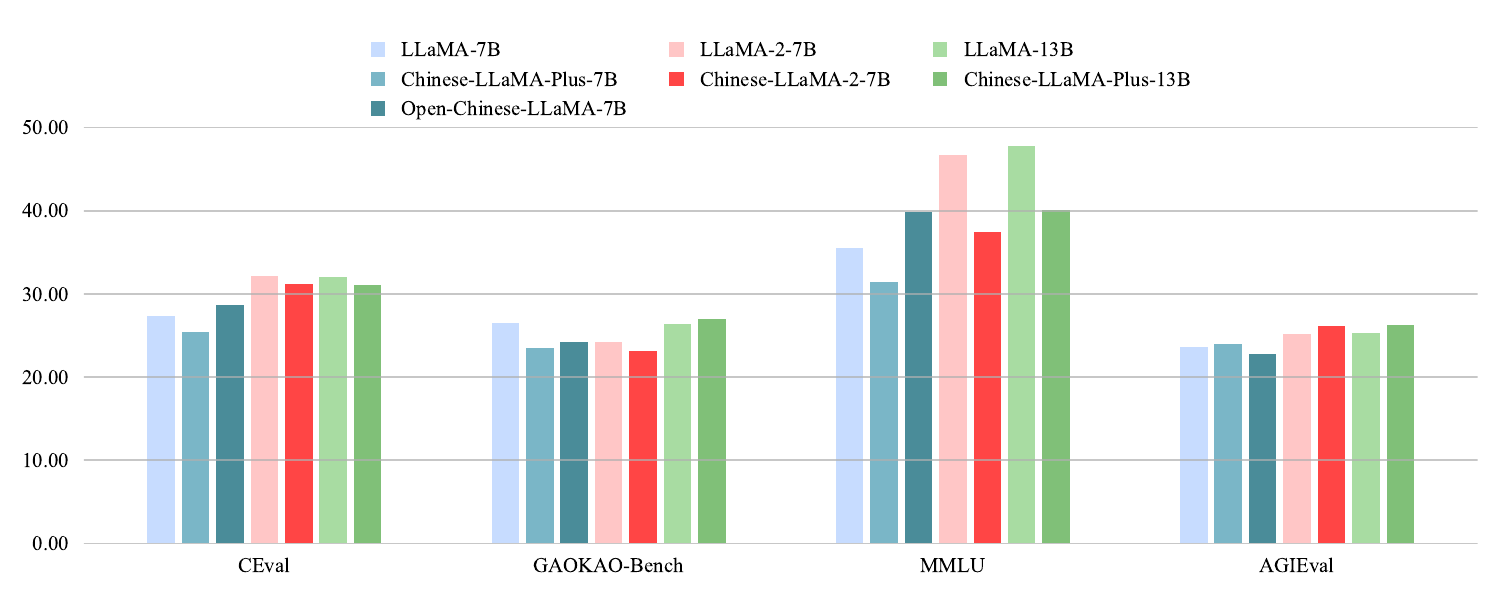}
        \caption{Knowledge-level evaluation results  on four benchmarks.}
        \label{fig:knowledge}
    \end{figure*}  
    
With the continuous growth of model size and pre-training data, many researchers believe that an enhancement in a model's linguistic competence will directly lead to an improvement in its knowledge and reasoning abilities. However, our research does not entirely support this viewpoint. Initially, to systematically verify whether the growth in a model's knowledge capability is directly related to the enhancement of its linguistic skills, we adopted four widely accepted knowledge evaluation standards: C-Eval \citep{huang2023ceval}, Gaokao-Bench \citep{zhang2023evaluating}, AGI-Eval \citep{zhong2023agieval}, and MMLU \citep{DBLP:journals/corr/abs-2009-03300}. In these assessments, we evaluated different versions of LLaMA, Chinese-LLaMA, and Open-Chinese-LLaMA, with results consolidated in Figure \ref{fig:knowledge}. Specifically, Chinese LLaMA 7B and Open Chinese LLaMA 7B are based on LLaMA 7B but underwent further Chinese pre-training on a base of 30B and 100B tokens respectively, leading to a significant improvement in their Chinese linguistic competence. Nevertheless, the scores of these two versions on C-Eval, Gaokao-Bench, and AGI-Eval were almost on par with the original LLaMA 7B. This implies that even if linguistic competence are enhanced, the corresponding knowledge reasoning capability doesn't necessarily improve. More importantly, we found that the LLaMA2-7B and LLaMA-13B, which had not undergone further Chinese pre-training, outperformed the Open Chinese LLaMA 7B across all four evaluation standards. Notably, the pre-training tokens of LLaMA2-7B stand at 2T, which is double that of LLaMA-7B, and the model size of LLaMA-13B is twice that of the 7B version. This highlights the crucial role of model scale and large-data pre-training in enhancing knowledge levels. In summary, our research reveals a distinction between linguistic competence and knowledge reasoning ability, suggesting that within LLMs, in addition to the linguistic region, there might also exist dedicated knowledge processing regions.

\section{Dissicusion and Future Work}
The core regions of linguistic competence and their dimensional dependence have guiding significance in the pre-training and fine-tuning of large language models. To achieve superior model performance, we believe the following recommendations are particularly important:

\textbf{Consideration of Data Ratios during Further Pre-training:}
\begin{enumerate}
    \item After pretraining, specific parameter regions of the language model are responsible for particular functions. Introducing a significant amount of knowledge that was missing during the pre-training may cause notable parameter shifts, potentially leading to a decline in model capabilities.
    \item For a set of fine-tuning data, consider mixing it with 5-10 times the original pre-training data before training.
\end{enumerate}

\textbf{Sensitivity of Linguistic Competence Regions in LLMs:}
\begin{enumerate}
    \item Overtraining with a small amount of data for many epochs might influence the linguistic competence region, subsequently impairing the model's overall capabilities.
    \item In supervised fine-tuning, to prevent substantial changes in key regions, one might consider adding general instruction data or original pre-training data.
\end{enumerate}

\textbf{Strict Noise Control and Adversarial Sample Generation in Training Data:}
\begin{enumerate}
    \item If pre-training data contains consecutive noise, such as repeated words or non-word sequences, it might trigger adjustments in specific dimensions, subsequently causing PPL fluctuations.
    \item If the supervised fine-tuning instructions contain numerous samples inconsistent with the original pre-training data, this could also result in adjustments in key dimensions, leading to a sharp decline in overall performance.
    \item Careful observation of the dynamic changes in parameters within core regions can guide the generation of adversarial samples, that is, understanding which data can adversely affect the parameters of the core regions.

\end{enumerate}
By adhering to these guidelines, one can ensure that large language models are trained and fine-tuned more effectively, maximizing their potential and minimizing potential pitfalls. In the future, we plan to delve deeper into the linguistic competence regions within large language models and their properties, such as the stability across multiple languages and inter-model consistency. Additionally, we will further explore potential functional regions and their interactions therein.
\section{Conclusions}
Inspired by cognitive science research, this paper investigates whether specific functional regions exist within LLMs. We identify a core region specifically responsible for language processing within LLMs. This region occupies only about 1\% of the model's parameters but plays a crucial role in maintaining the overall linguistic competence of the model. Invalid changes in the parameters of this region can severely impair the model's linguistic competence. We also observe a pronounced dimension dependence in the core region of linguistic competence. Surprisingly, in a large model like LLaMA-13B, which boasts 13 billion parameters, altering just one parameter could potentially inflict significant damage to its linguistic competence. This study further elucidates the relationship between linguistic competence and knowledge in large language models. We find that an improvement in linguistic competence does not necessarily imply an enhancement in knowledge level. This suggests the presence of a knowledge storage region in LLMs that operates independently of language processing. In summary, the findings of this paper shed new light on how the capabilities and knowledge are structured in large language models and help explain why the pre-training and fine-tuning processes of these large models differ significantly from their smaller predecessors.
\bibliography{iclr2021_conference}
\bibliographystyle{iclr2021_conference}

\appendix
\section{Appendix}
\subsection{The languages in evaluation corpus}
\label{sec:app_lang}
We use evaluation data composed of 30 languages to assess the model's linguistic competence. The 30 languages and their respective token counts are as follows:
Arabic (4702998), Chinese (2869208), Czech   (1362041), Danish  (36467), Dutch (3991305), English (1216599), Finnish (372303), French  (6755281), German  (2884921), Greek   (474622), Hungarian   (1229433), Indonesian  (19226), Italian (6332560), Japanese    (501899), Korean  (2730794), Malay   (5842), Malayalam   (1489244), Norwegian   (42289), Persian (1736589), Polish  (4948702), Portuguese  (7598161), Romanian    (1381598), Russian (5205716), Spanish (7163860), Swahili (630), Swedish (1450236), Tamil   (2920808), Turkish (2484186), Ukrainian   (455720), Vietnamese  (3606202).

\subsection{The Illustration of the Calculation Workflow}
\begin{figure*}[h]
        \includegraphics[width=\linewidth]{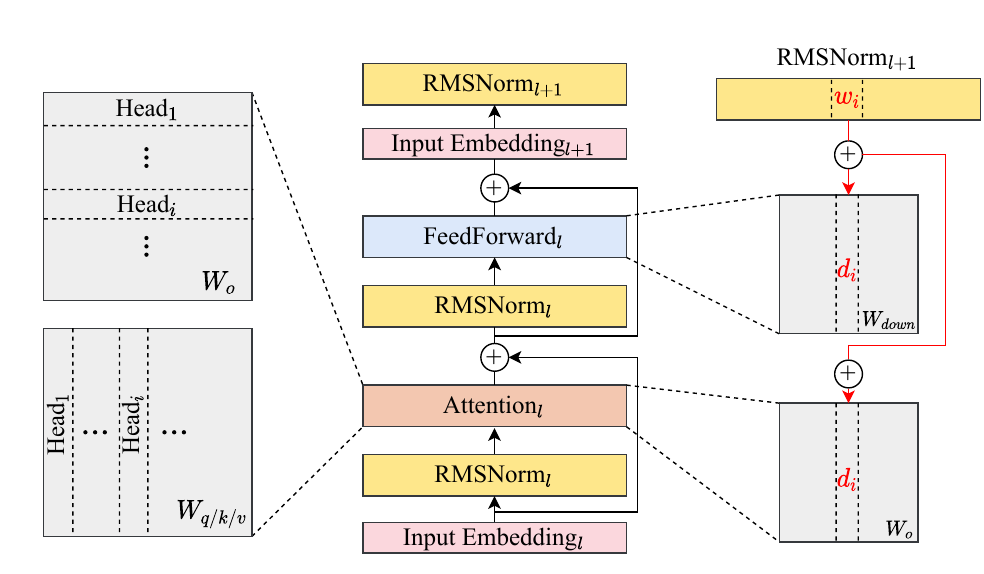}
        \caption{One can see from the left that each row of the Attn.o ($W_o$) corresponds to a particular attention head, and each column of the Attn.q/k/v ($W_{q/k/v}$) matrix corresponds to one as well. On the right, one can observe the perturbation applied to one weight within RMSNorm, which can be seen as affecting a column of the FFN.down and the Attn.o.}
        \label{fig:app_structure_visual}
    \end{figure*}  

\subsection{Output Comparison and Region Visualization}
\begin{figure*}[t]
        \includegraphics[width=\linewidth]{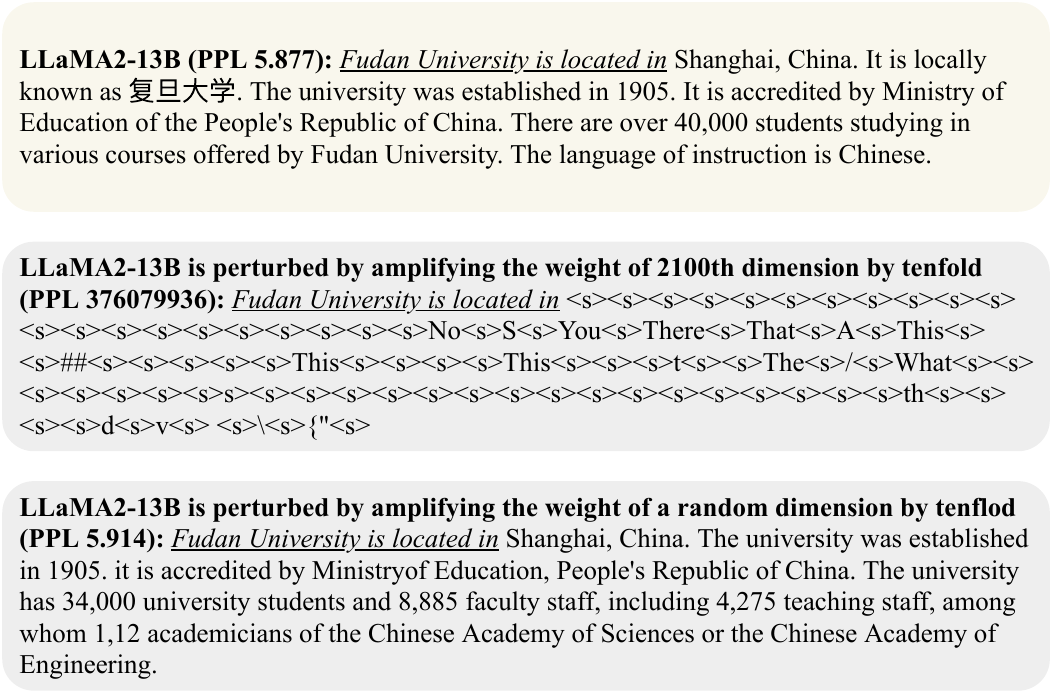}
        \caption{Comparison of linguistic competence. Perturbing a single parameter leads to complete language incapacity in LLaMA2-13B, a 13 billion-parameter LLM.}
        \label{fig:app_case}
    \end{figure*}

\begin{figure*}[h]
	\centering
	\subfigure{
		\centering
	\begin{minipage}[t]{0.23\textwidth}
        \includegraphics[width=3.6cm]{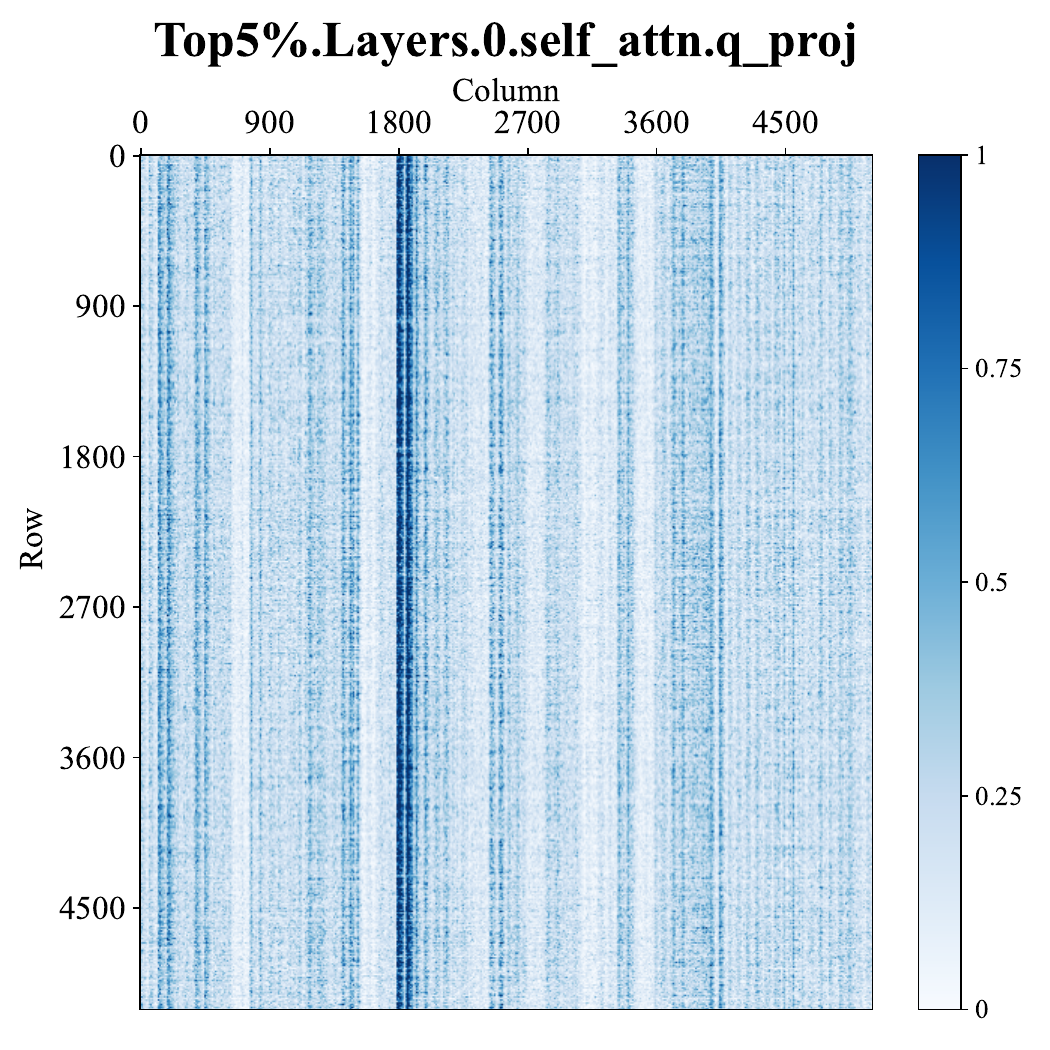}
        \includegraphics[width=3.6cm]{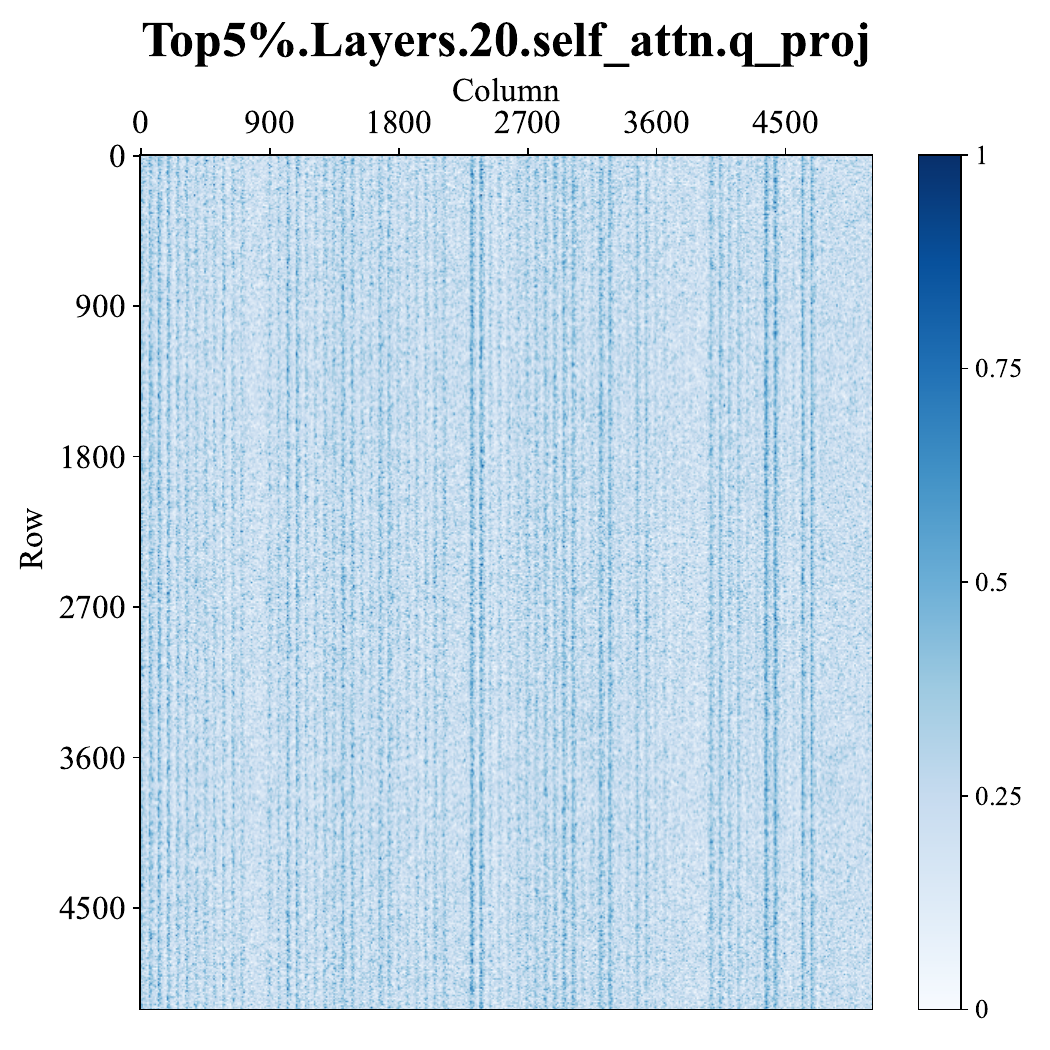}
	\end{minipage}
	}
	\subfigure{
		\centering
	\begin{minipage}[t]{0.23\textwidth}
	       \includegraphics[width=3.6cm]{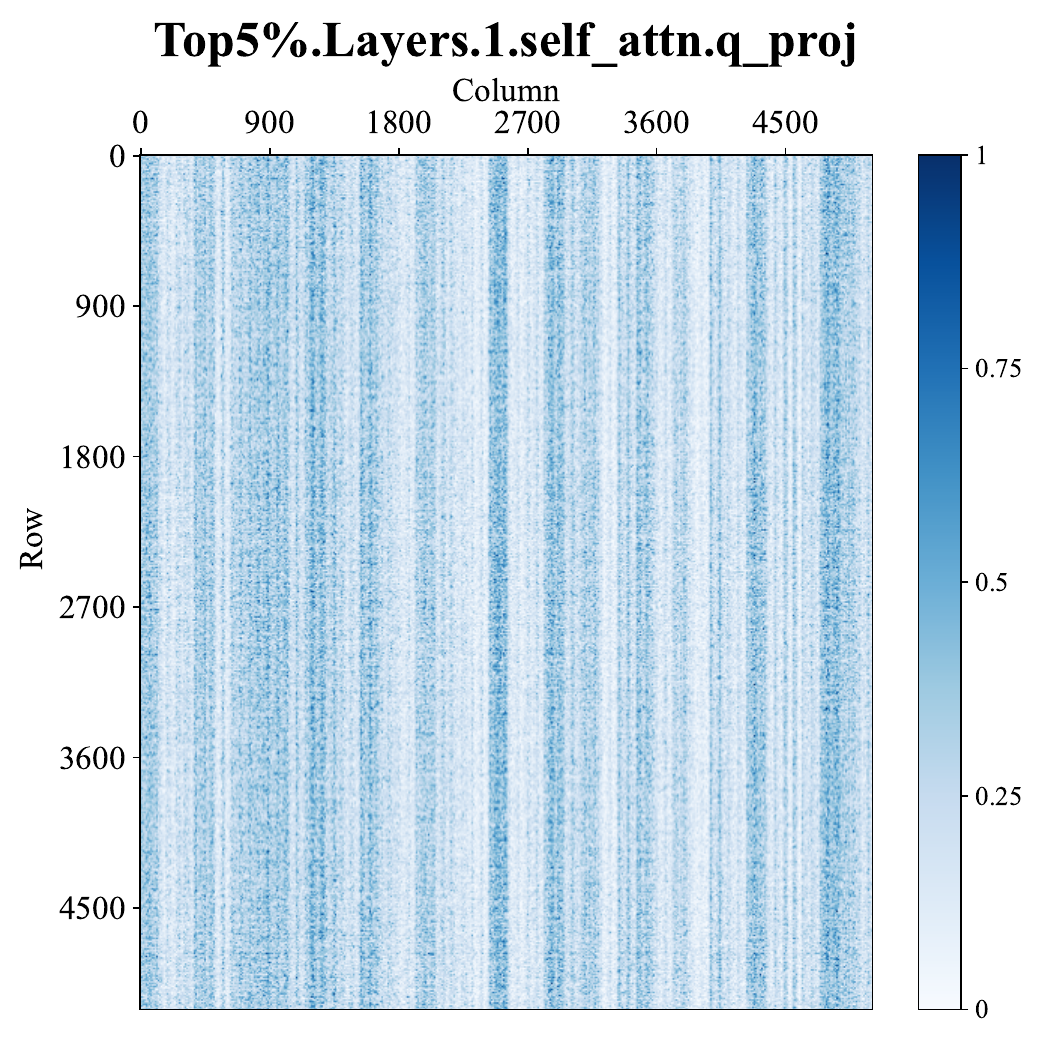}
            \includegraphics[width=3.6cm]{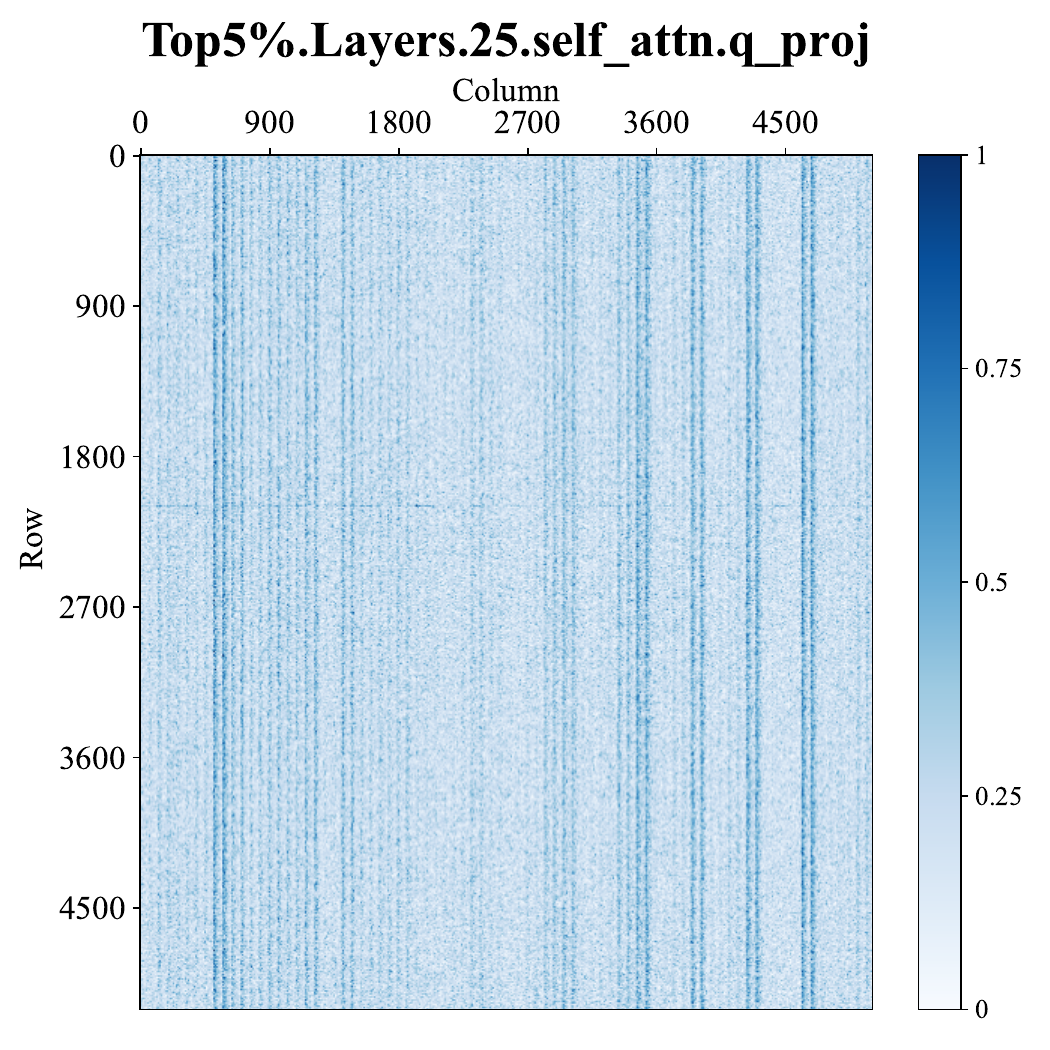}
	\end{minipage}
	}
	\subfigure{
		\centering
	\begin{minipage}[t]{0.23\textwidth}
		\includegraphics[width=3.6cm]{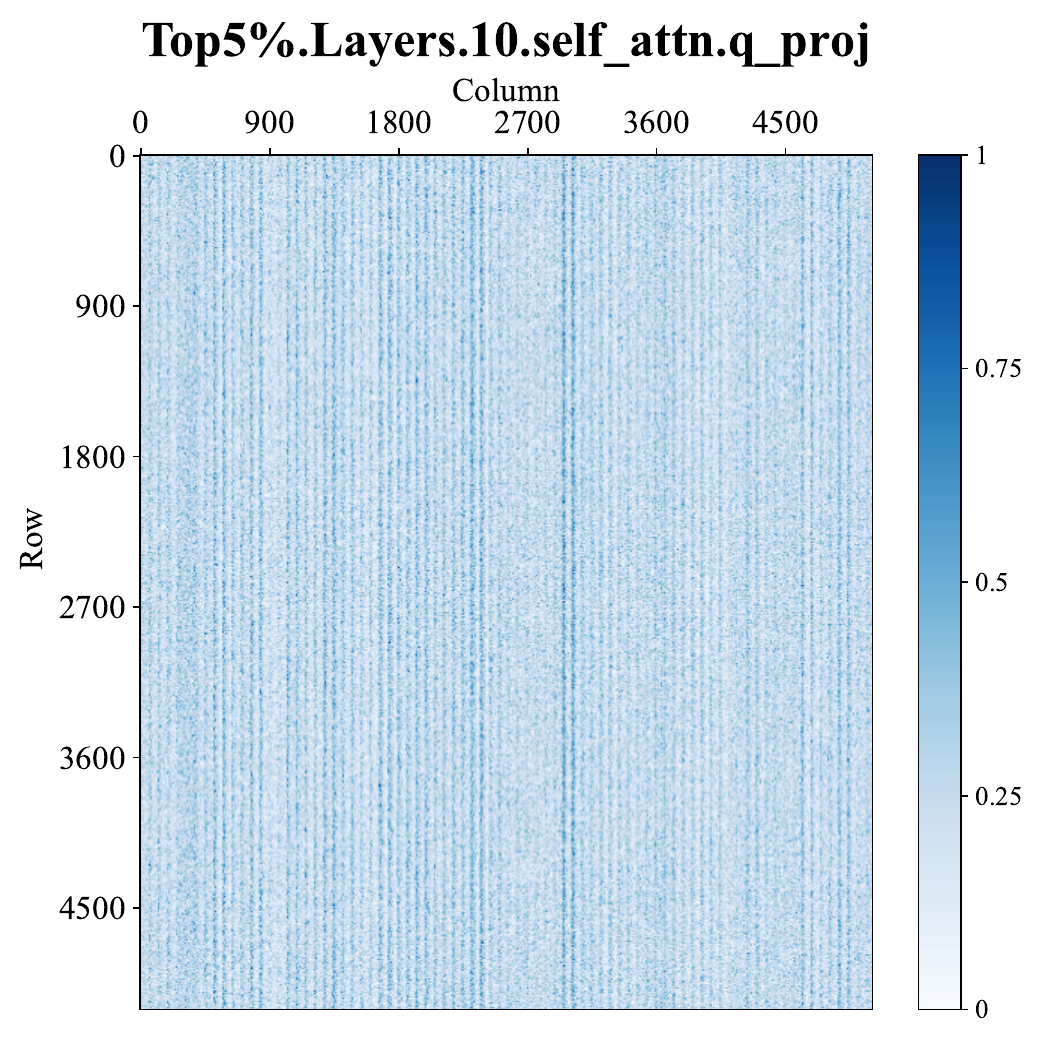}
		\includegraphics[width=3.6cm]{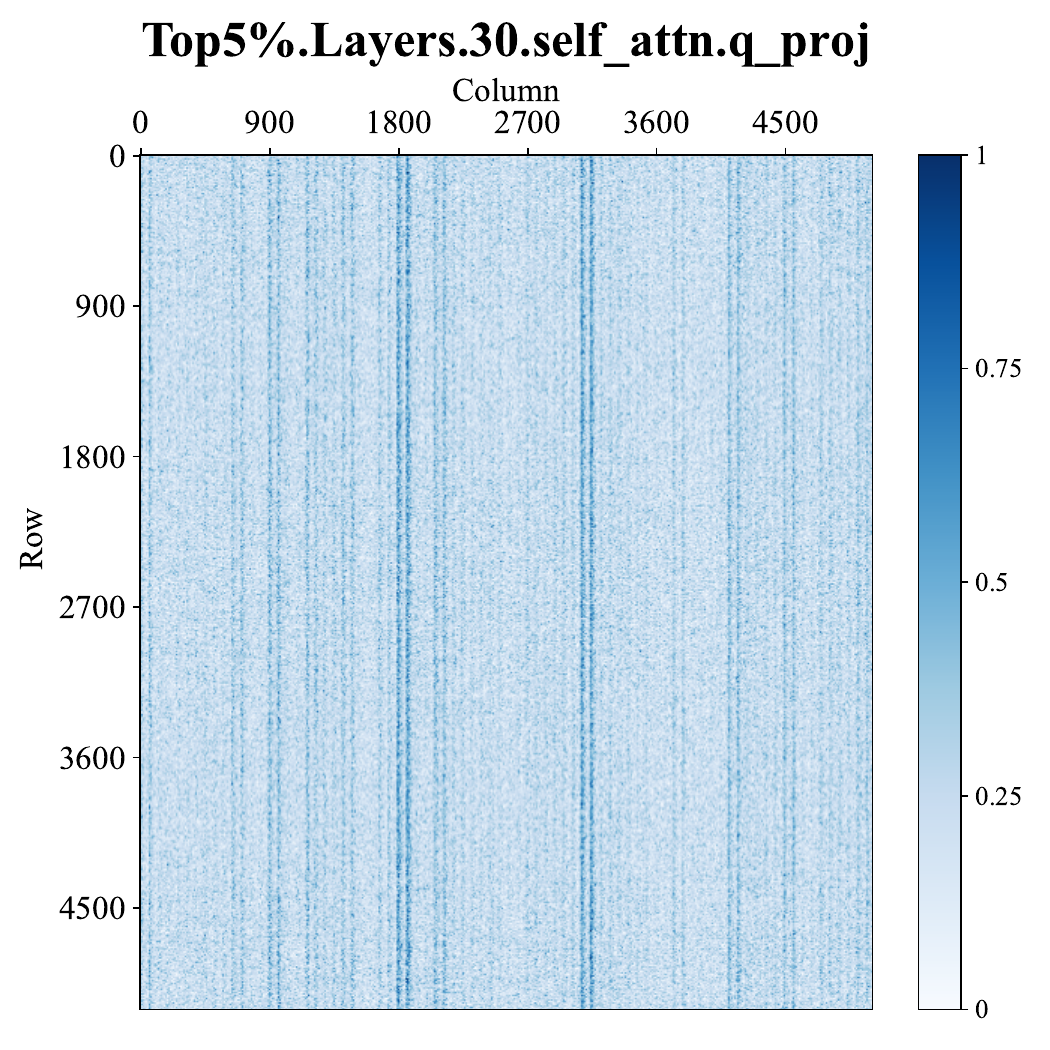}
	\end{minipage}
	}
 \subfigure{
		\centering
	\begin{minipage}[t]{0.23\textwidth}
		\includegraphics[width=3.6cm]{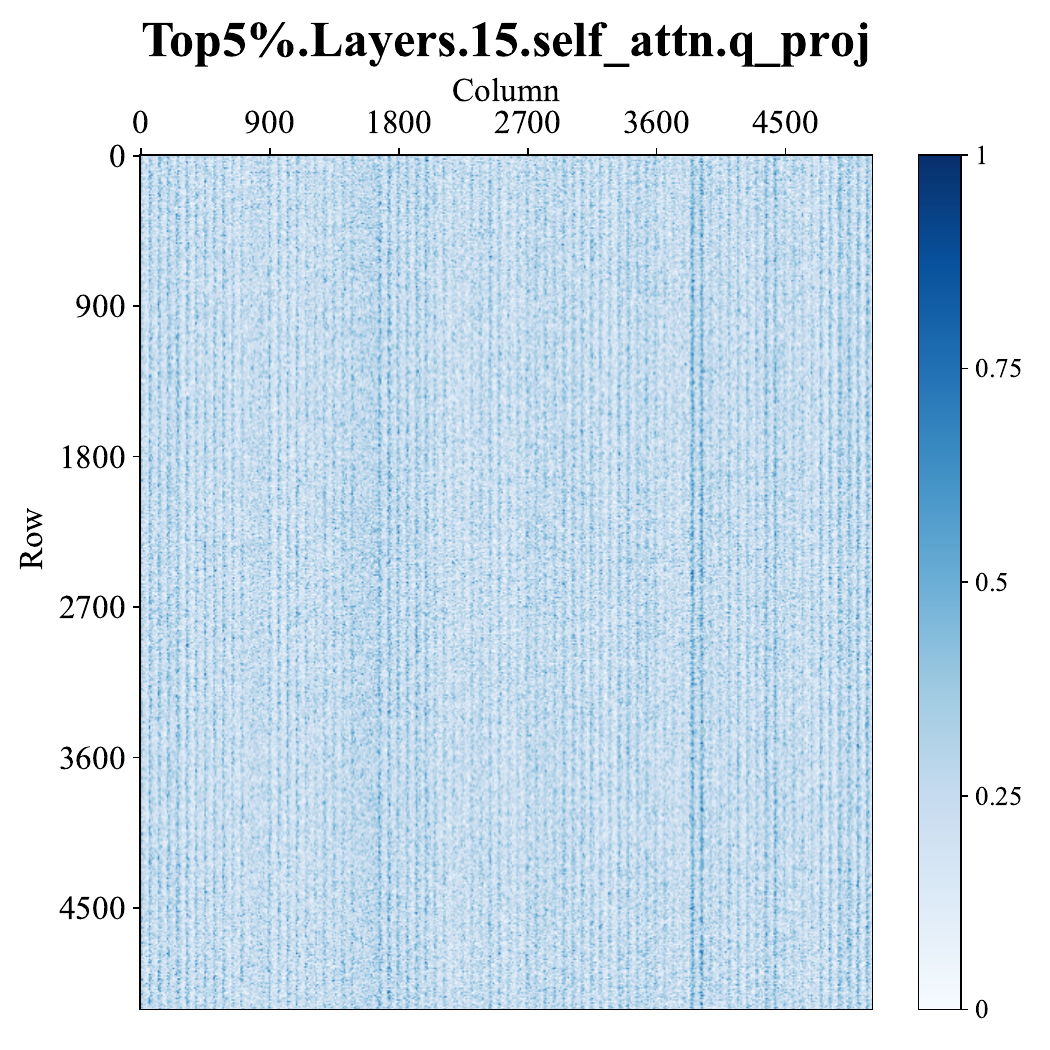}
		\includegraphics[width=3.6cm]{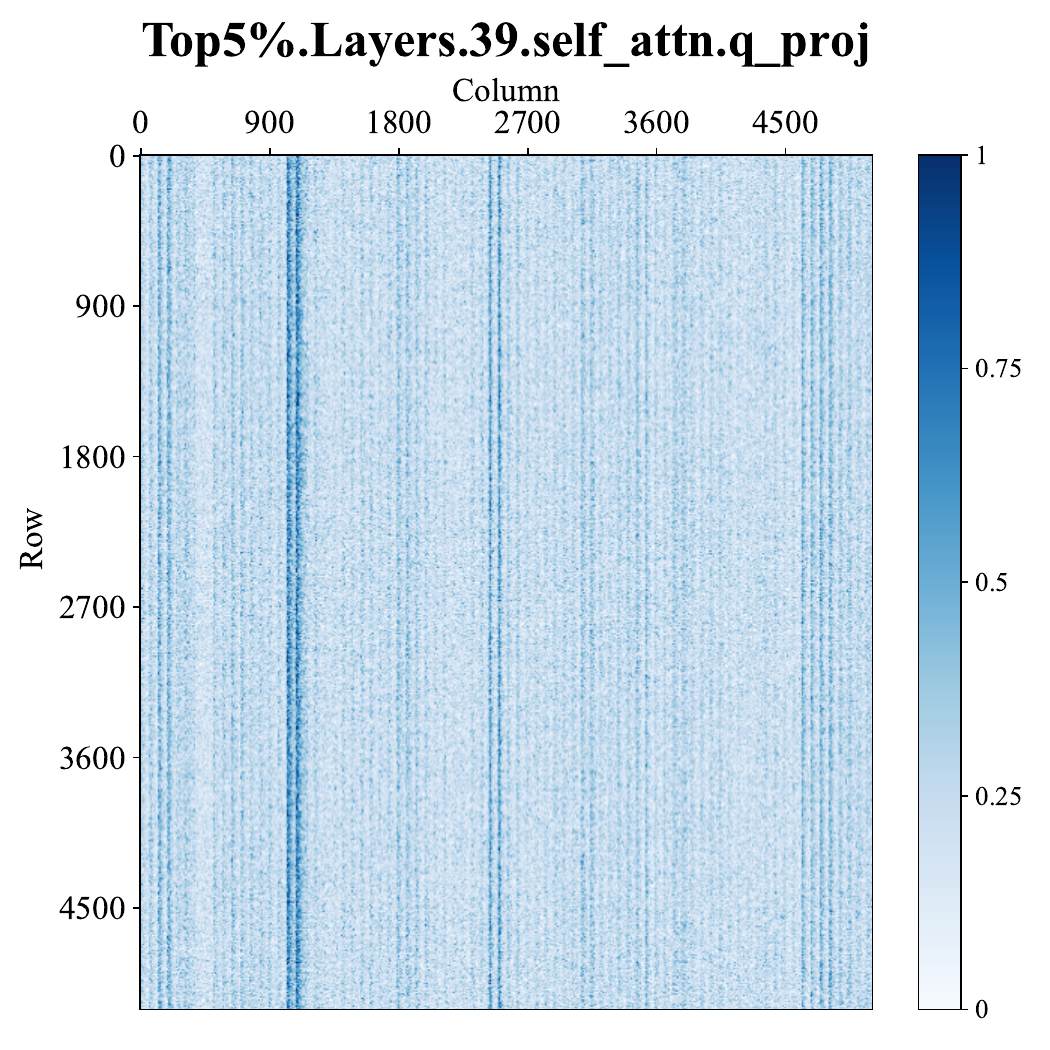}
	\end{minipage}
	}
 \caption{Visualization of Attn.q's `Top' region in LLaMA2-13b. The scale from 0 to 1 (after normalization) represent the proportion of parameters within a $3\times3$ vicinity that belong to the Bottom region.}
\label{fig:app_visualize_13b_q}
\end{figure*}

\begin{figure*}[t]
	\centering
	\subfigure{
		\centering
	\begin{minipage}[t]{0.23\textwidth}
        \includegraphics[width=3.6cm]{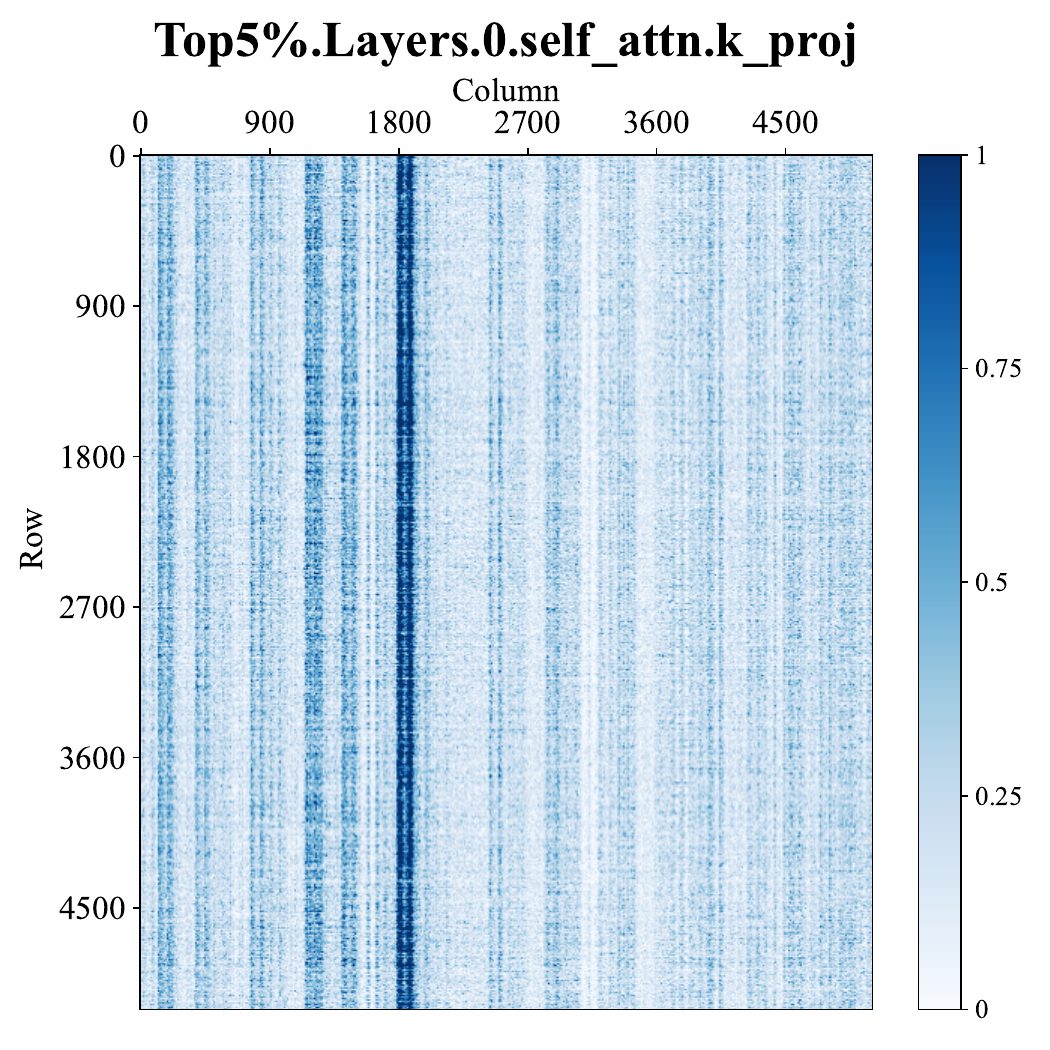}
        \includegraphics[width=3.6cm]{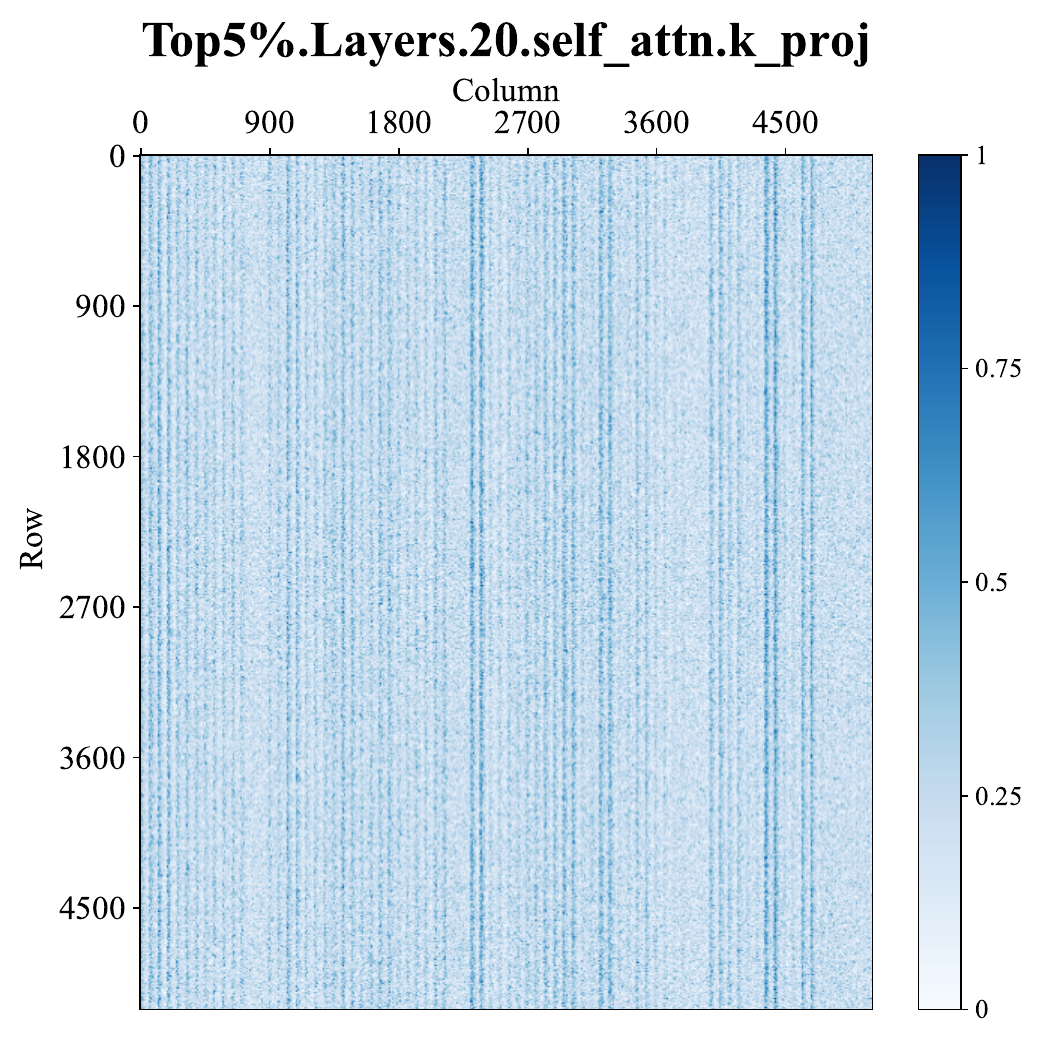}
	\end{minipage}
	}
	\subfigure{
		\centering
	\begin{minipage}[t]{0.23\textwidth}
	       \includegraphics[width=3.6cm]{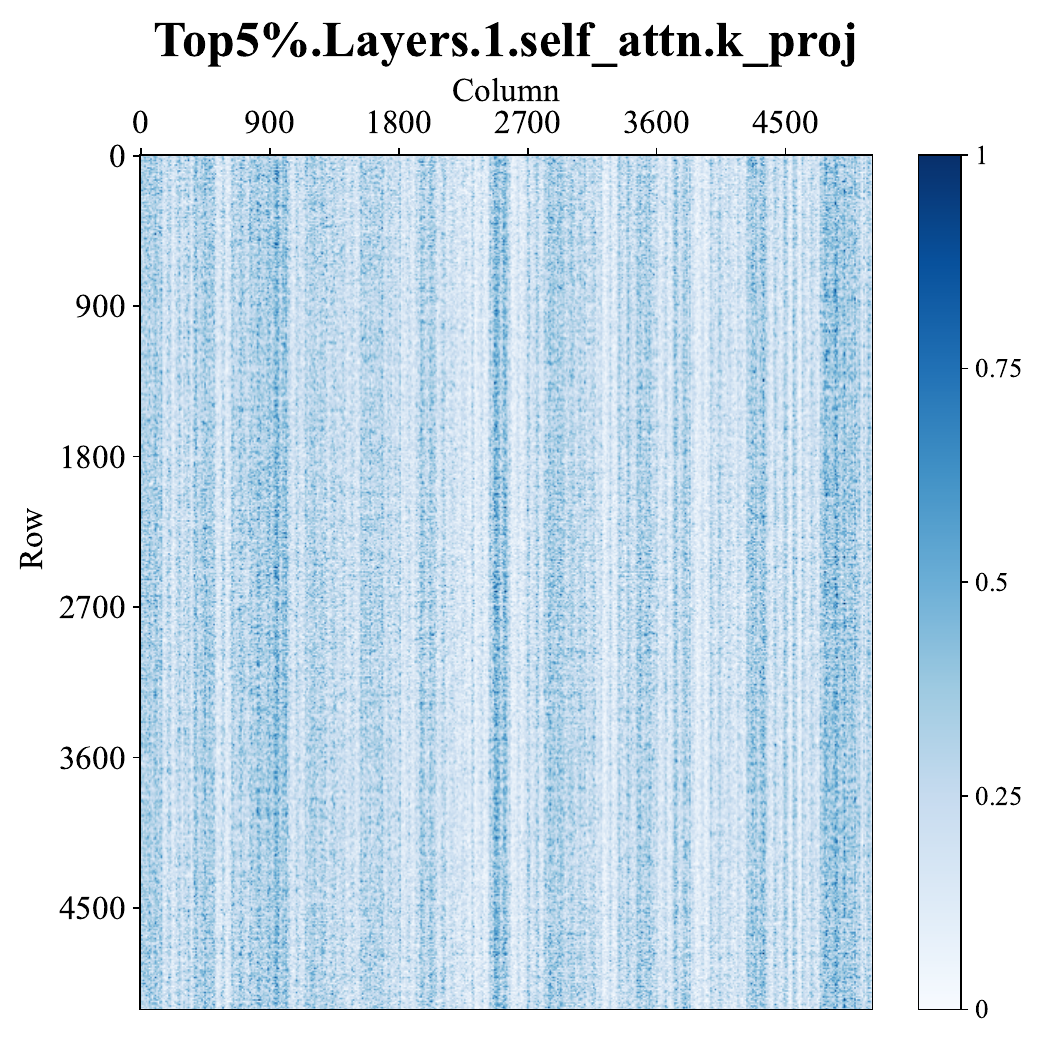}
            \includegraphics[width=3.6cm]{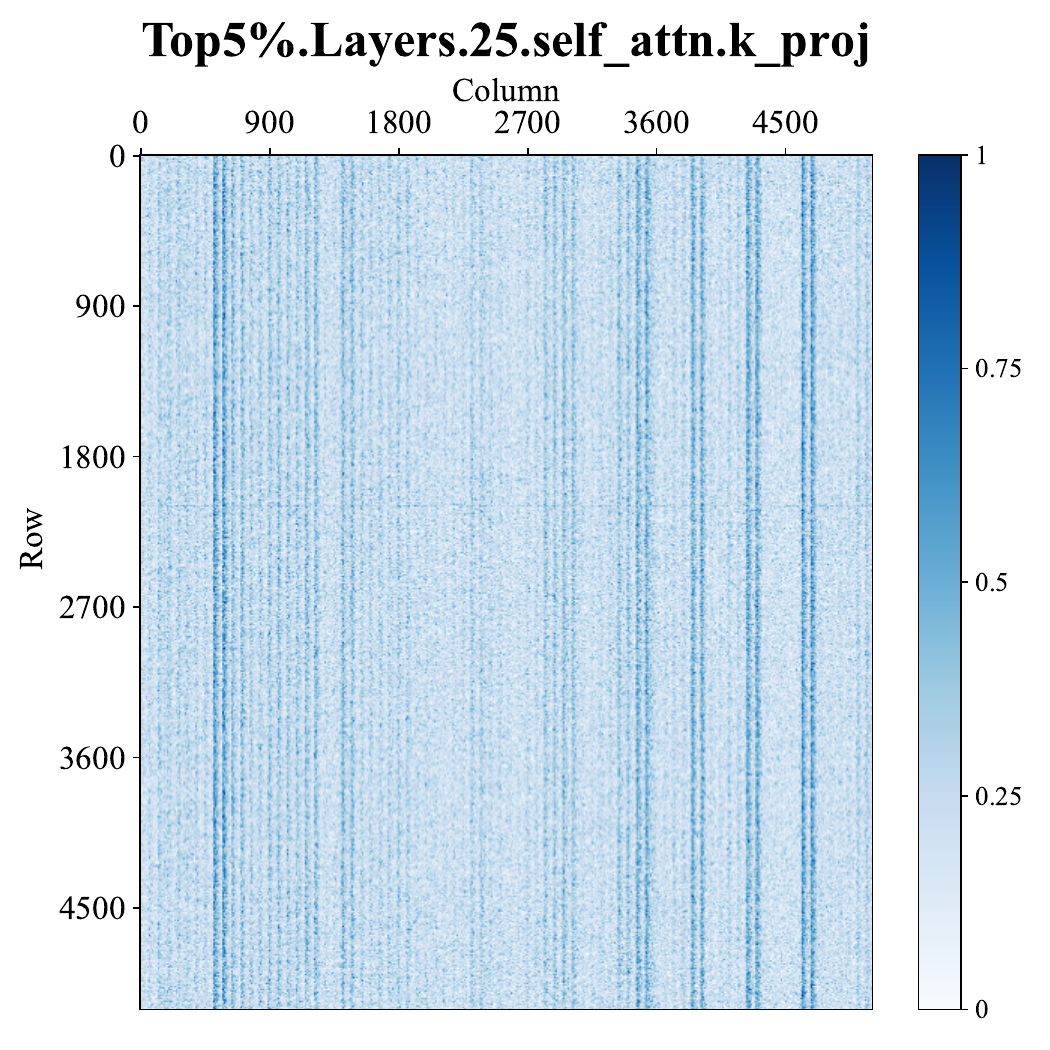}
	\end{minipage}
	}
	\subfigure{
		\centering
	\begin{minipage}[t]{0.23\textwidth}
		\includegraphics[width=3.6cm]{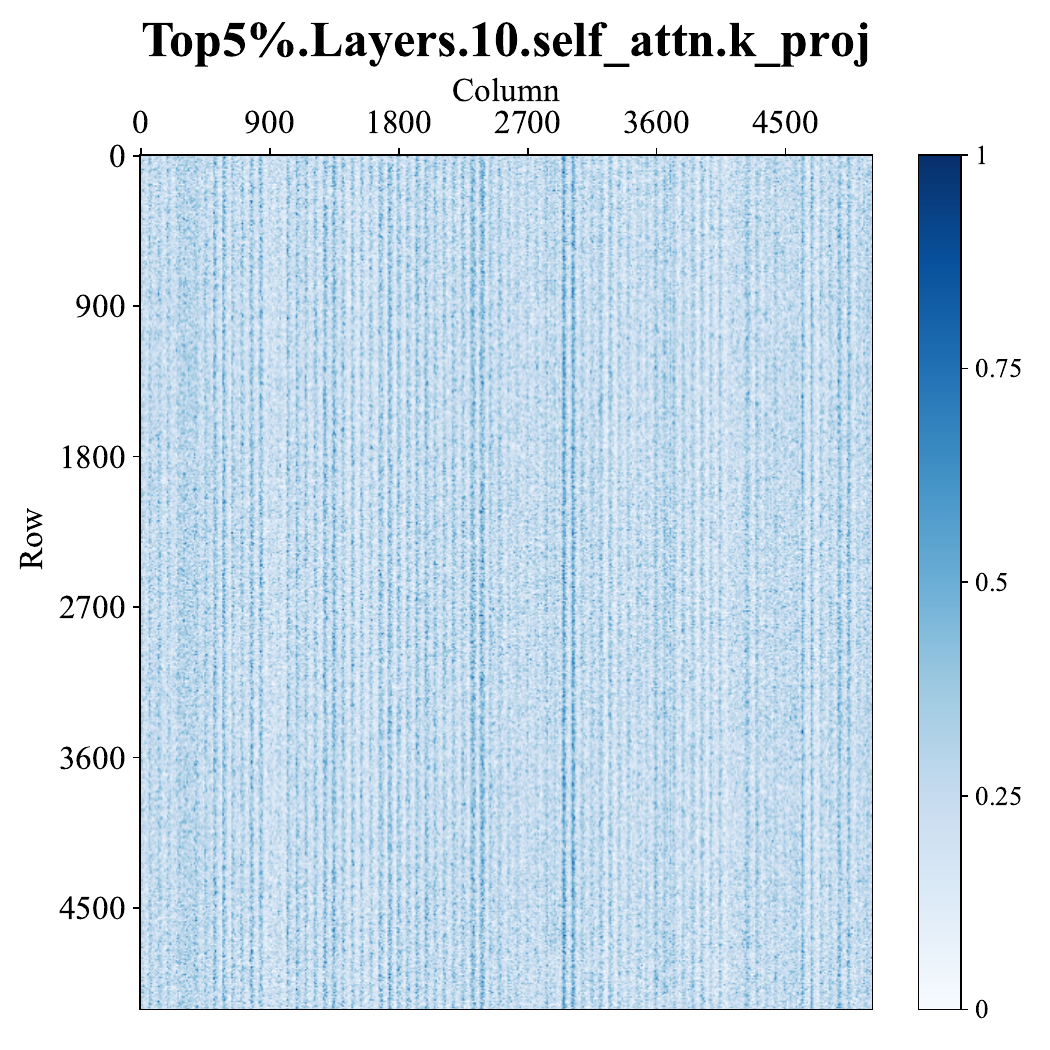}
		\includegraphics[width=3.6cm]{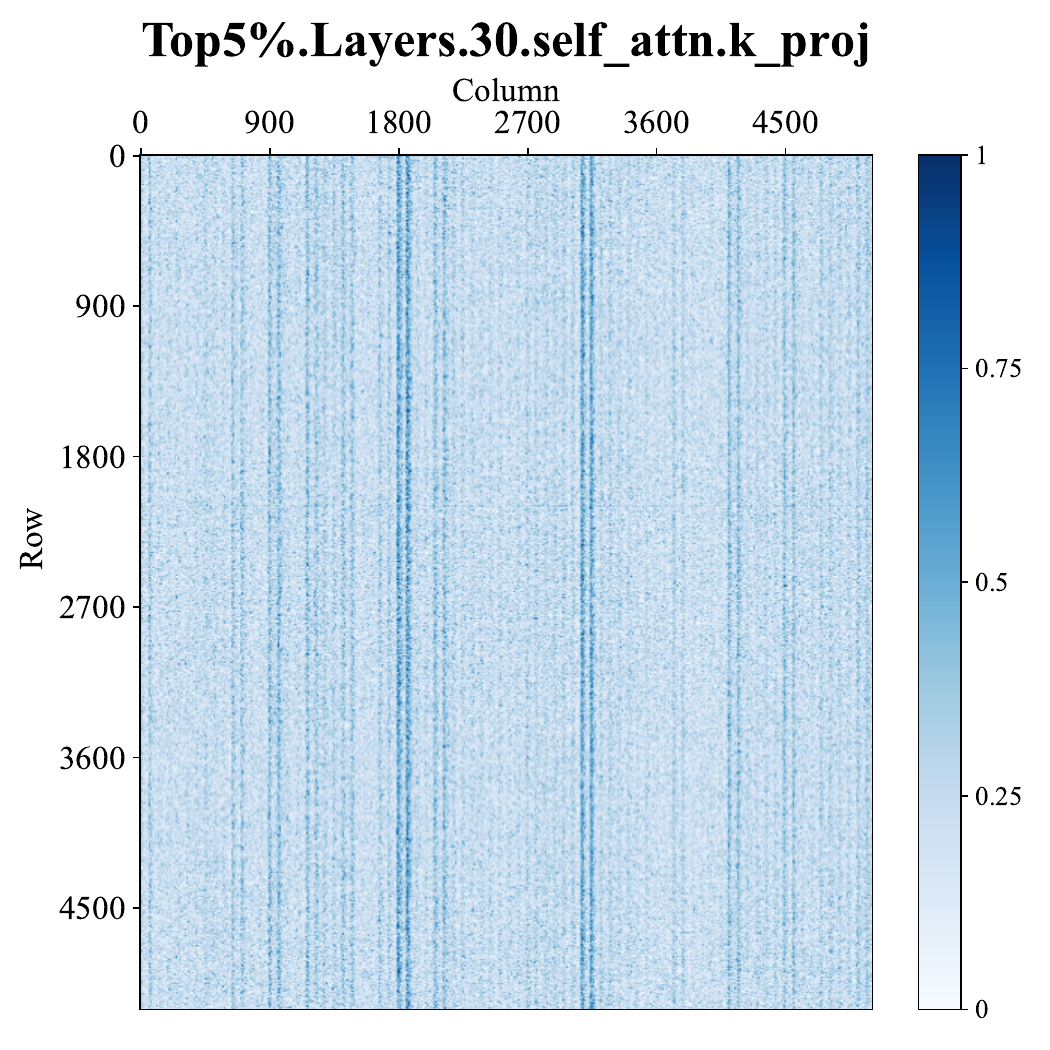}
	\end{minipage}
	}
 \subfigure{
		\centering
	\begin{minipage}[t]{0.23\textwidth}
		\includegraphics[width=3.6cm]{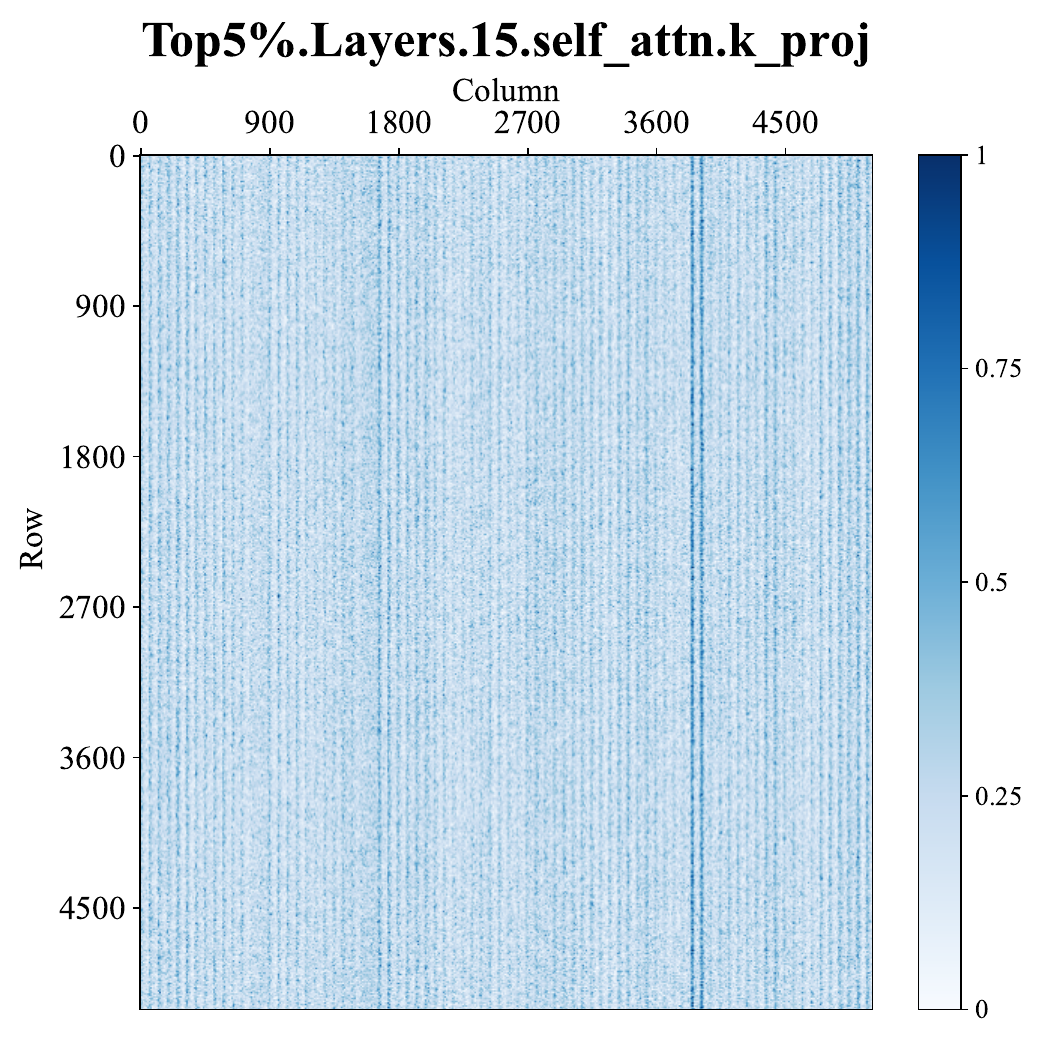}
		\includegraphics[width=3.6cm]{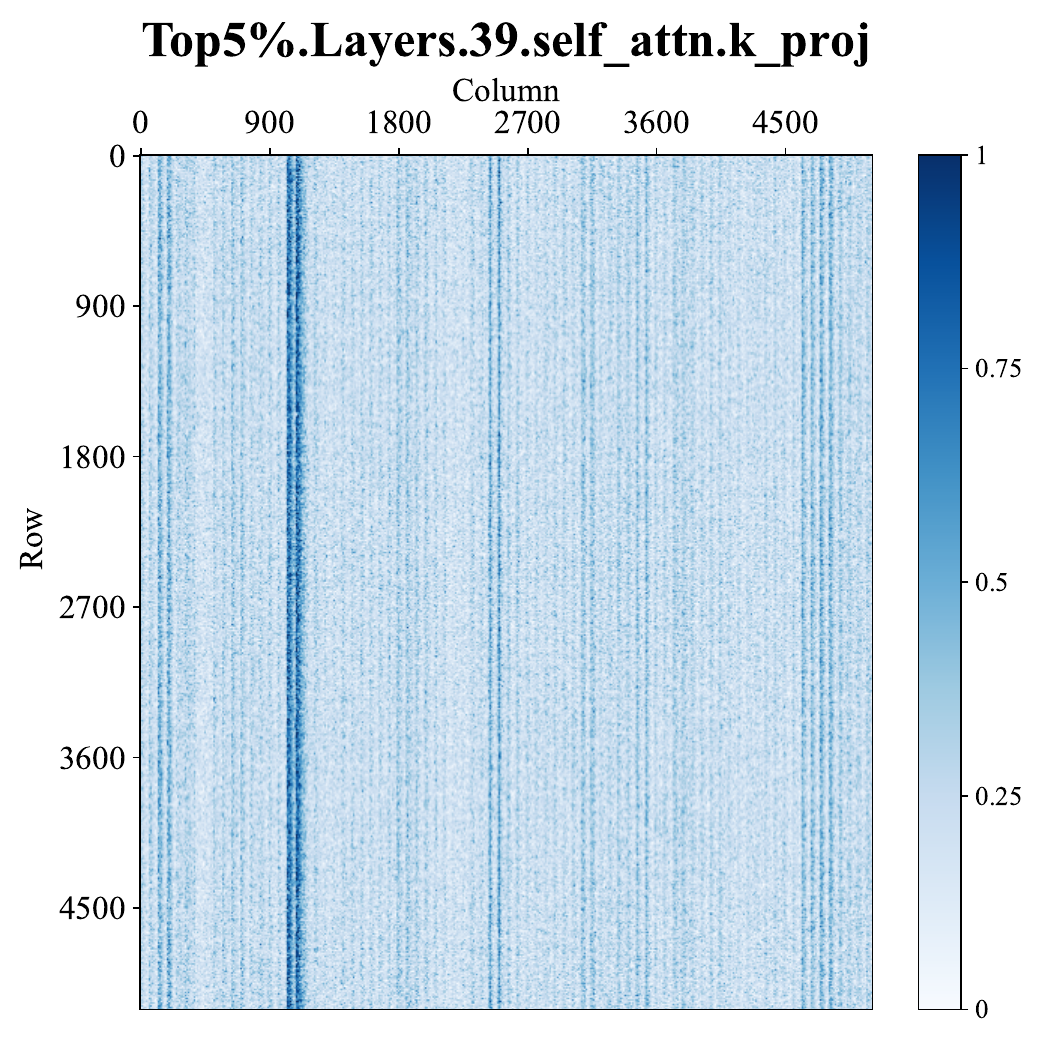}
	\end{minipage}
	}
 \caption{Visualization of Attn.k's `Top' region in LLaMA2-13b. The scale from 0 to 1 (after normalization) represent the proportion of parameters within a $3\times3$ vicinity that belong to the Bottom region.}
\label{fig:app_visualize_13b_k}
\end{figure*}

\begin{figure*}[t]
	\centering
	\subfigure{
		\centering
	\begin{minipage}[t]{0.23\textwidth}
        \includegraphics[width=3.6cm]{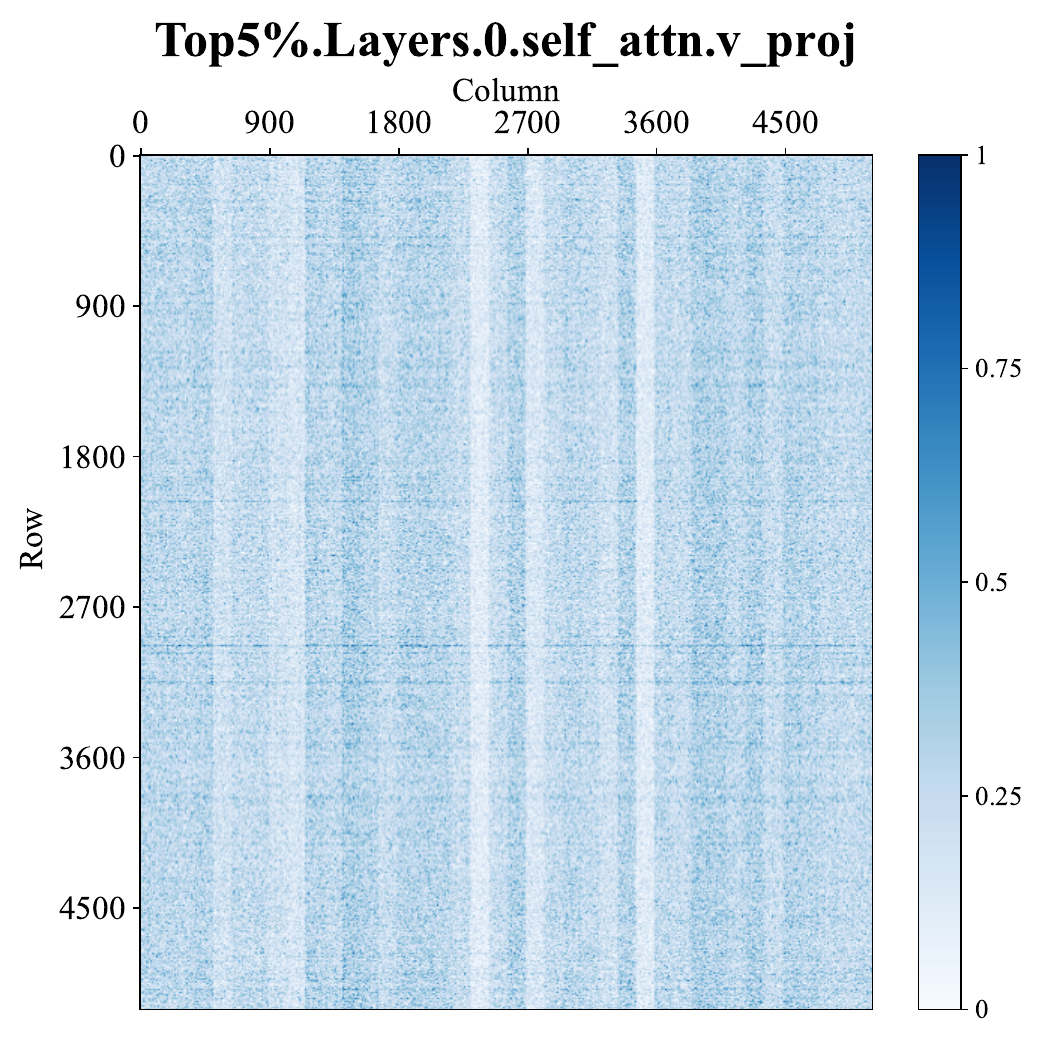}
        \includegraphics[width=3.6cm]{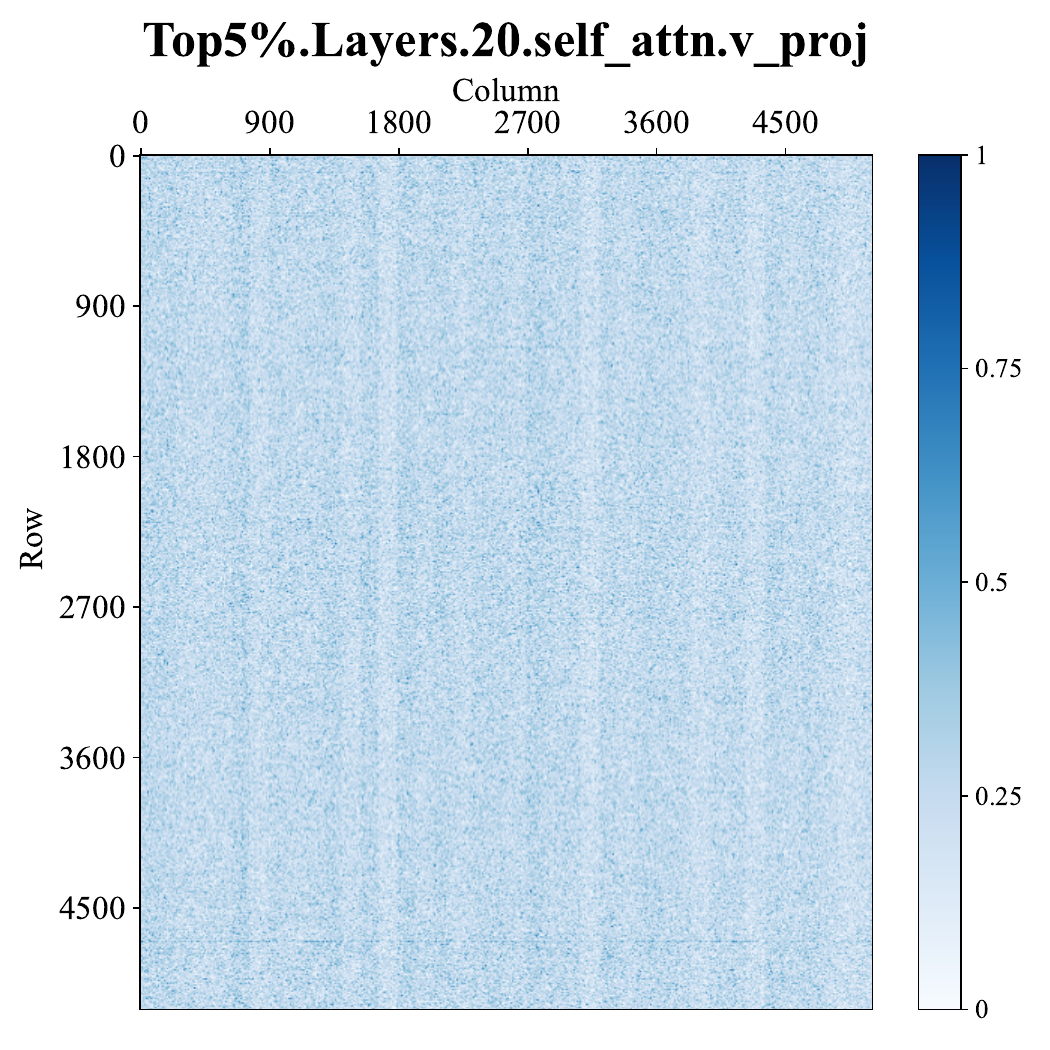}
	\end{minipage}
	}
	\subfigure{
		\centering
	\begin{minipage}[t]{0.23\textwidth}
	       \includegraphics[width=3.6cm]{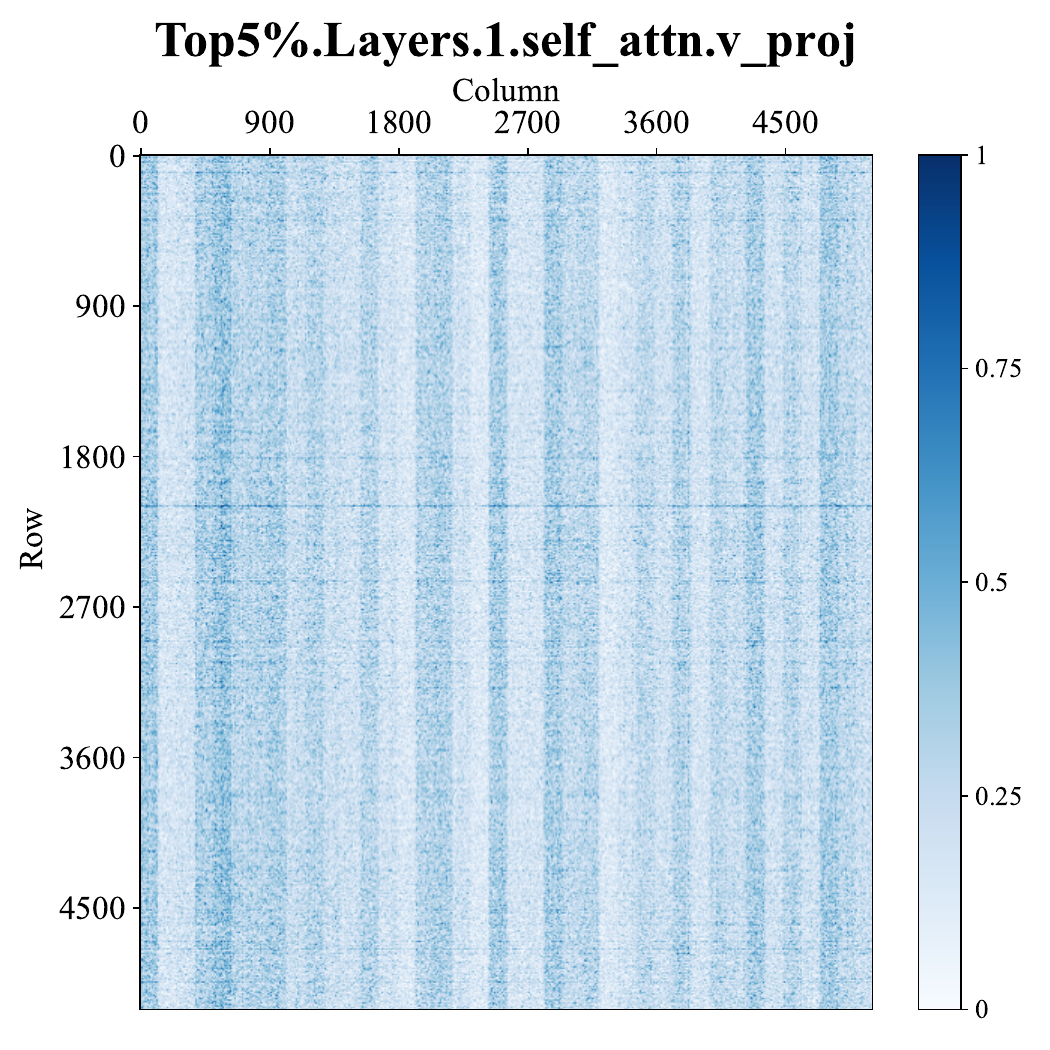}
            \includegraphics[width=3.6cm]{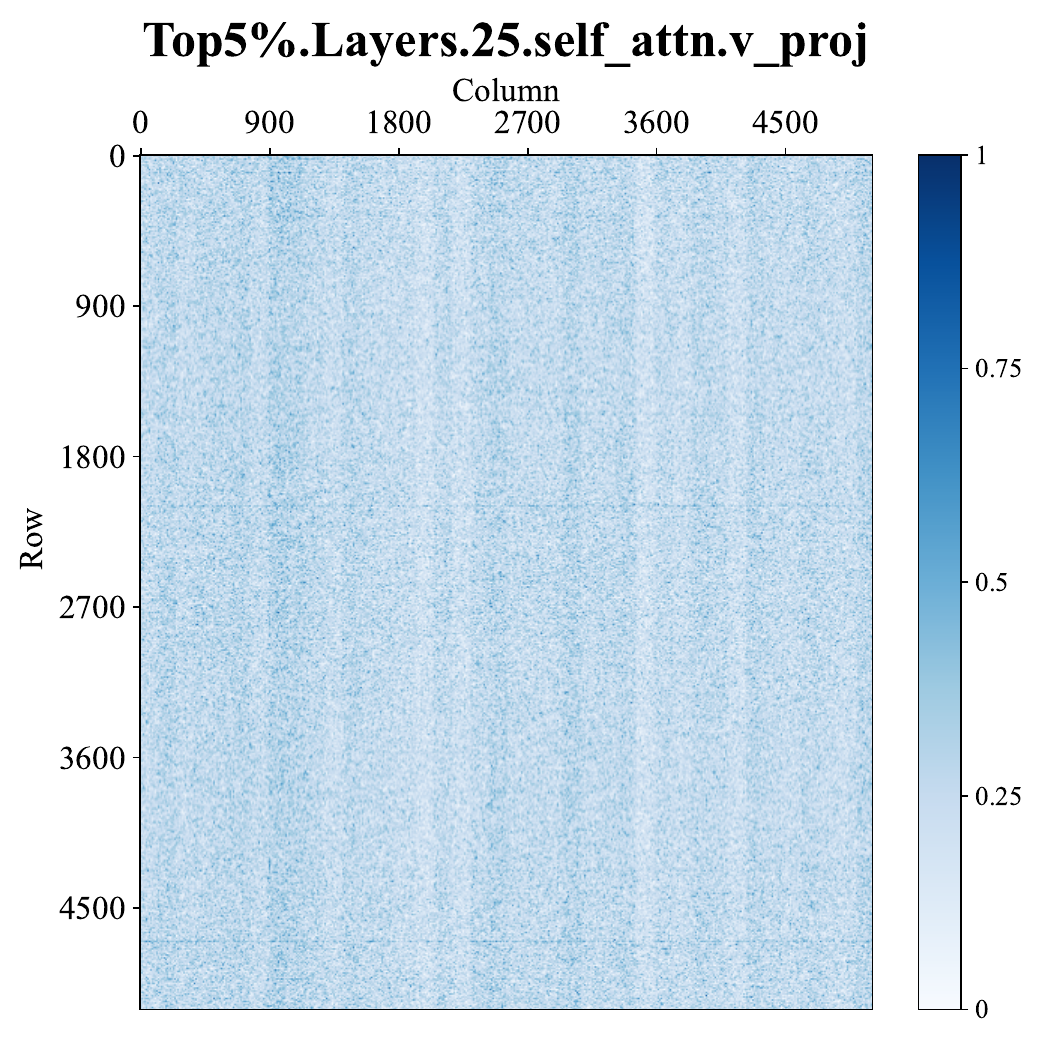}
	\end{minipage}
	}
	\subfigure{
		\centering
	\begin{minipage}[t]{0.23\textwidth}
		\includegraphics[width=3.6cm]{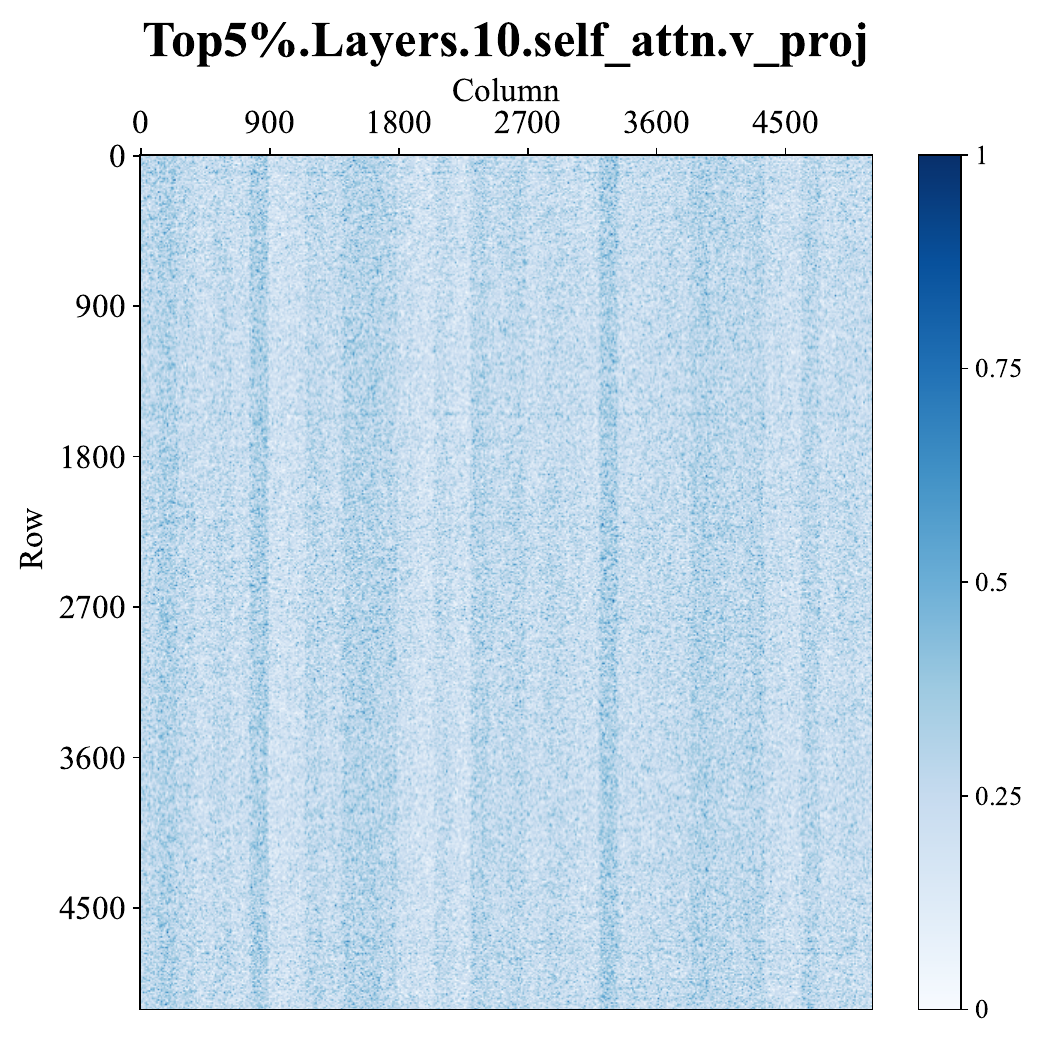}
		\includegraphics[width=3.6cm]{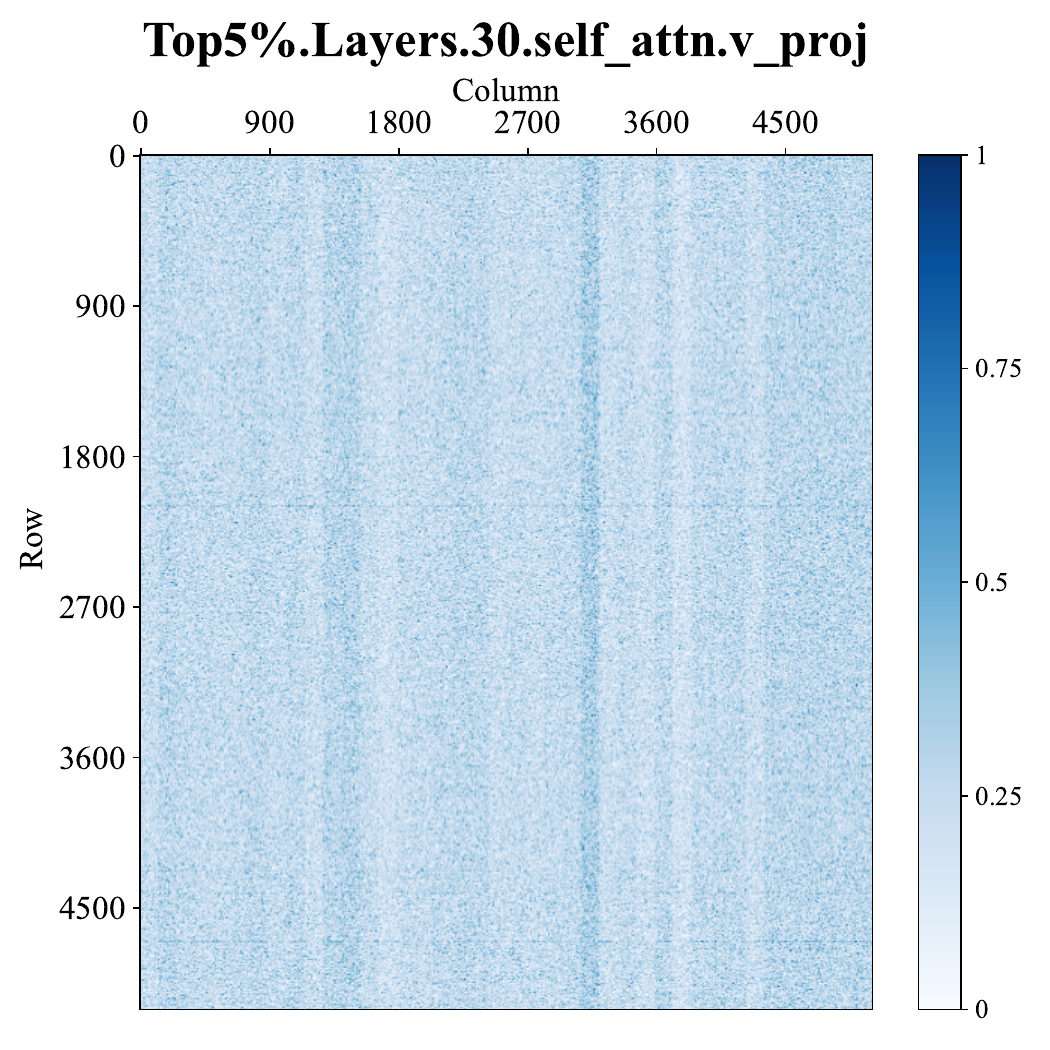}
	\end{minipage}
	}
 \subfigure{
		\centering
	\begin{minipage}[t]{0.23\textwidth}
		\includegraphics[width=3.6cm]{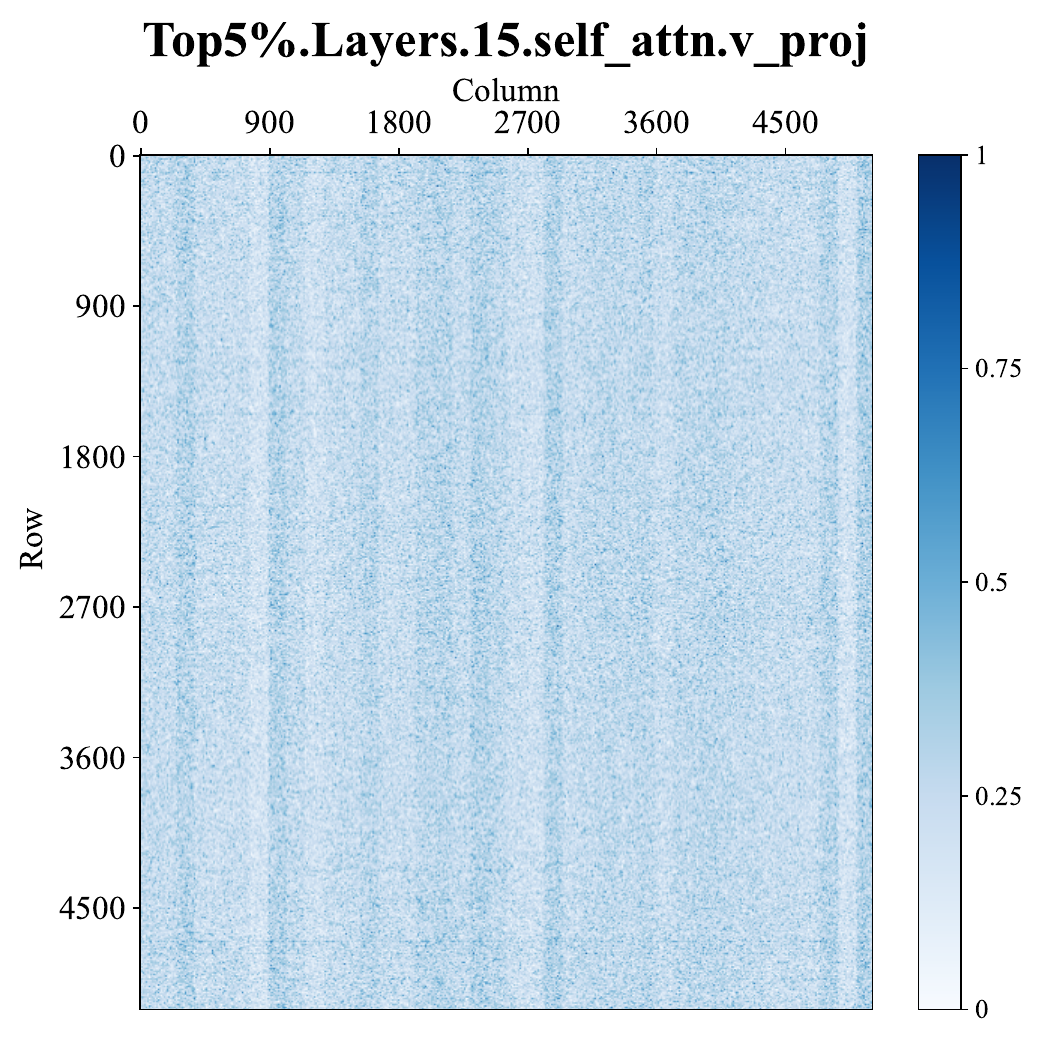}
		\includegraphics[width=3.6cm]{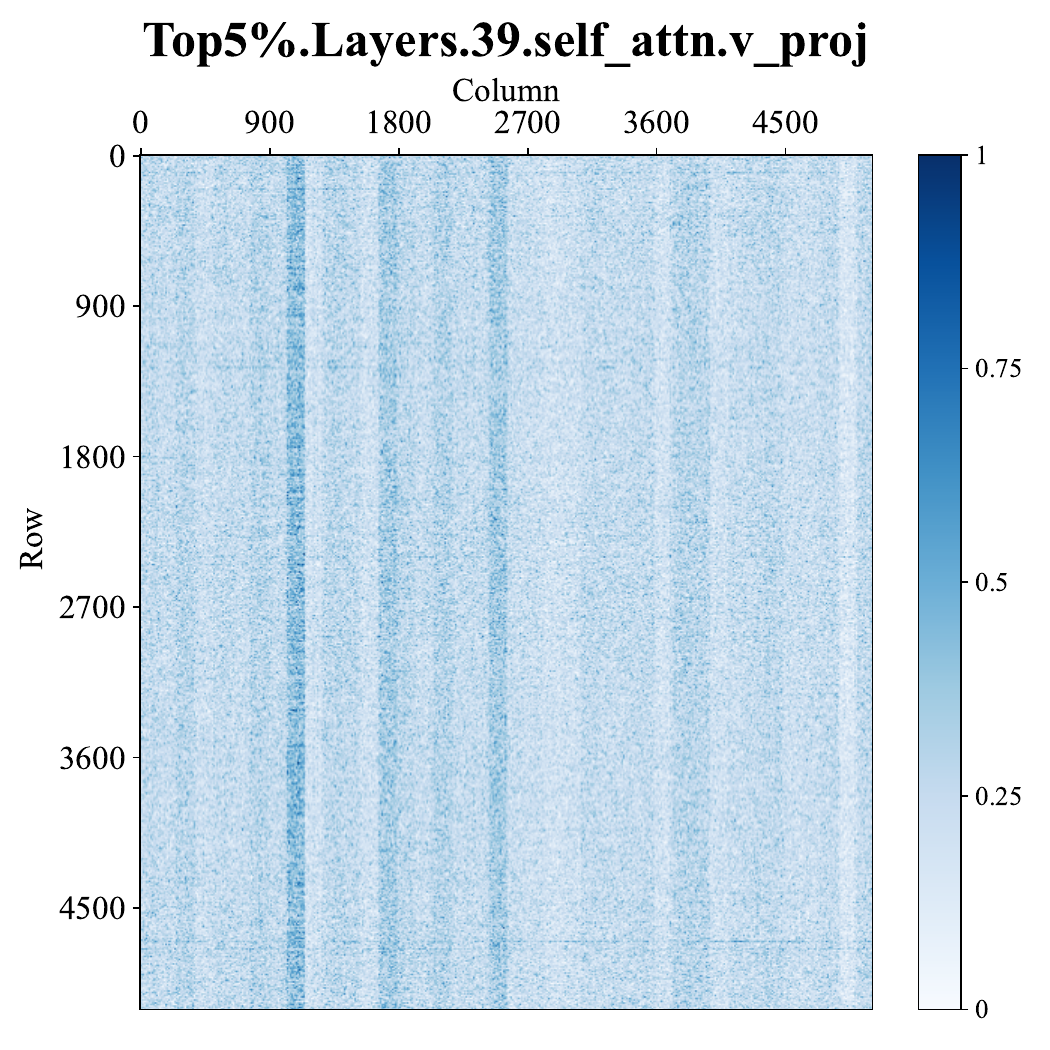}
	\end{minipage}
	}
 \caption{Visualization of Attn.v's `Top' region in LLaMA2-13b. The scale from 0 to 1 (after normalization) represent the proportion of parameters within a $3\times3$ vicinity that belong to the Bottom region.}
\label{fig:app_visualize_13b_v}
\end{figure*}

\begin{figure*}[t]
	\centering
	\subfigure{
		\centering
	\begin{minipage}[t]{0.23\textwidth}
        \includegraphics[width=3.6cm]{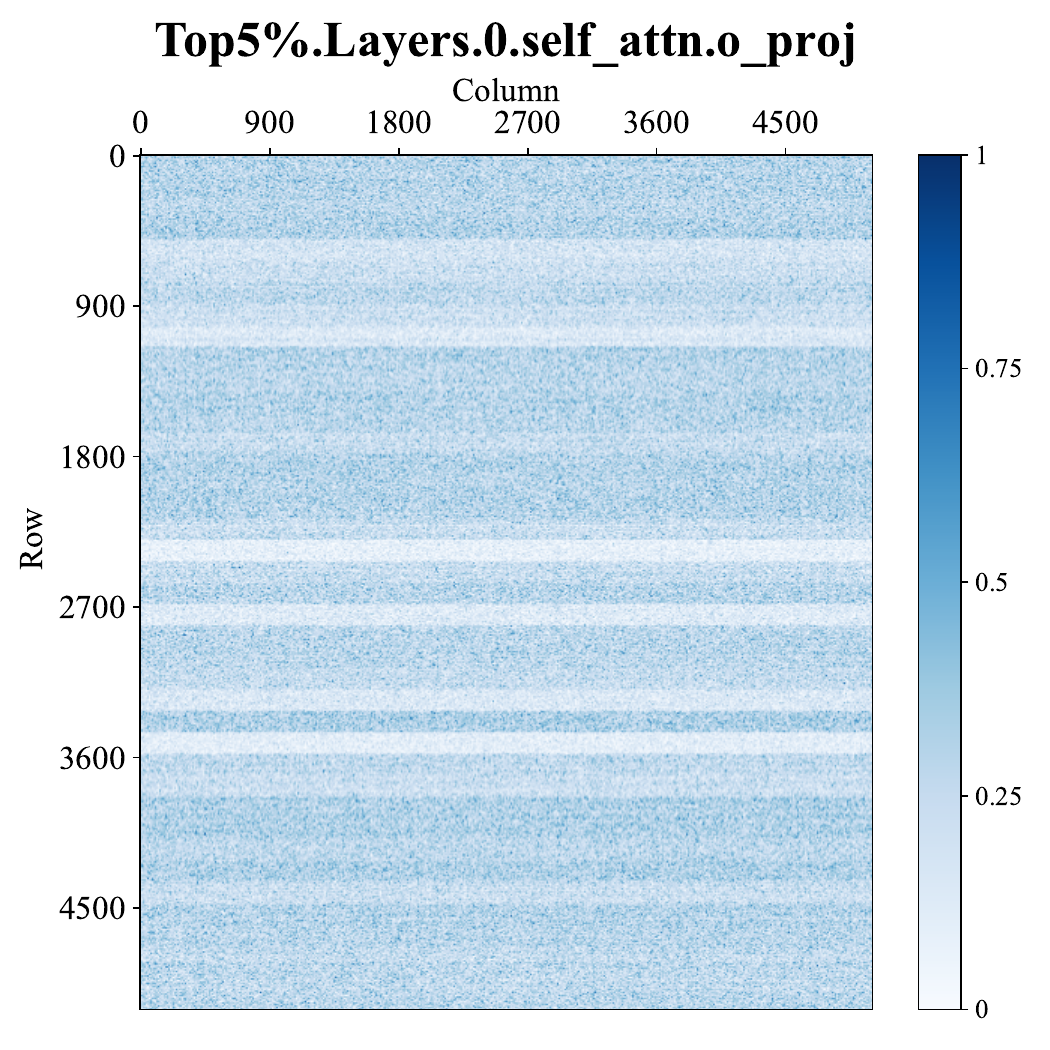}
        \includegraphics[width=3.6cm]{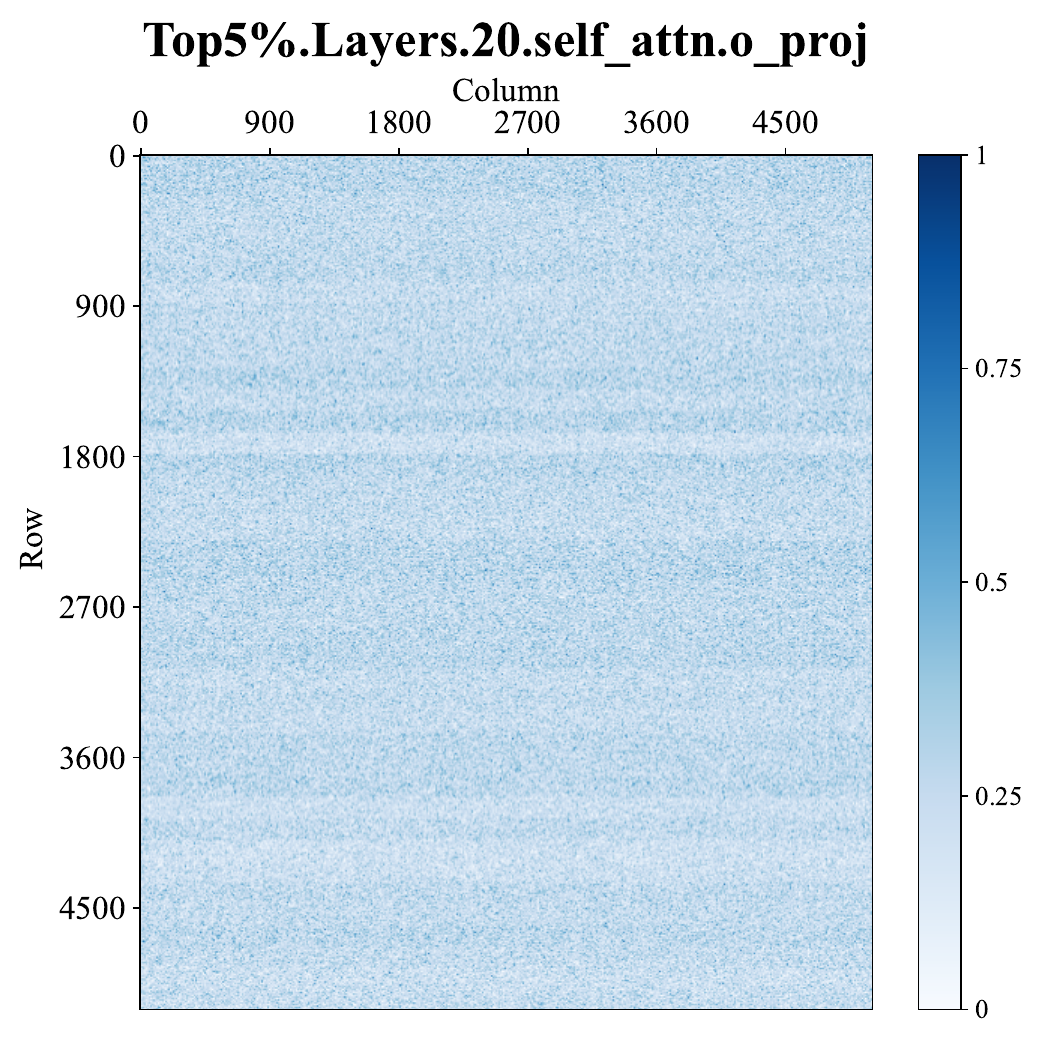}
	\end{minipage}
	}
	\subfigure{
		\centering
	\begin{minipage}[t]{0.23\textwidth}
	       \includegraphics[width=3.6cm]{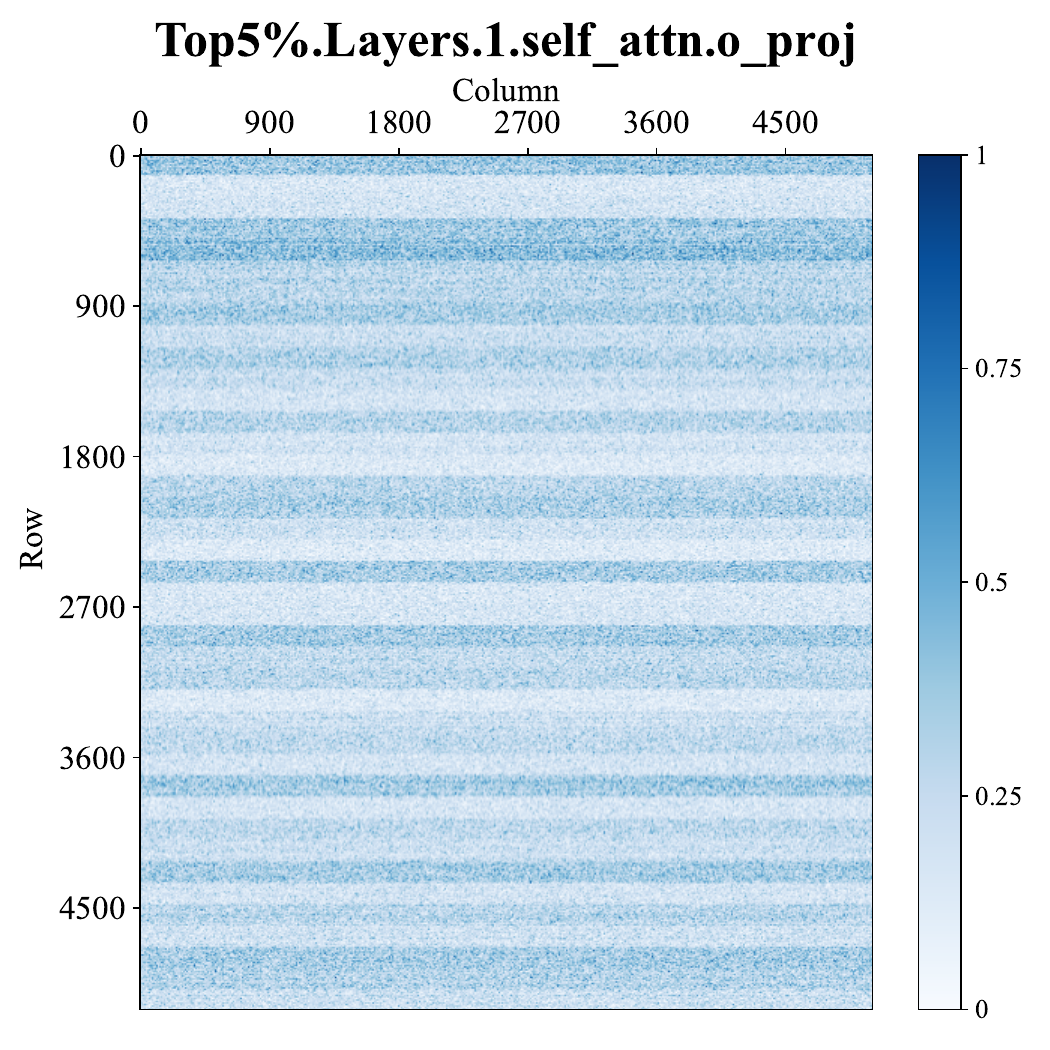}
            \includegraphics[width=3.6cm]{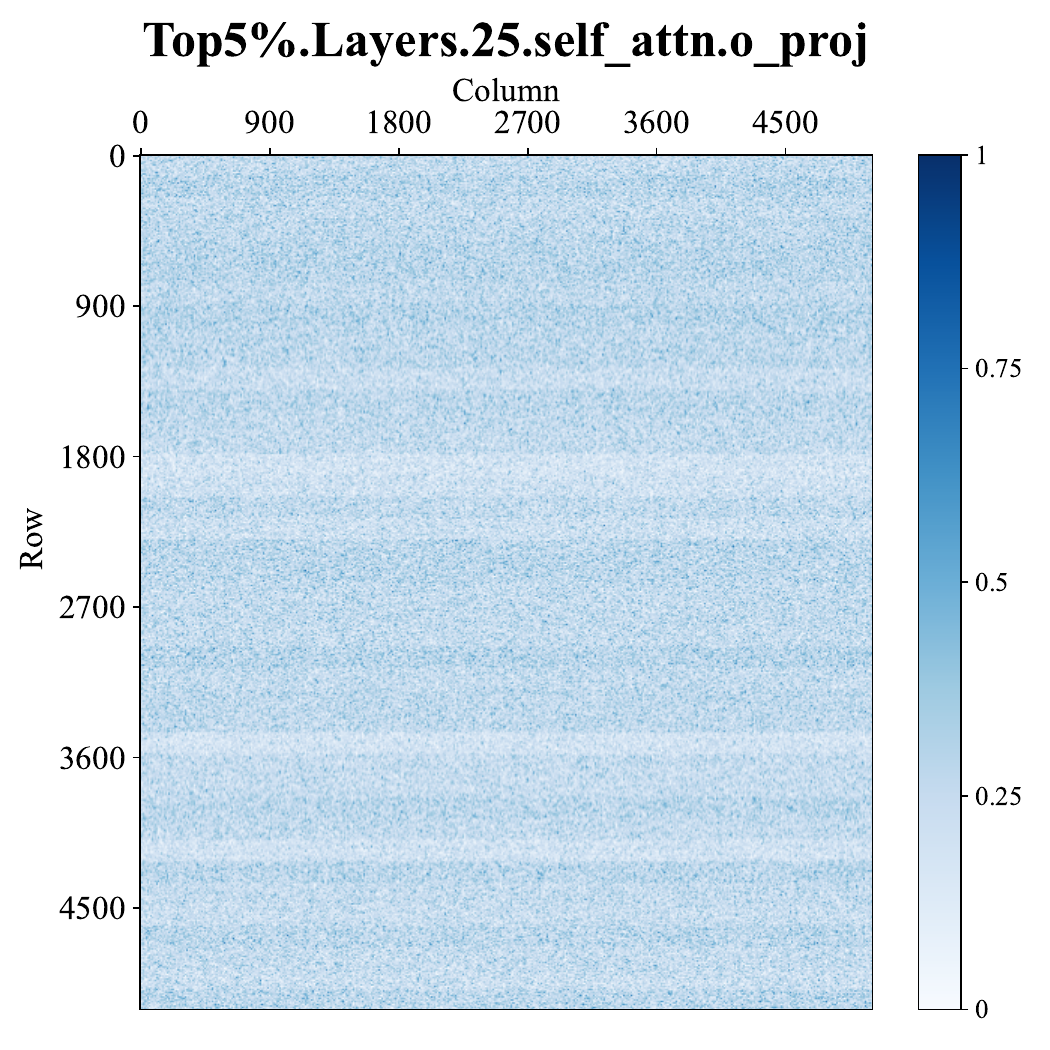}
	\end{minipage}
	}
	\subfigure{
		\centering
	\begin{minipage}[t]{0.23\textwidth}
		\includegraphics[width=3.6cm]{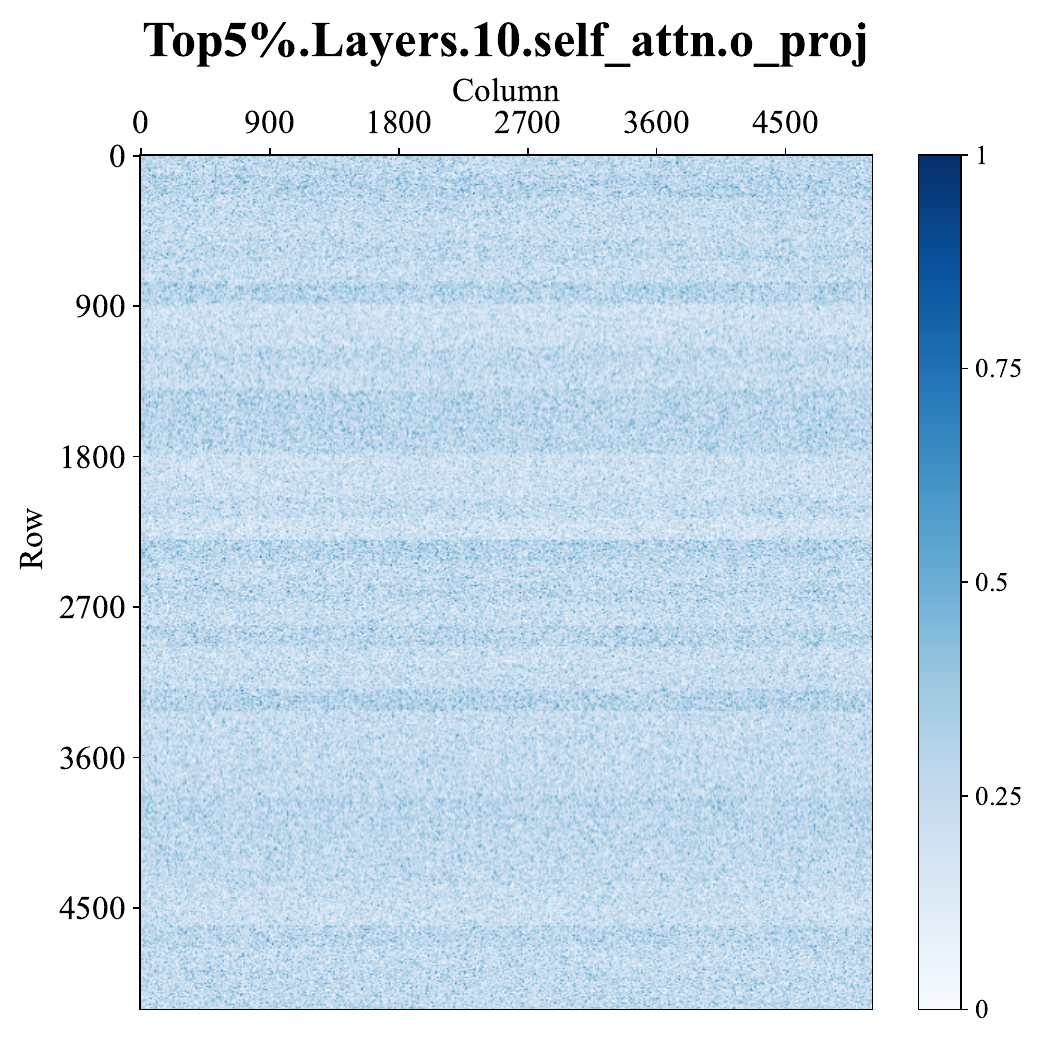}
		\includegraphics[width=3.6cm]{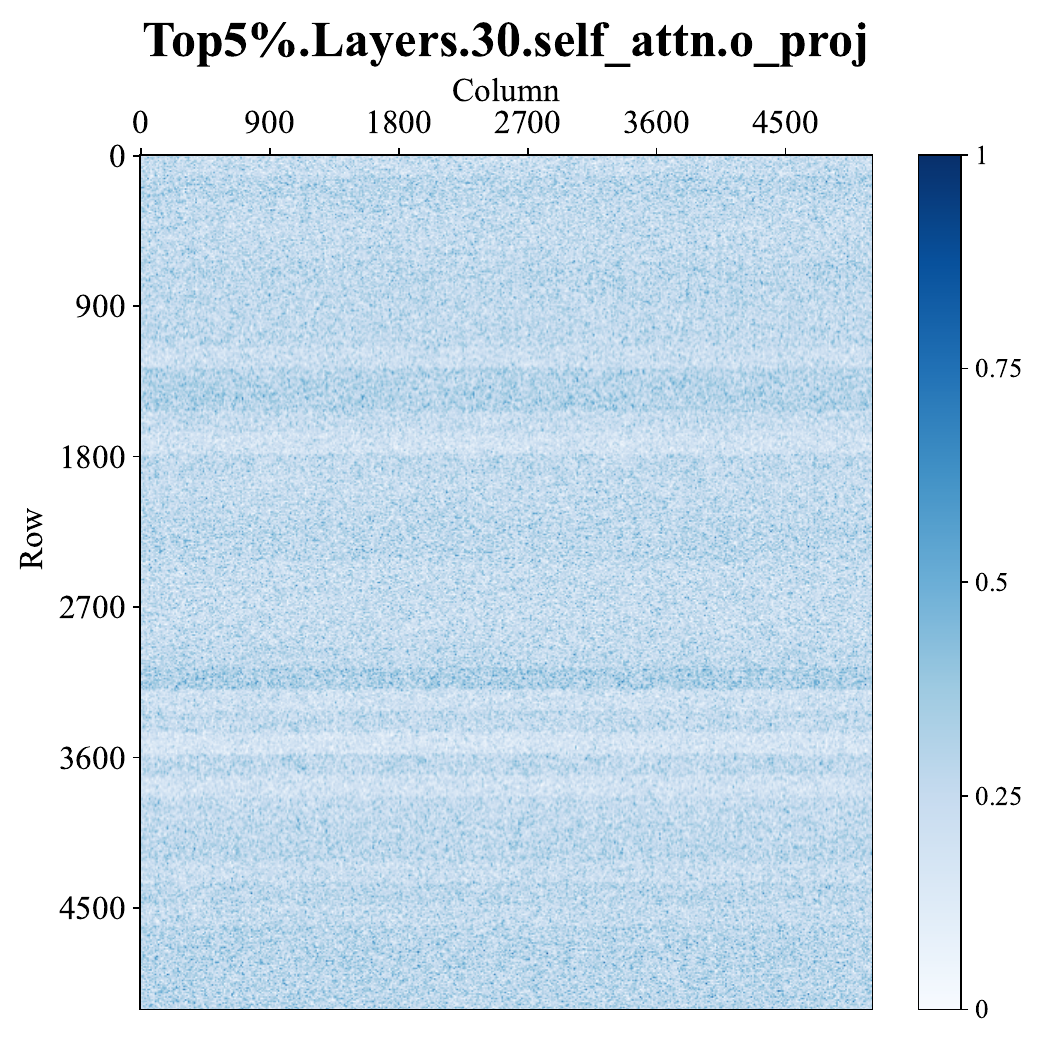}
	\end{minipage}
	}
 \subfigure{
		\centering
	\begin{minipage}[t]{0.23\textwidth}
		\includegraphics[width=3.6cm]{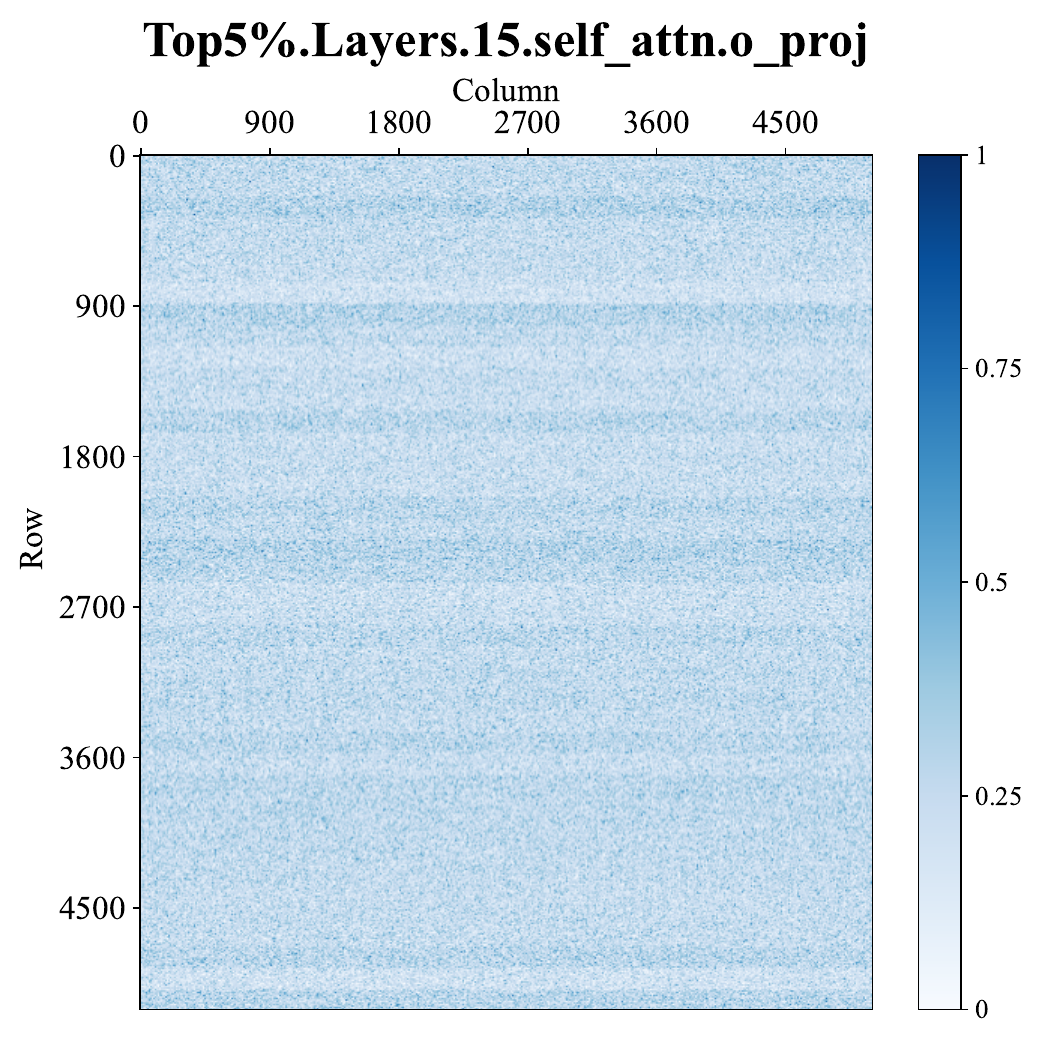}
		\includegraphics[width=3.6cm]{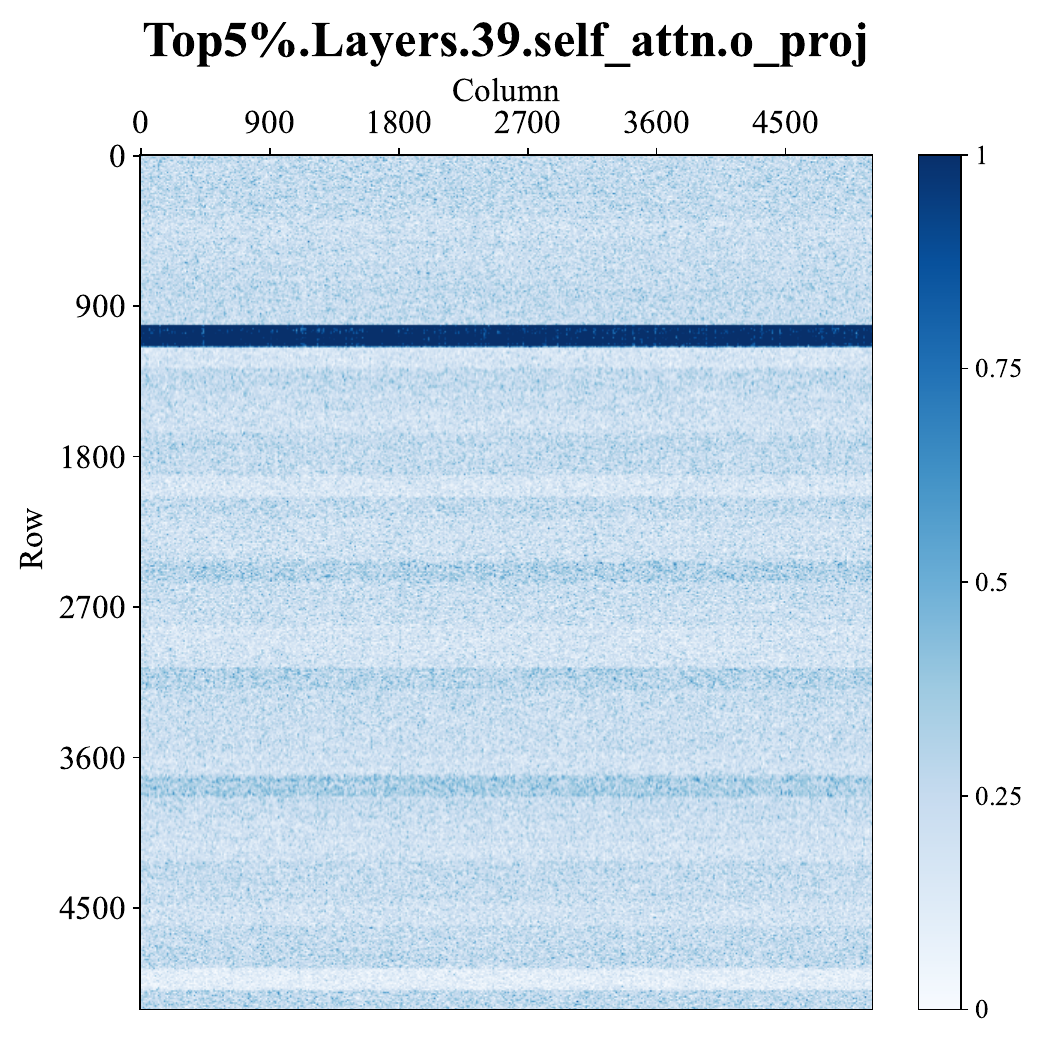}
	\end{minipage}
	}
 \caption{Visualization of Attn.o's `Top' region in LLaMA2-13b. The scale from 0 to 1 (after normalization) represent the proportion of parameters within a $3\times3$ vicinity that belong to the Bottom region.}
\label{fig:app_visualize_13b_o}
\end{figure*}

\begin{figure*}[t]
	\centering
	\subfigure{
		\centering
	\begin{minipage}[t]{0.23\textwidth}
        \includegraphics[width=3.6cm]{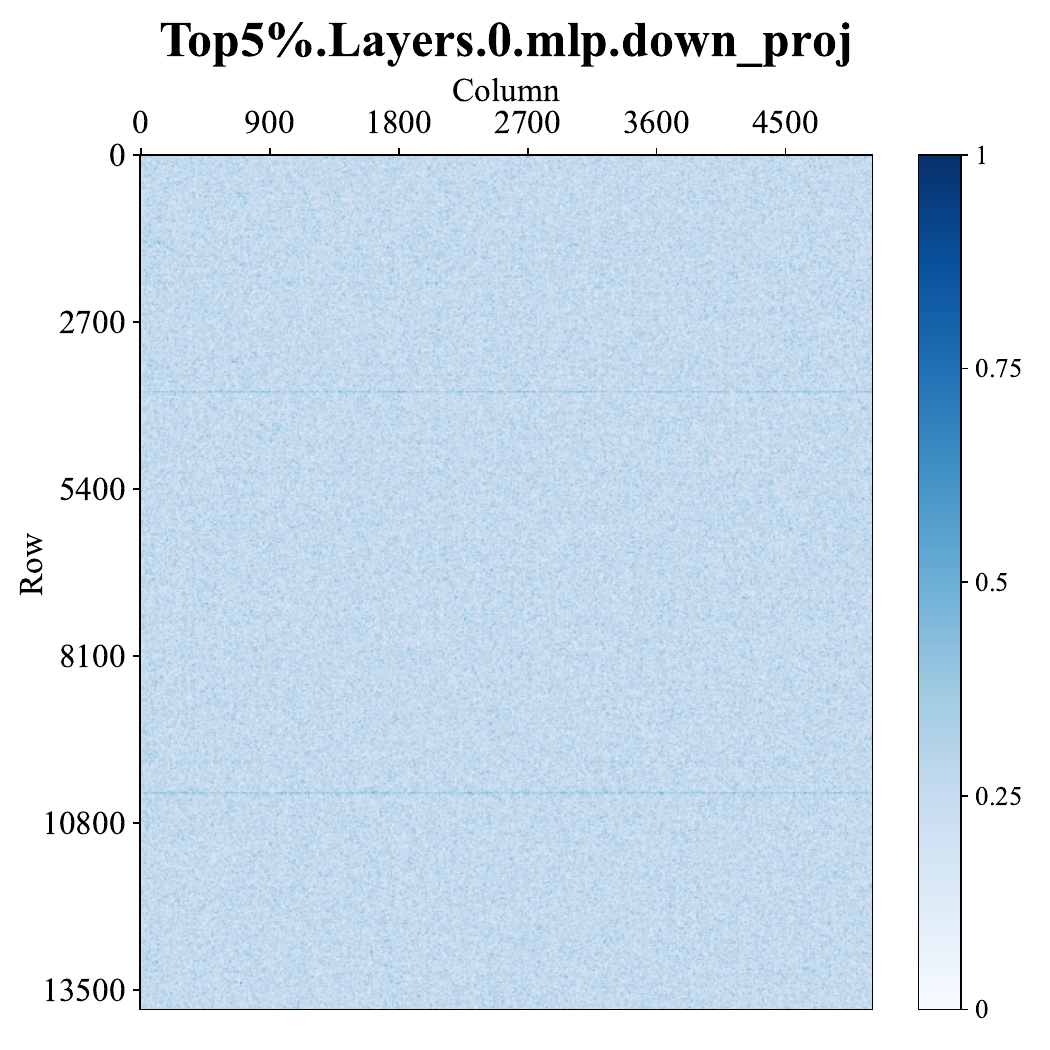}
        \includegraphics[width=3.6cm]{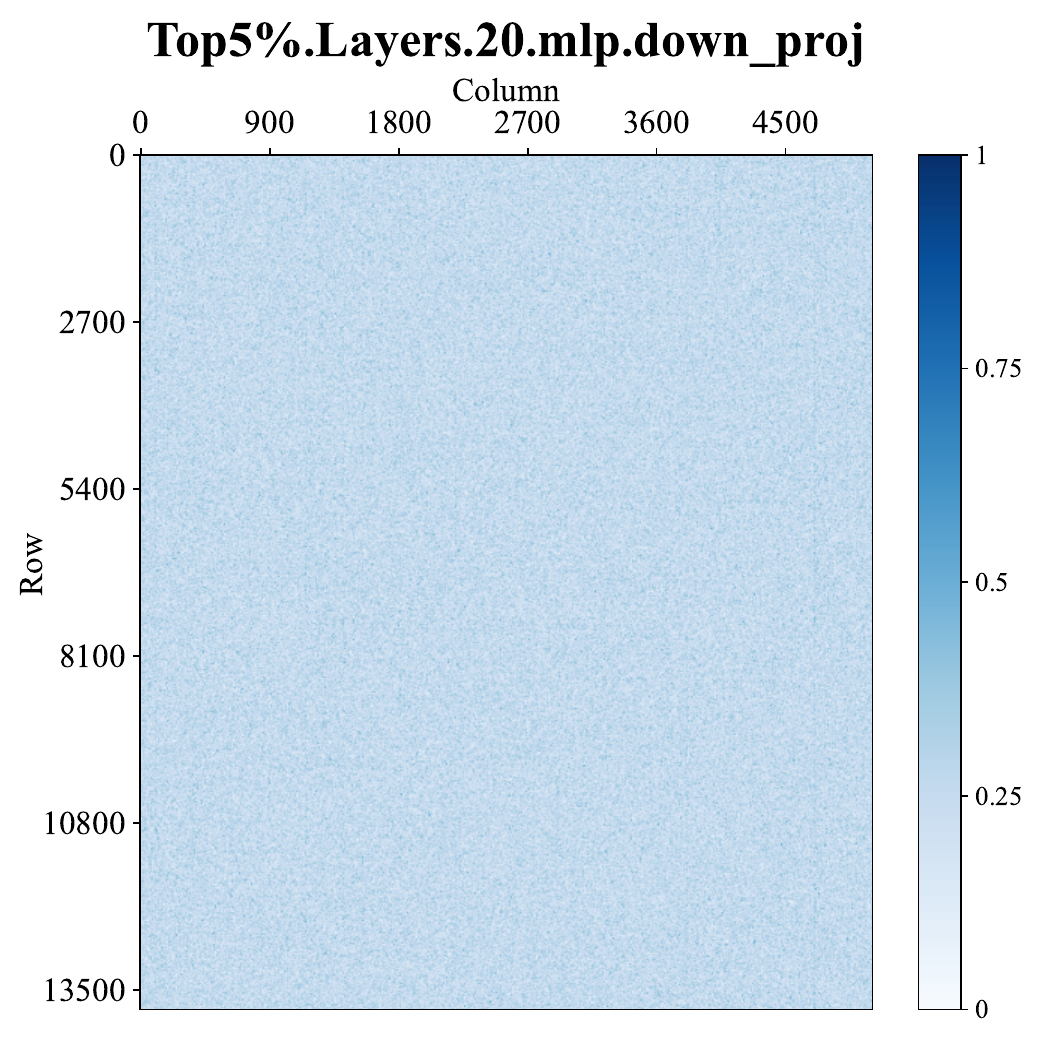}
	\end{minipage}
	}
	\subfigure{
		\centering
	\begin{minipage}[t]{0.23\textwidth}
	       \includegraphics[width=3.6cm]{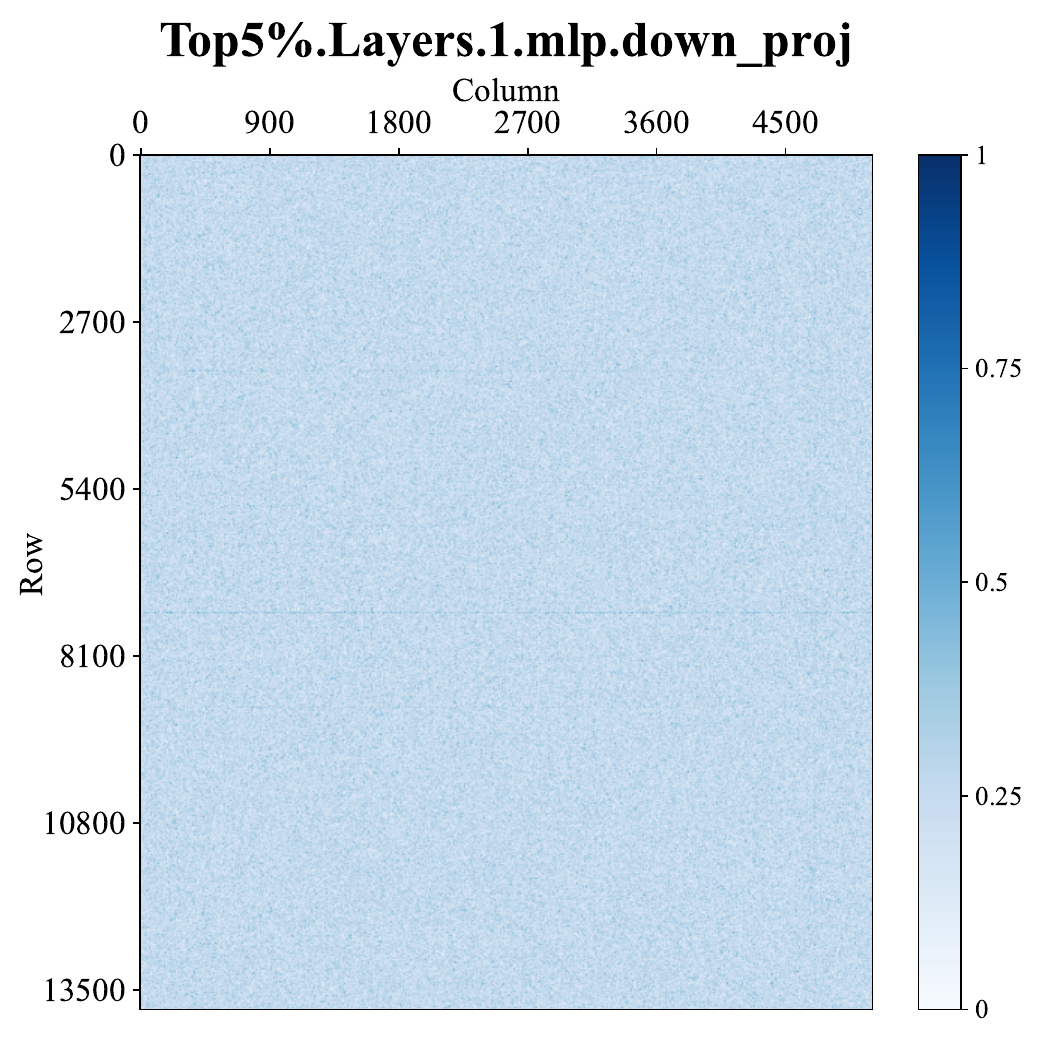}
            \includegraphics[width=3.6cm]{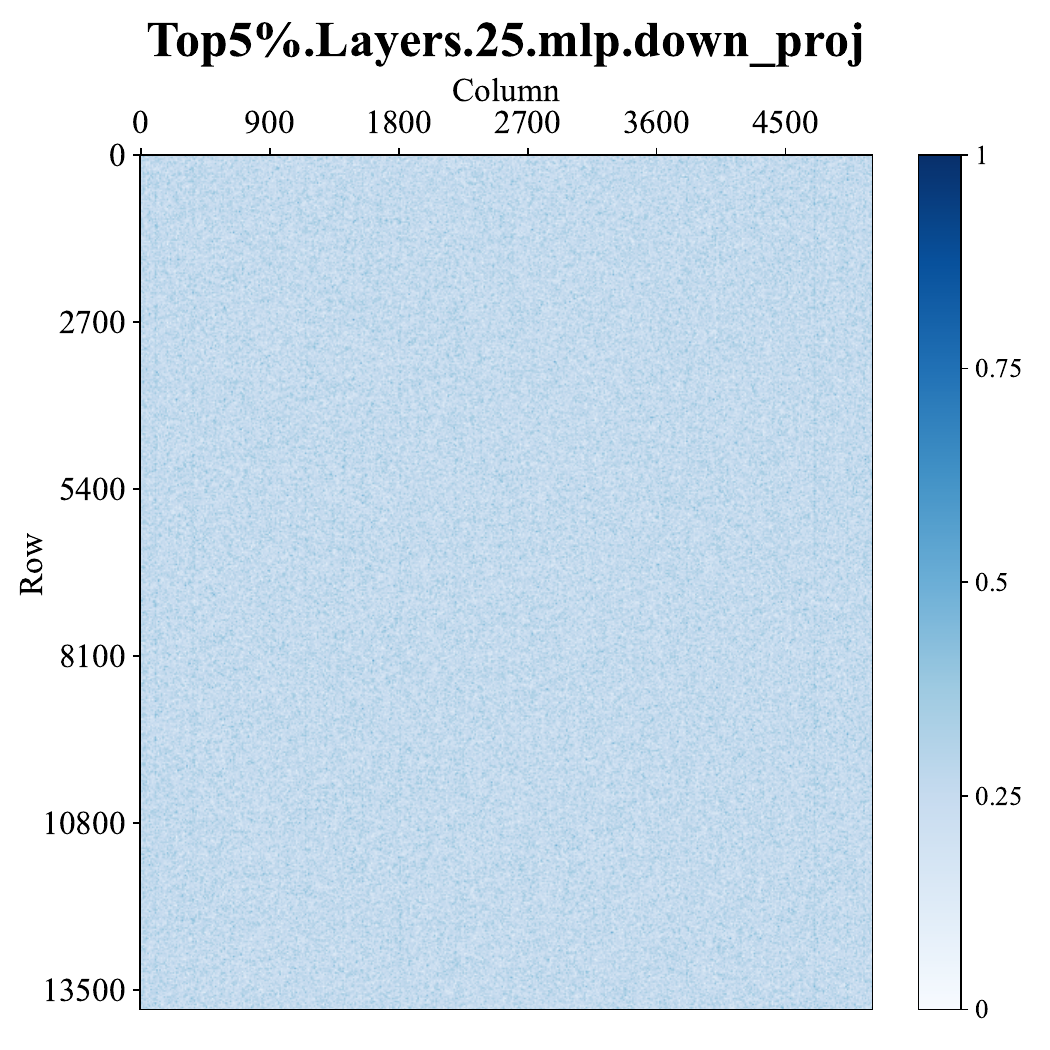}
	\end{minipage}
	}
	\subfigure{
		\centering
	\begin{minipage}[t]{0.23\textwidth}
		\includegraphics[width=3.6cm]{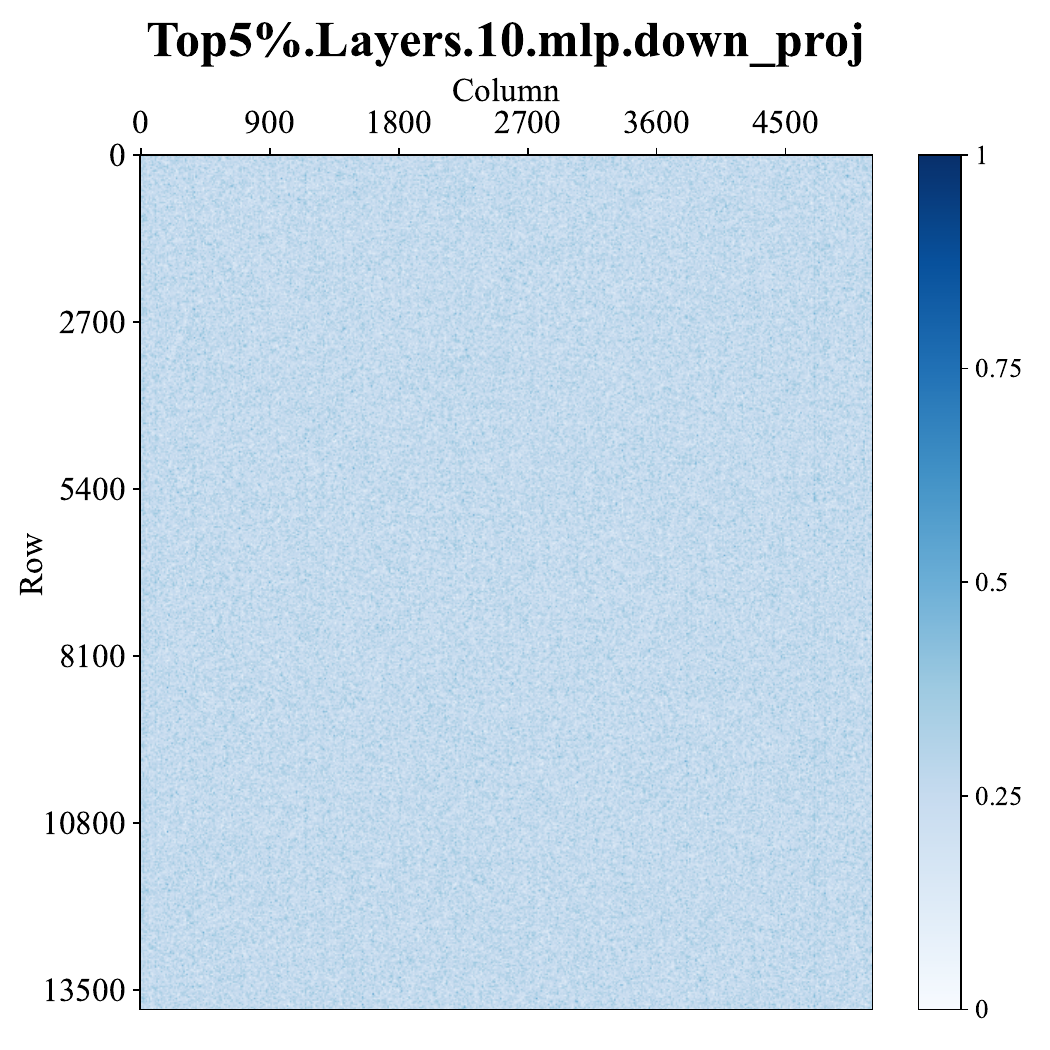}
		\includegraphics[width=3.6cm]{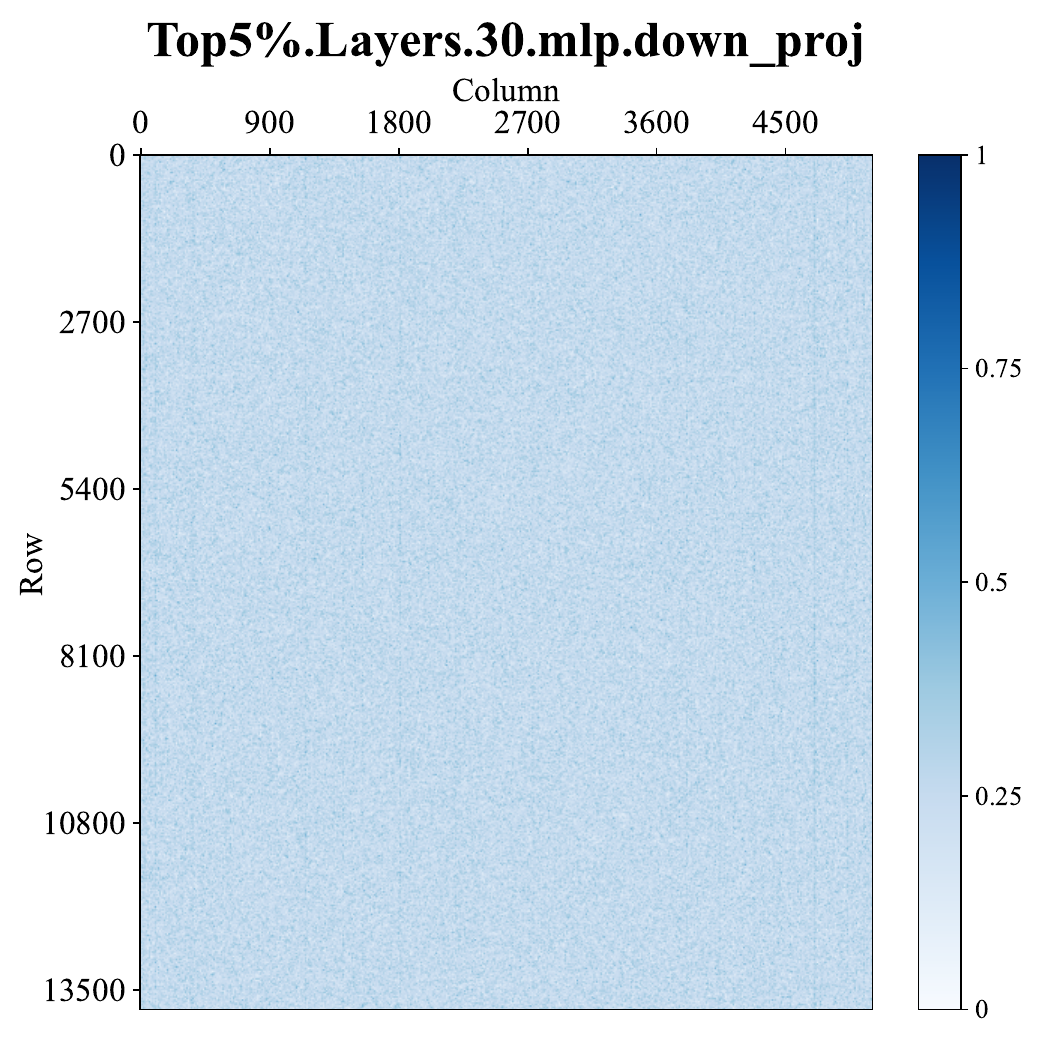}
	\end{minipage}
	}
 \subfigure{
		\centering
	\begin{minipage}[t]{0.23\textwidth}
		\includegraphics[width=3.6cm]{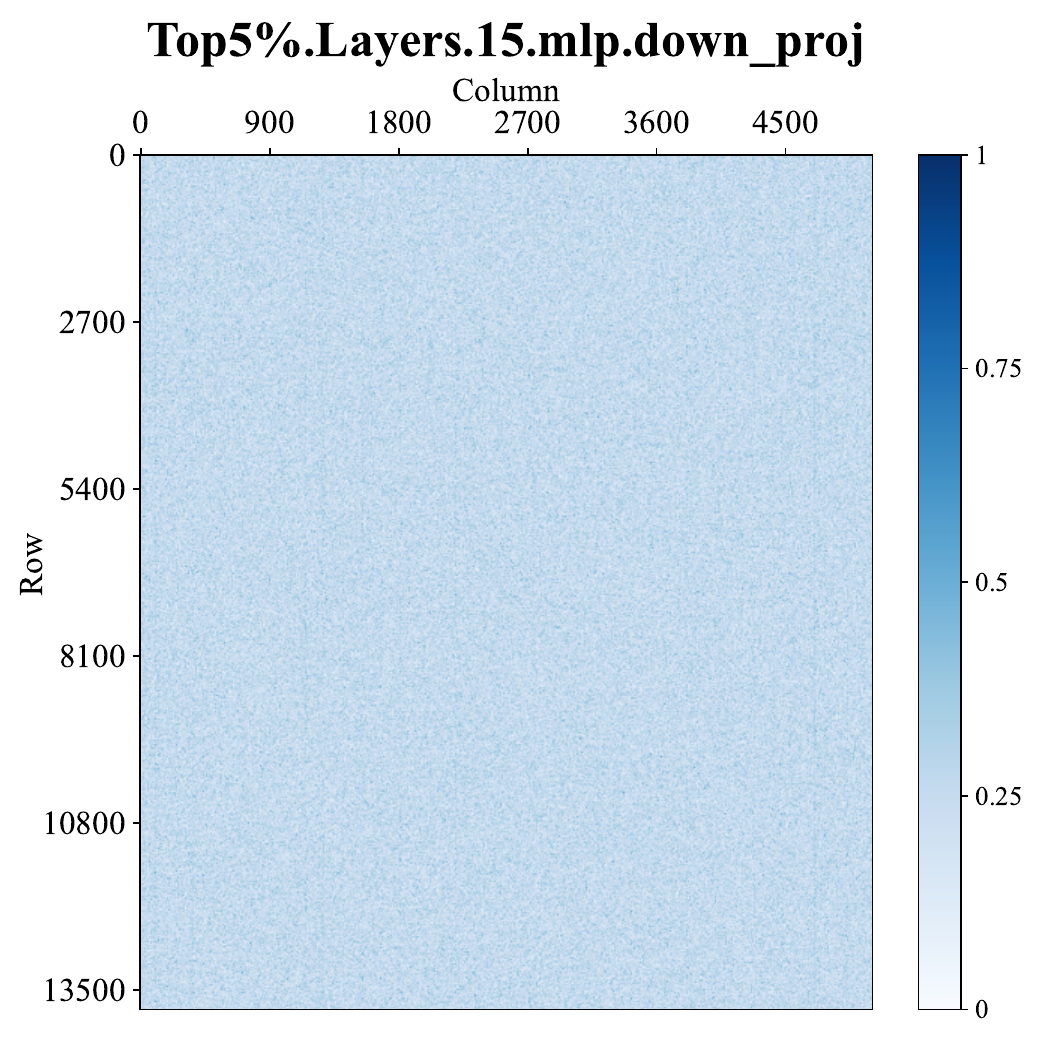}
		\includegraphics[width=3.6cm]{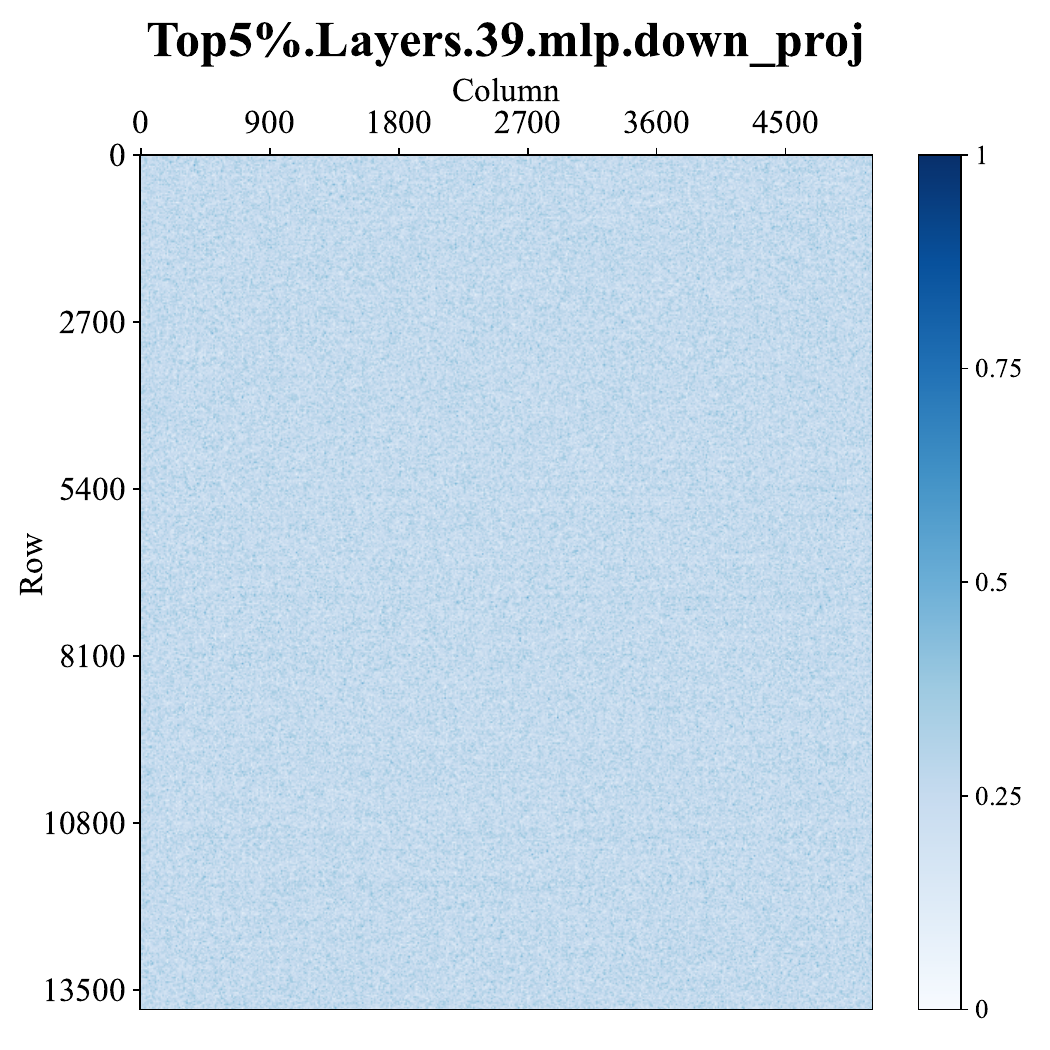}
	\end{minipage}
	}
 \caption{Visualization of FFn.down's `Top' region in LLaMA2-13b. The scale from 0 to 1 (after normalization) represent the proportion of parameters within a $3\times3$ vicinity that belong to the Bottom region.}
\label{fig:app_visualize_13b_down}
\end{figure*}

\begin{figure*}[t]
	\centering
	\subfigure{
		\centering
	\begin{minipage}[t]{0.23\textwidth}
        \includegraphics[width=3.6cm]{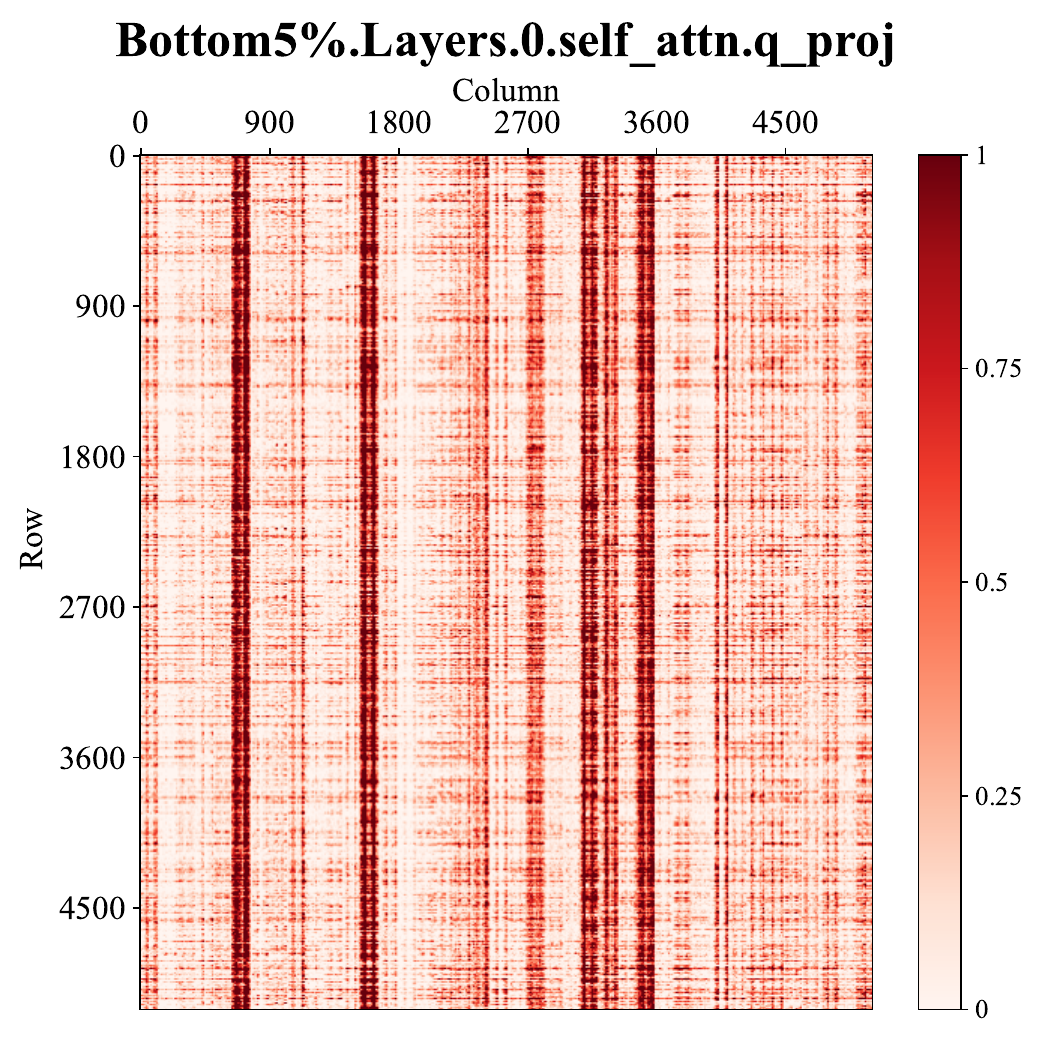}
        \includegraphics[width=3.6cm]{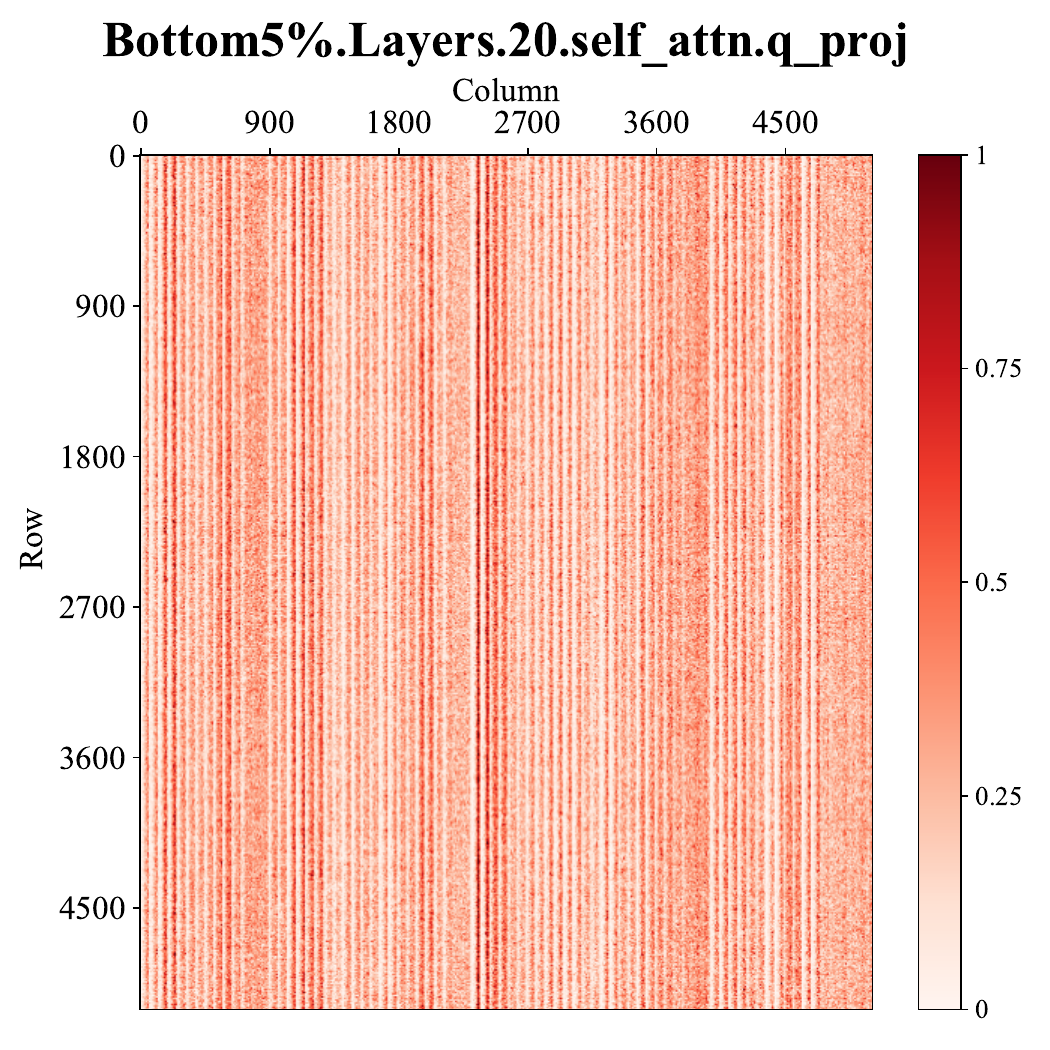}
	\end{minipage}
	}
	\subfigure{
		\centering
	\begin{minipage}[t]{0.23\textwidth}
	       \includegraphics[width=3.6cm]{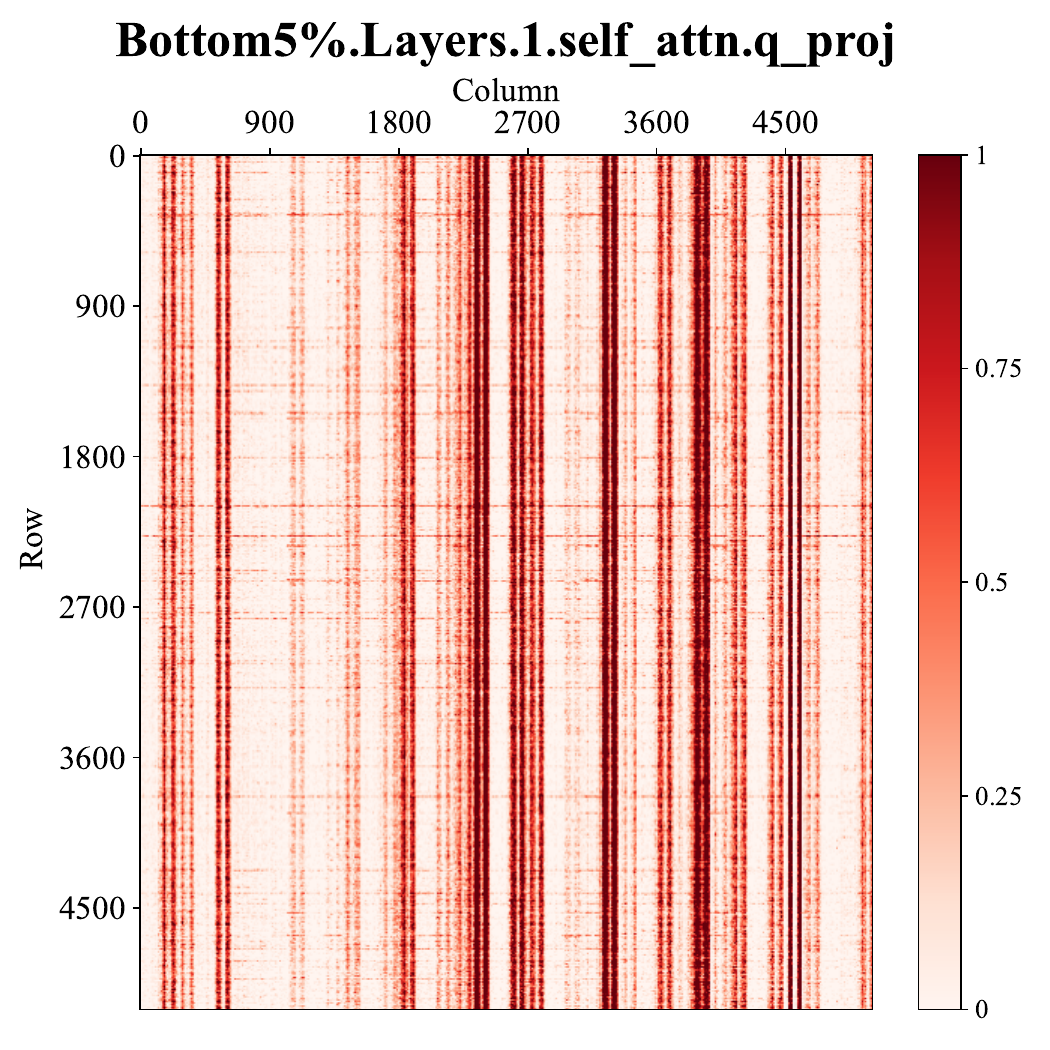}
            \includegraphics[width=3.6cm]{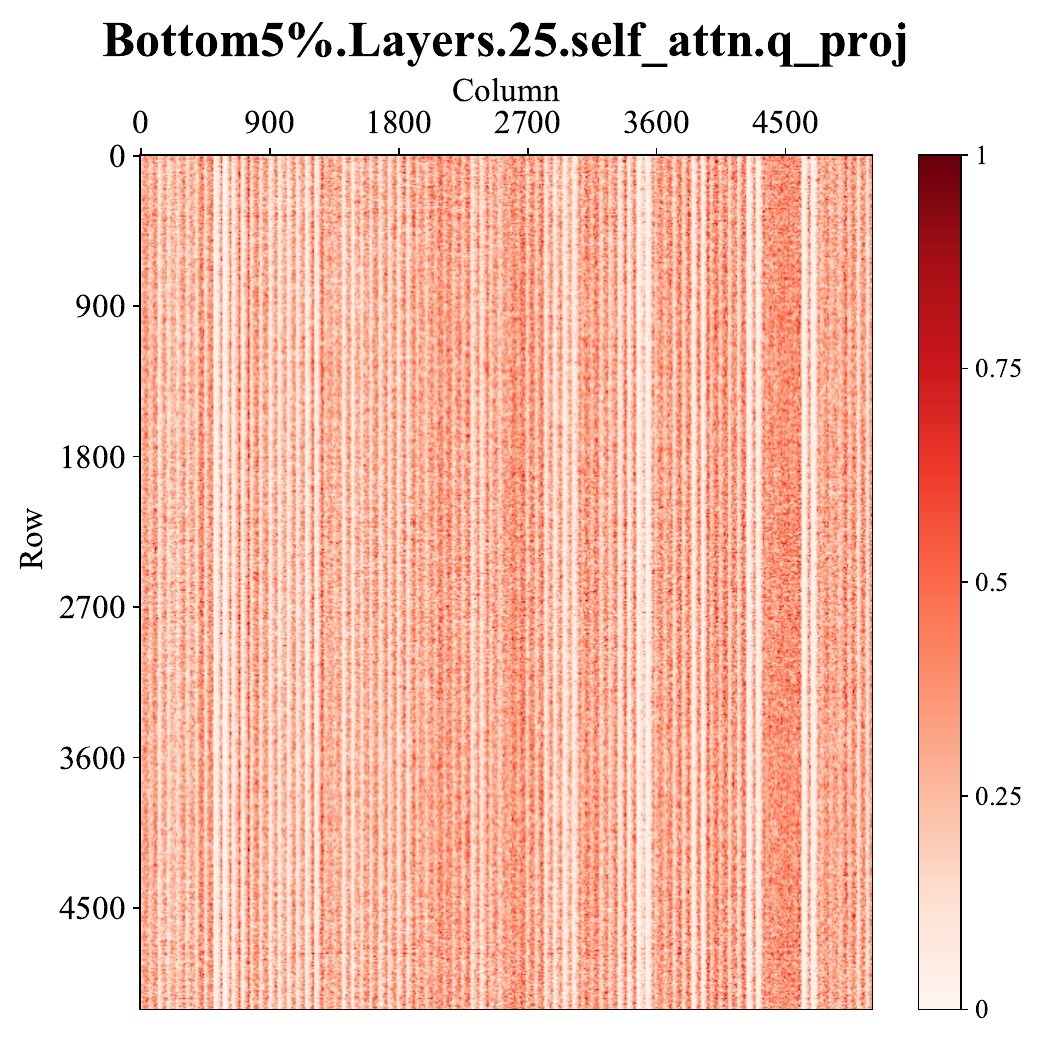}
	\end{minipage}
	}
	\subfigure{
		\centering
	\begin{minipage}[t]{0.23\textwidth}
		\includegraphics[width=3.6cm]{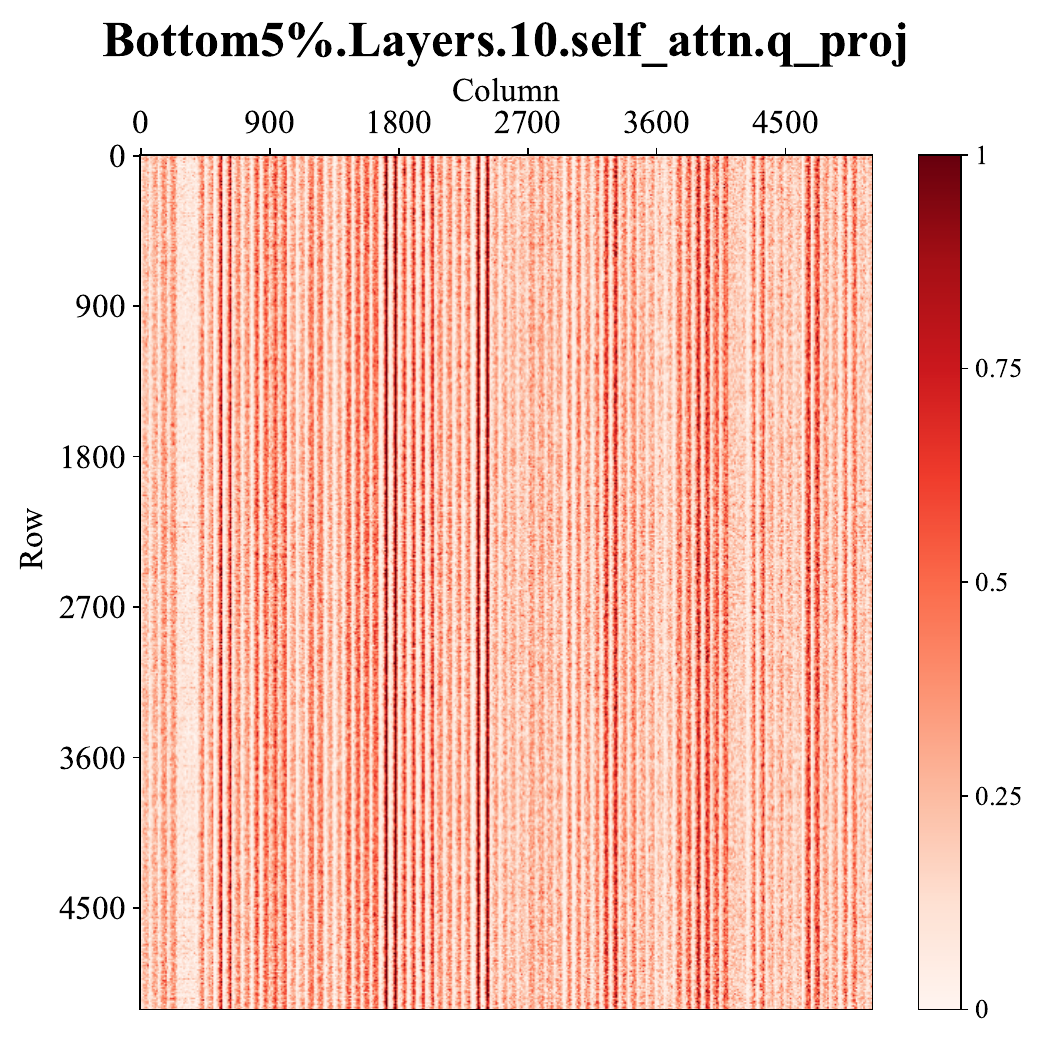}
		\includegraphics[width=3.6cm]{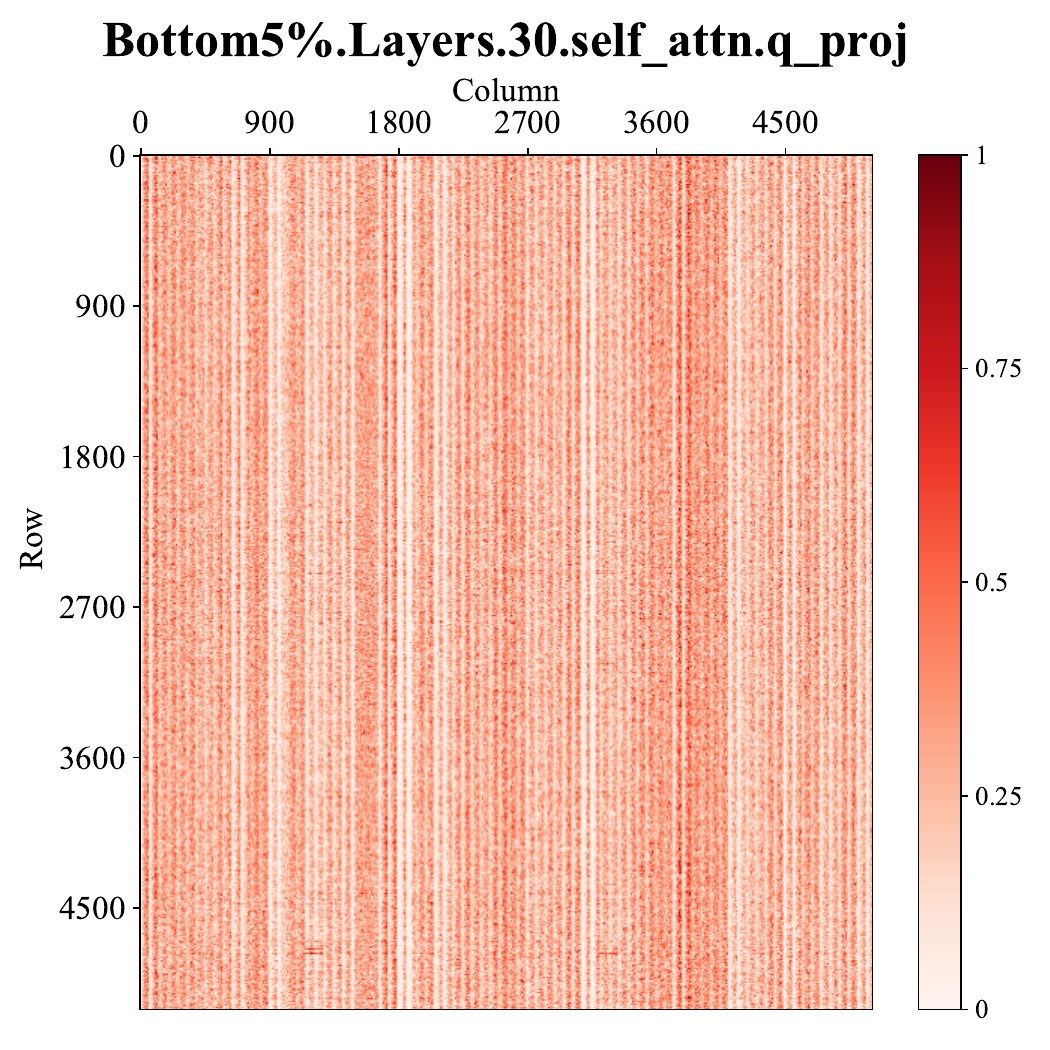}
	\end{minipage}
	}
 \subfigure{
		\centering
	\begin{minipage}[t]{0.23\textwidth}
		\includegraphics[width=3.6cm]{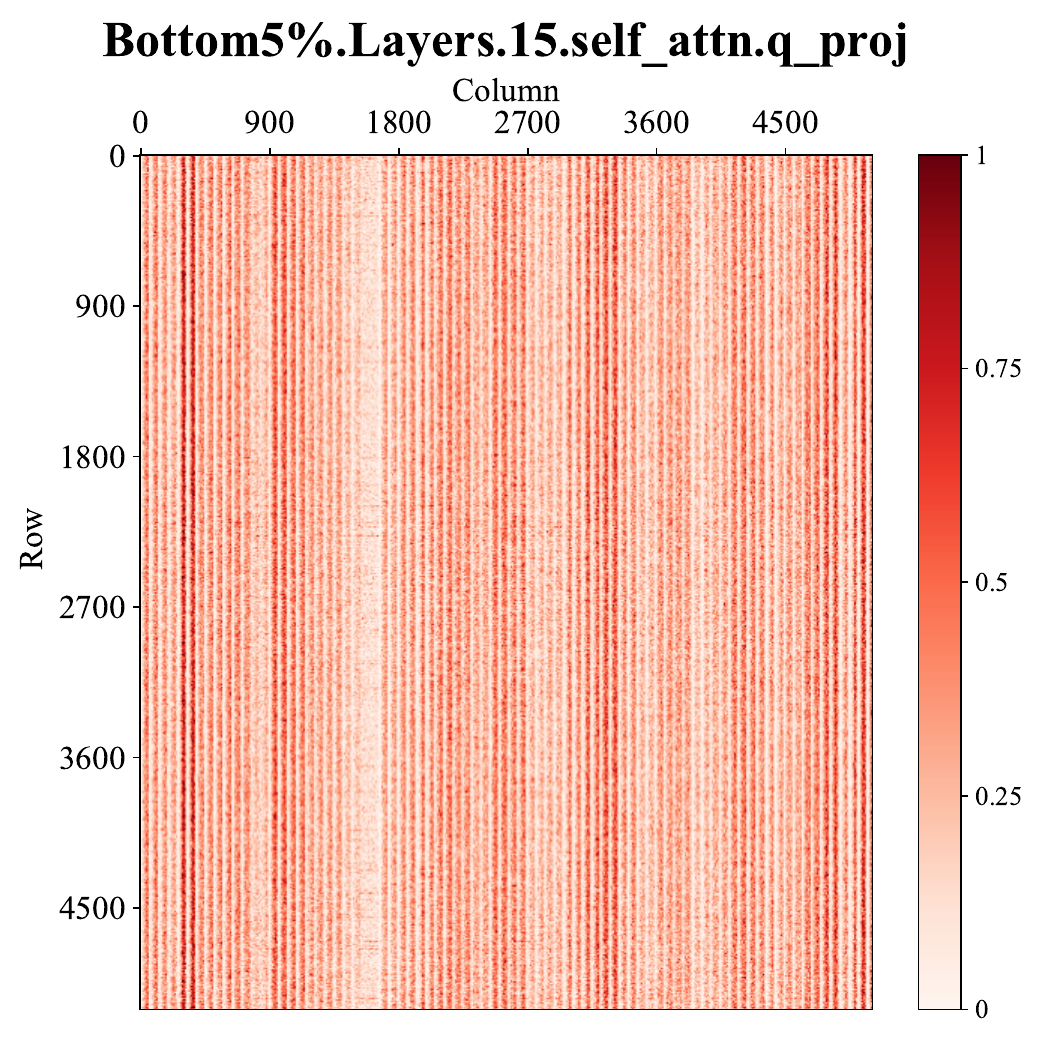}
		\includegraphics[width=3.6cm]{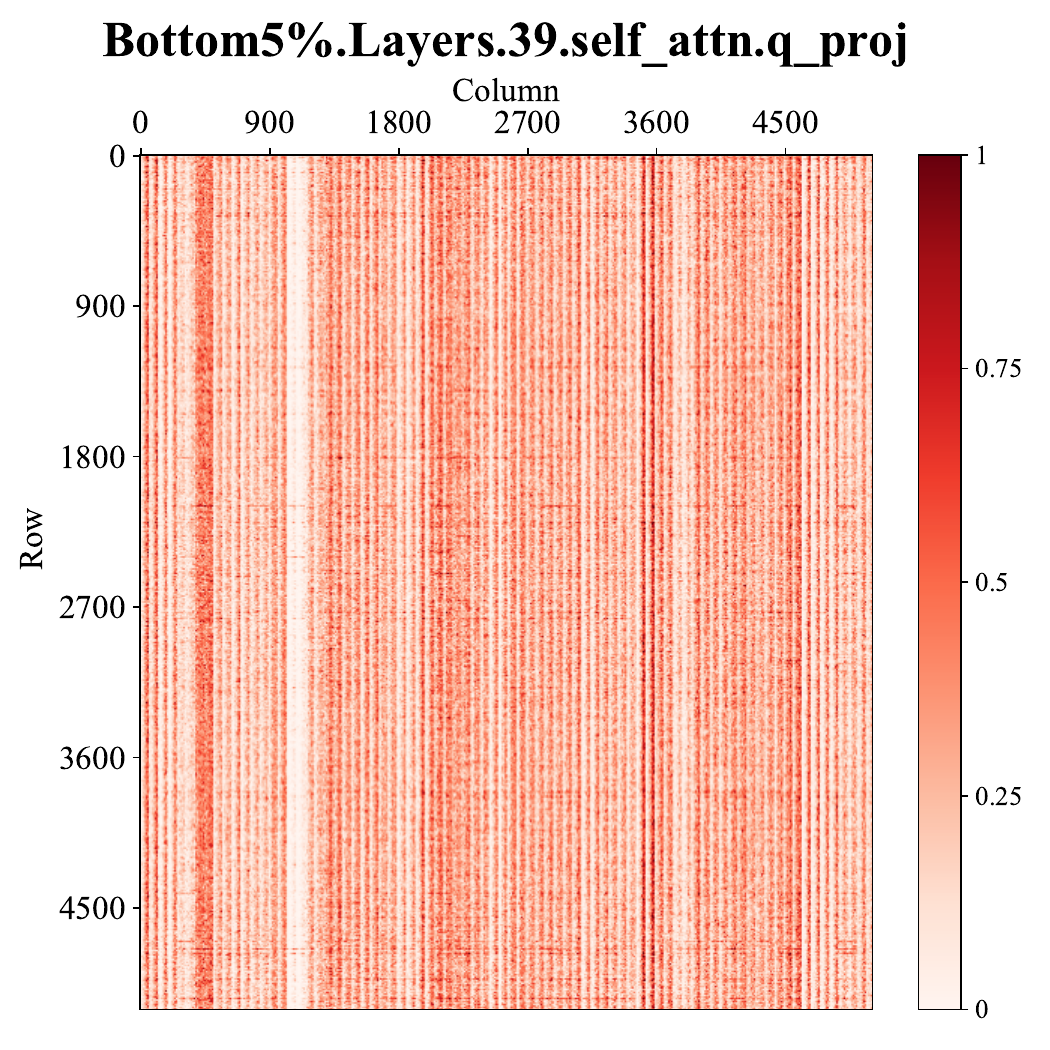}
	\end{minipage}
	}
 \caption{Visualization of Attn.q's `Bottom' region in LLaMA2-13b. The scale from 0 to 1 (after normalization) represent the proportion of parameters within a $3\times3$ vicinity that belong to the Bottom region.}
\label{fig:app_visualize_13b_q_bot}
\end{figure*}

\begin{figure*}[t]
	\centering
	\subfigure{
		\centering
	\begin{minipage}[t]{0.23\textwidth}
        \includegraphics[width=3.6cm]{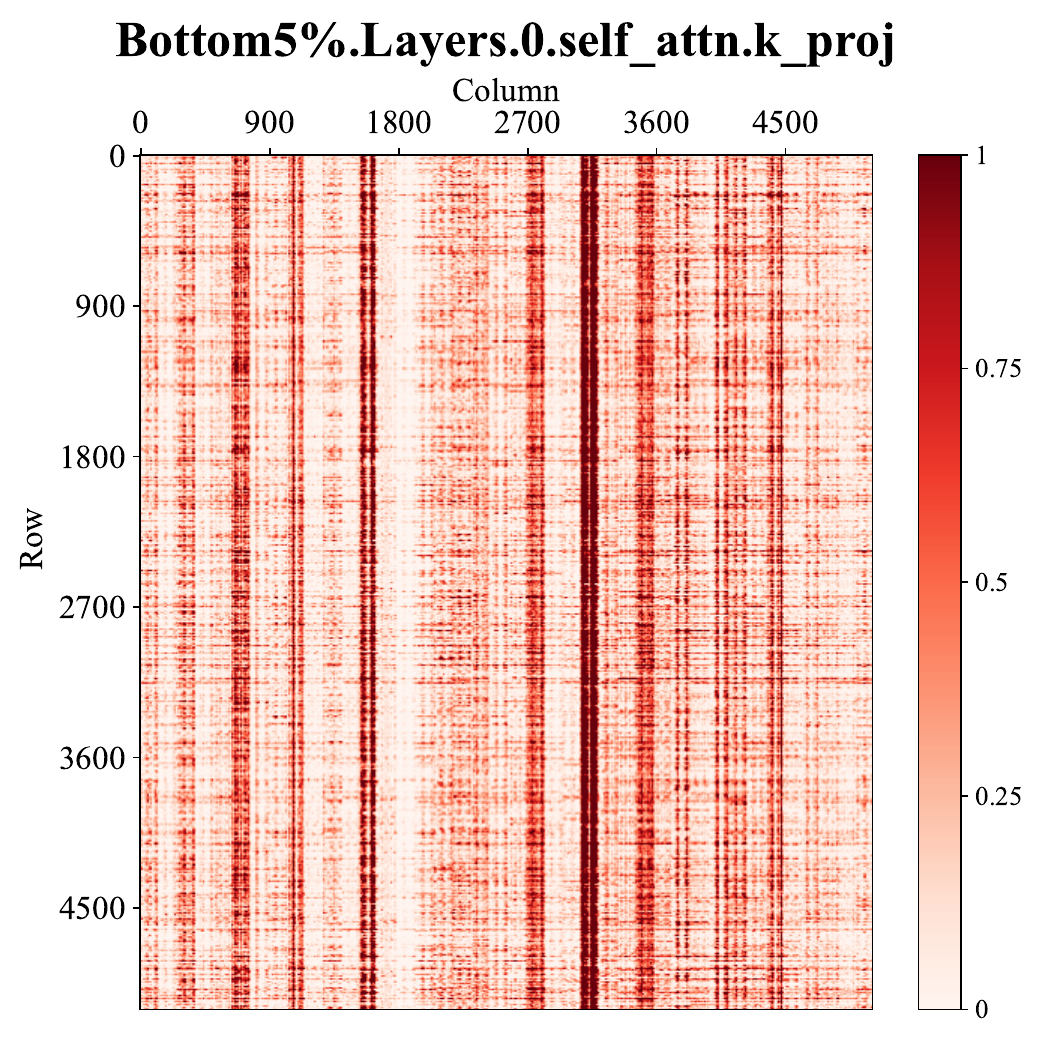}
        \includegraphics[width=3.6cm]{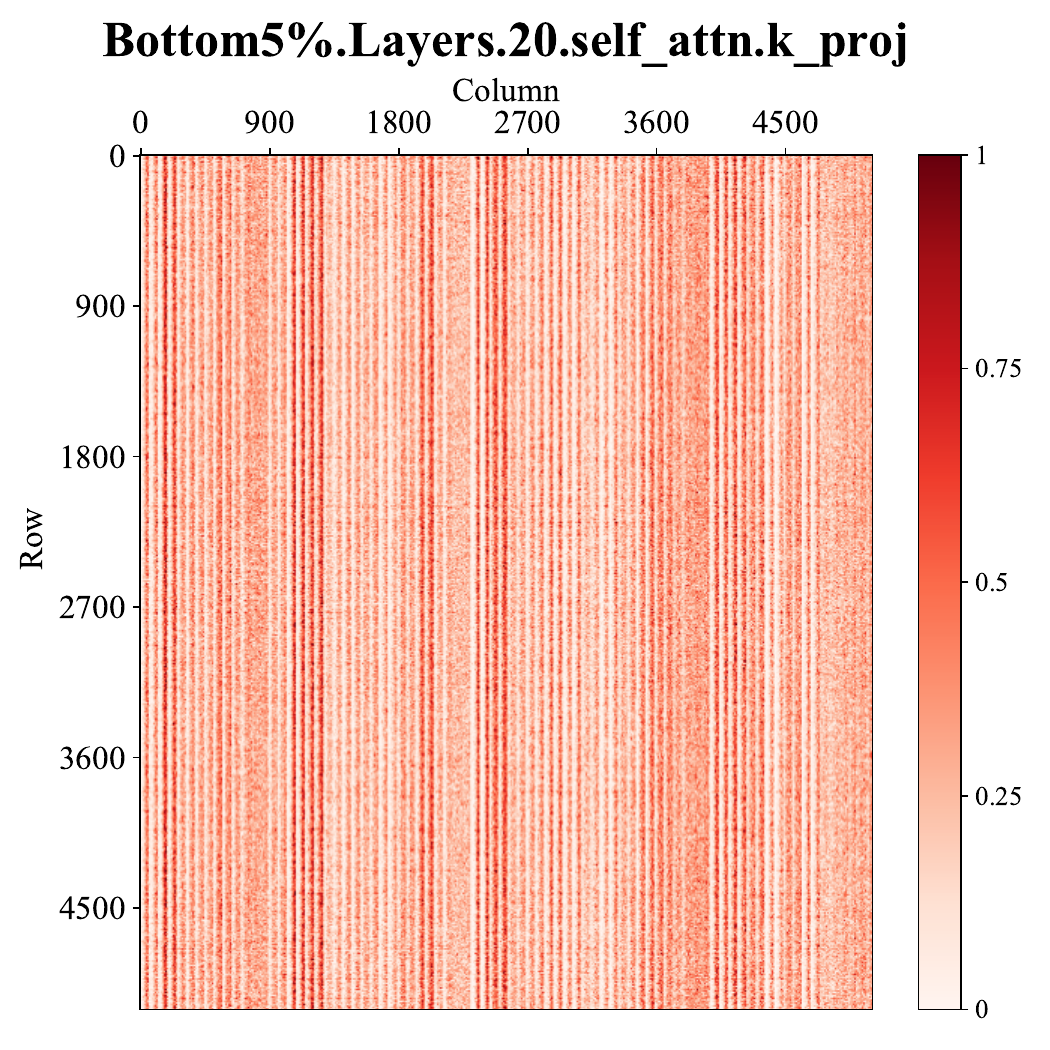}
	\end{minipage}
	}
	\subfigure{
		\centering
	\begin{minipage}[t]{0.23\textwidth}
	       \includegraphics[width=3.6cm]{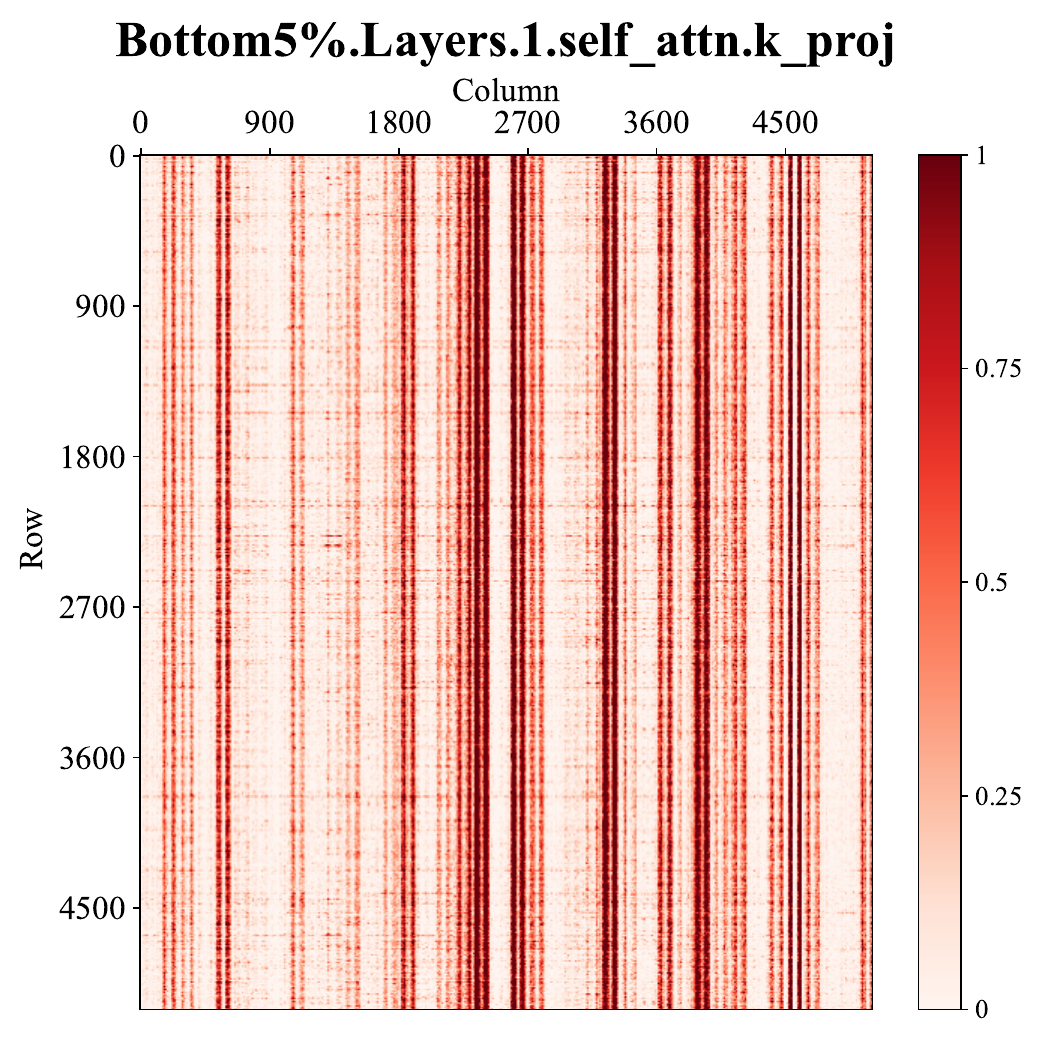}
            \includegraphics[width=3.6cm]{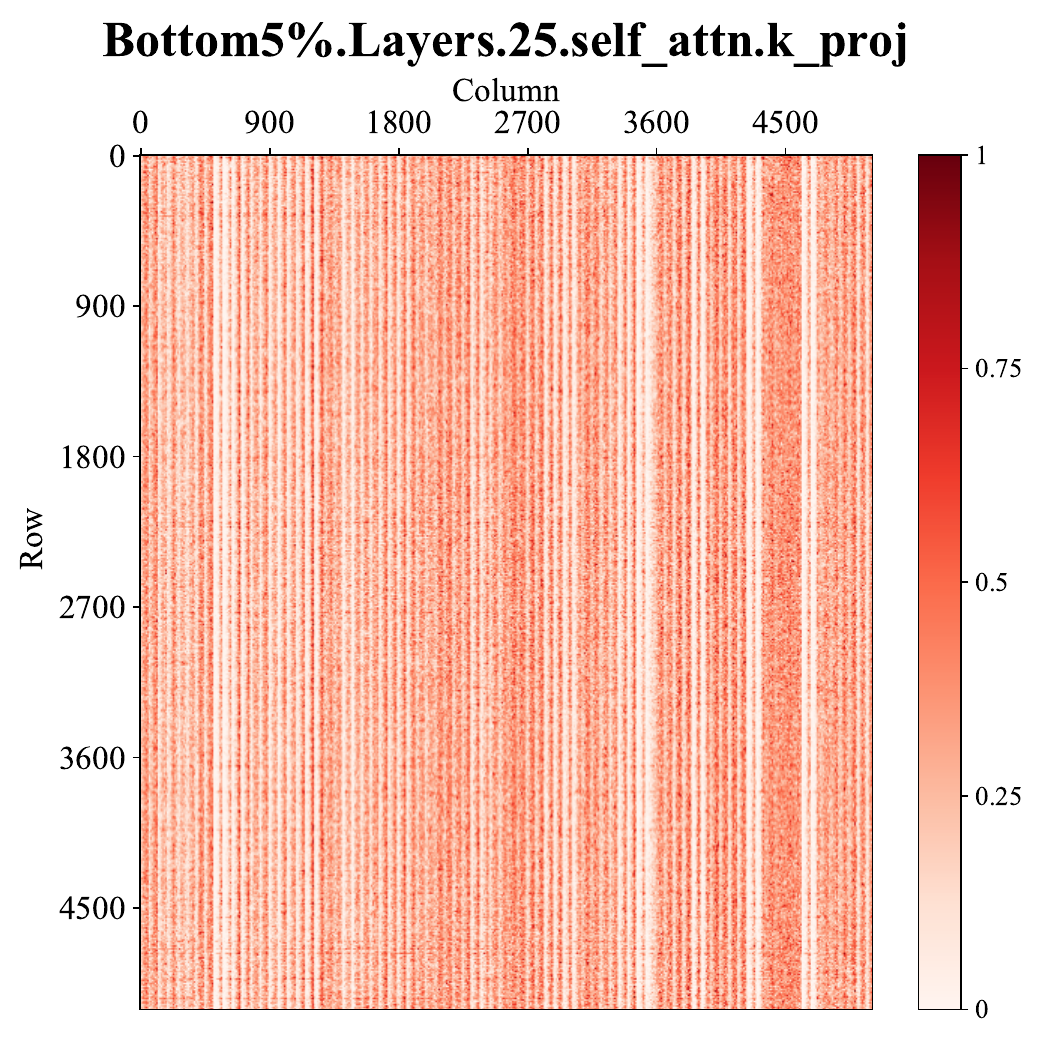}
	\end{minipage}
	}
	\subfigure{
		\centering
	\begin{minipage}[t]{0.23\textwidth}
		\includegraphics[width=3.6cm]{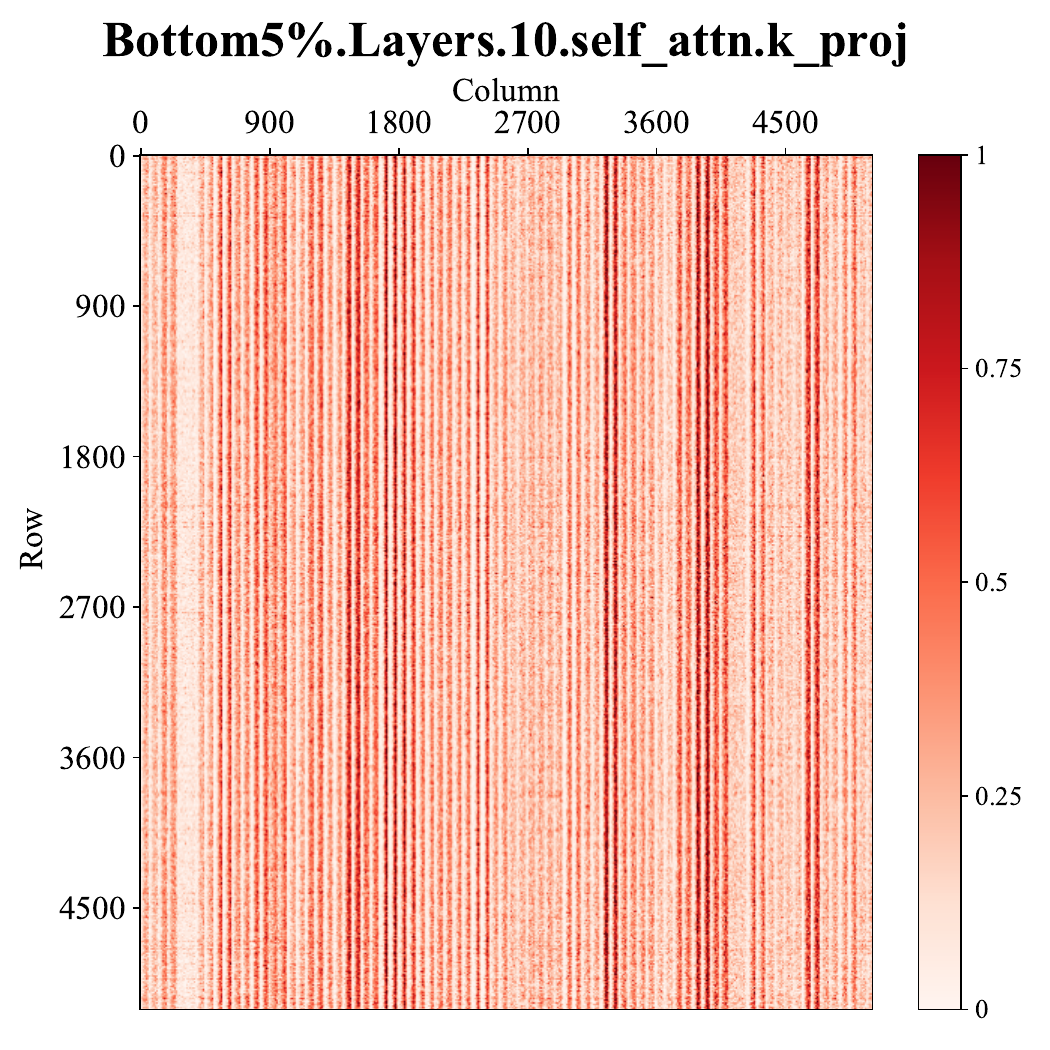}
		\includegraphics[width=3.6cm]{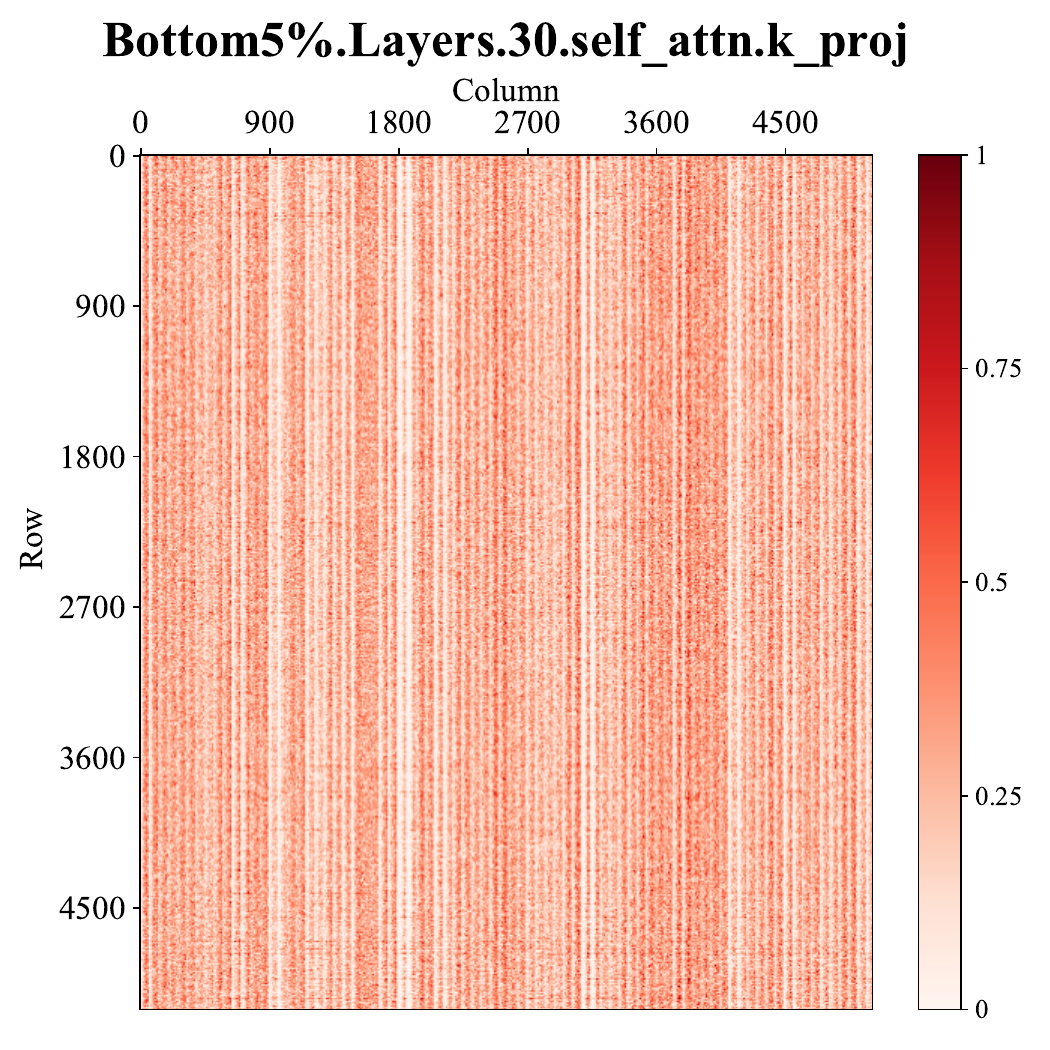}
	\end{minipage}
	}
 \subfigure{
		\centering
	\begin{minipage}[t]{0.23\textwidth}
		\includegraphics[width=3.6cm]{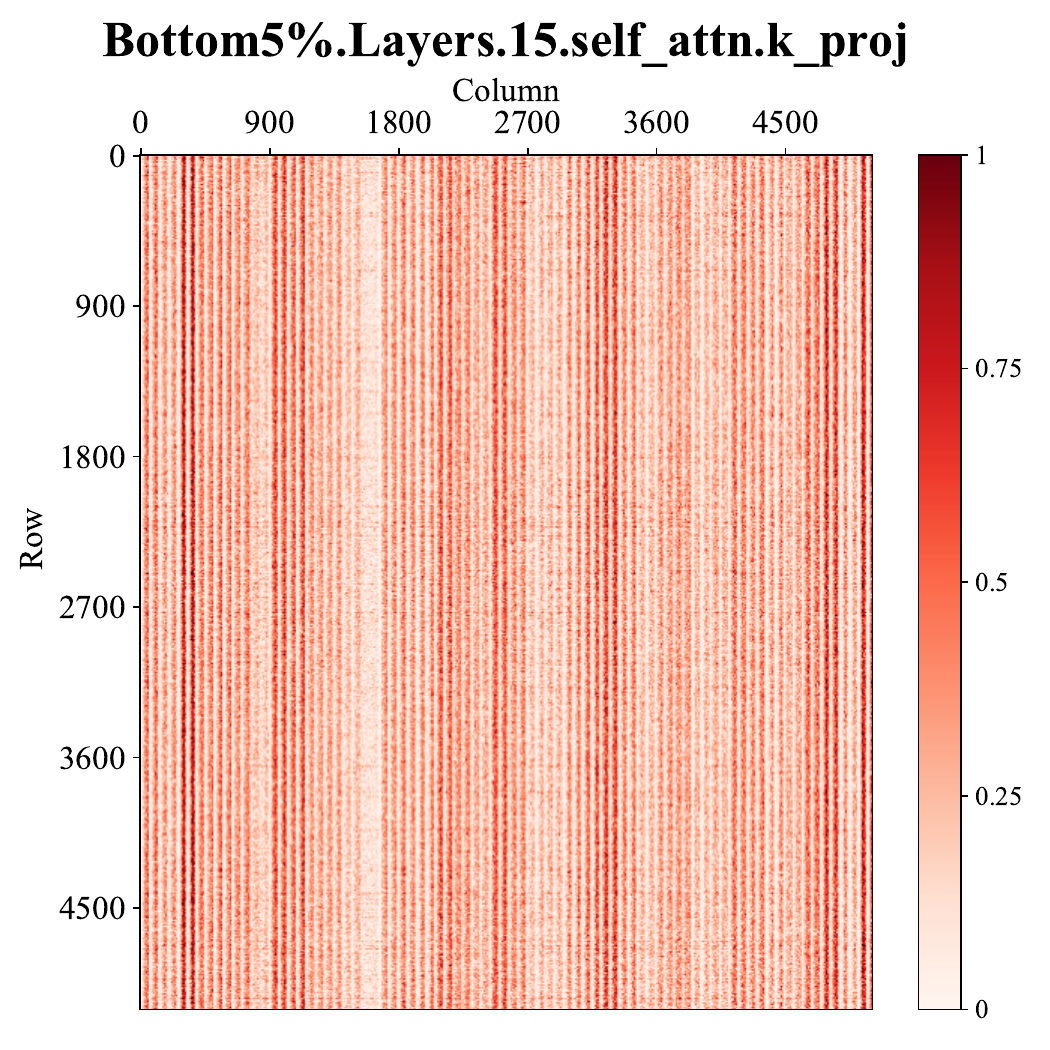}
		\includegraphics[width=3.6cm]{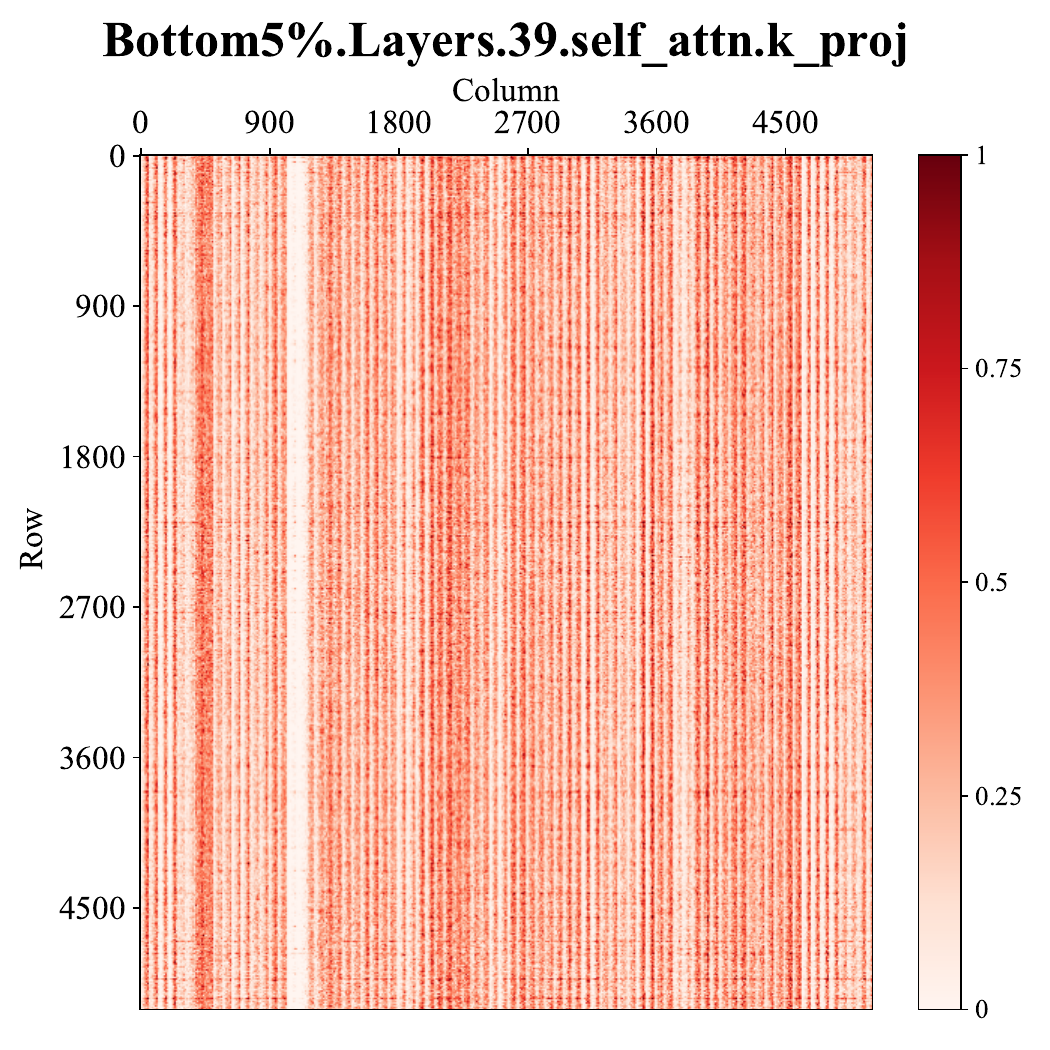}
	\end{minipage}
	}
 \caption{Visualization of Attn.k's `Bottom' region in LLaMA2-13b. The scale from 0 to 1 (after normalization) represent the proportion of parameters within a $3\times3$ vicinity that belong to the Bottom region.}
\label{fig:app_visualize_13b_k_bot}
\end{figure*}

\begin{figure*}[t]
	\centering
	\subfigure{
		\centering
	\begin{minipage}[t]{0.23\textwidth}
        \includegraphics[width=3.6cm]{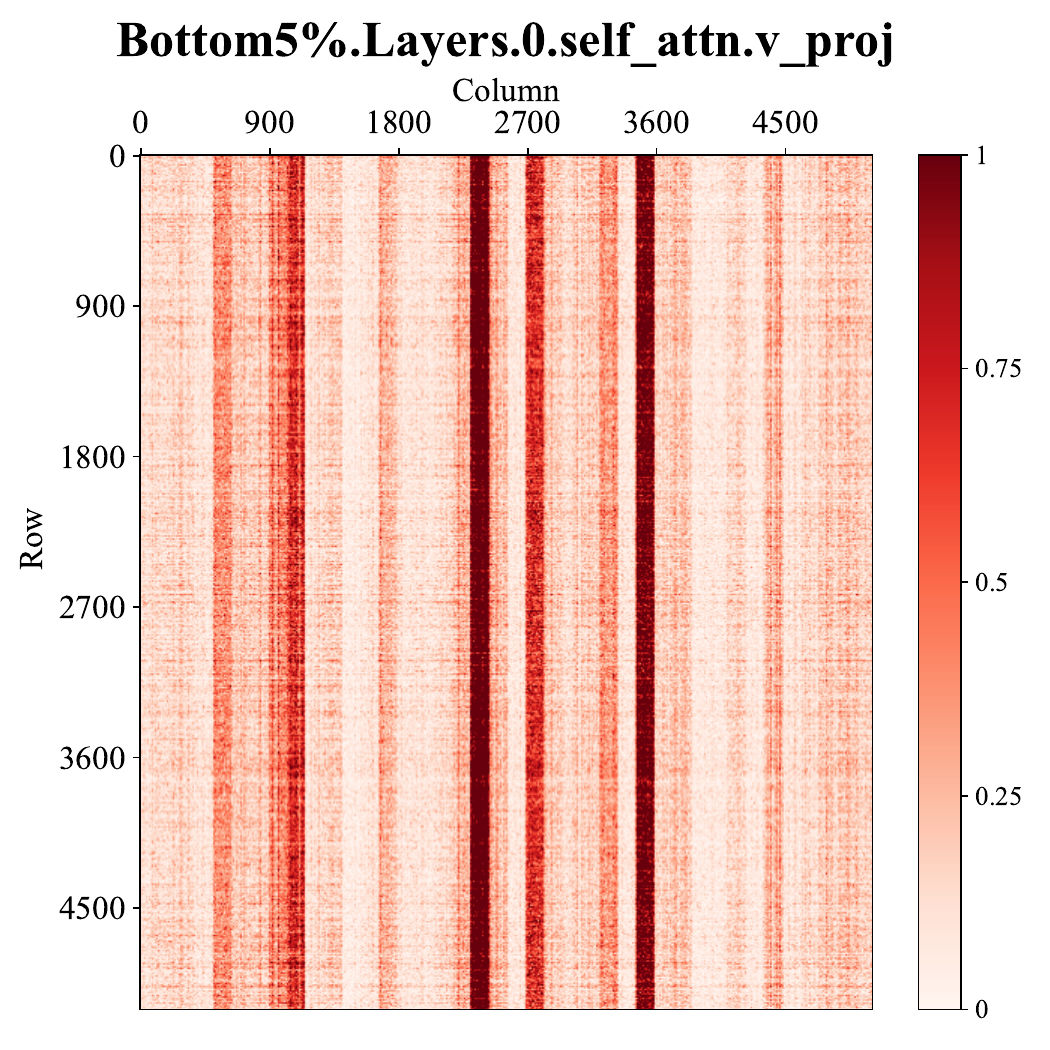}
        \includegraphics[width=3.6cm]{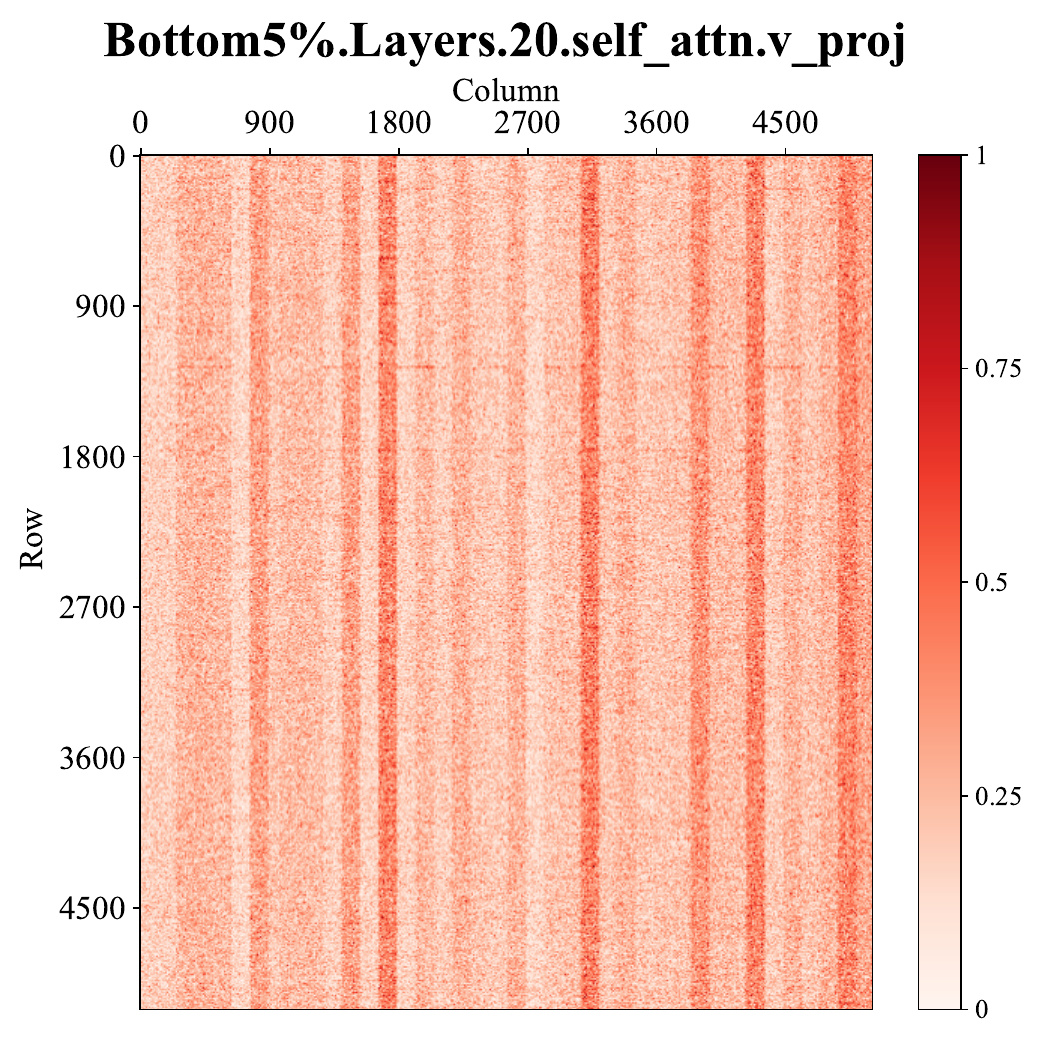}
	\end{minipage}
	}
	\subfigure{
		\centering
	\begin{minipage}[t]{0.23\textwidth}
	       \includegraphics[width=3.6cm]{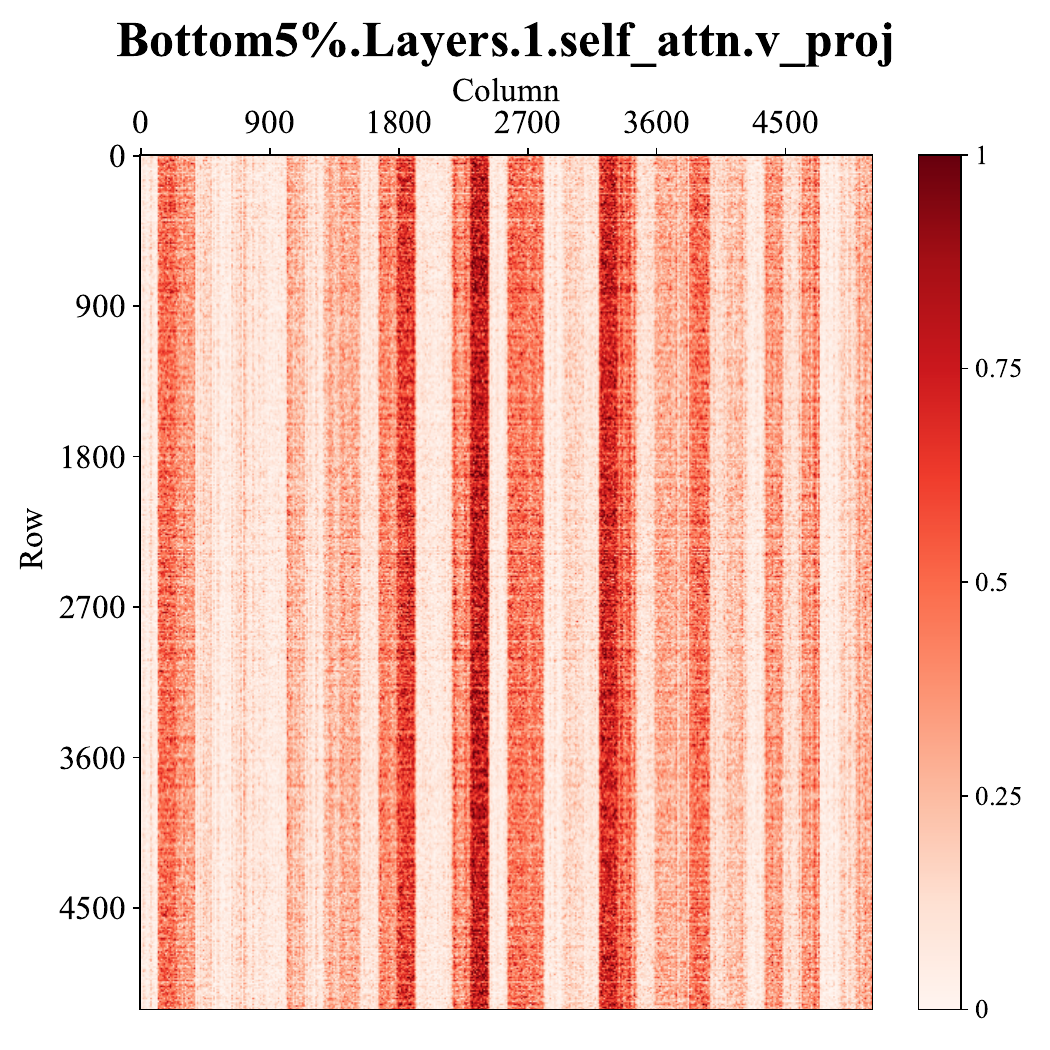}
            \includegraphics[width=3.6cm]{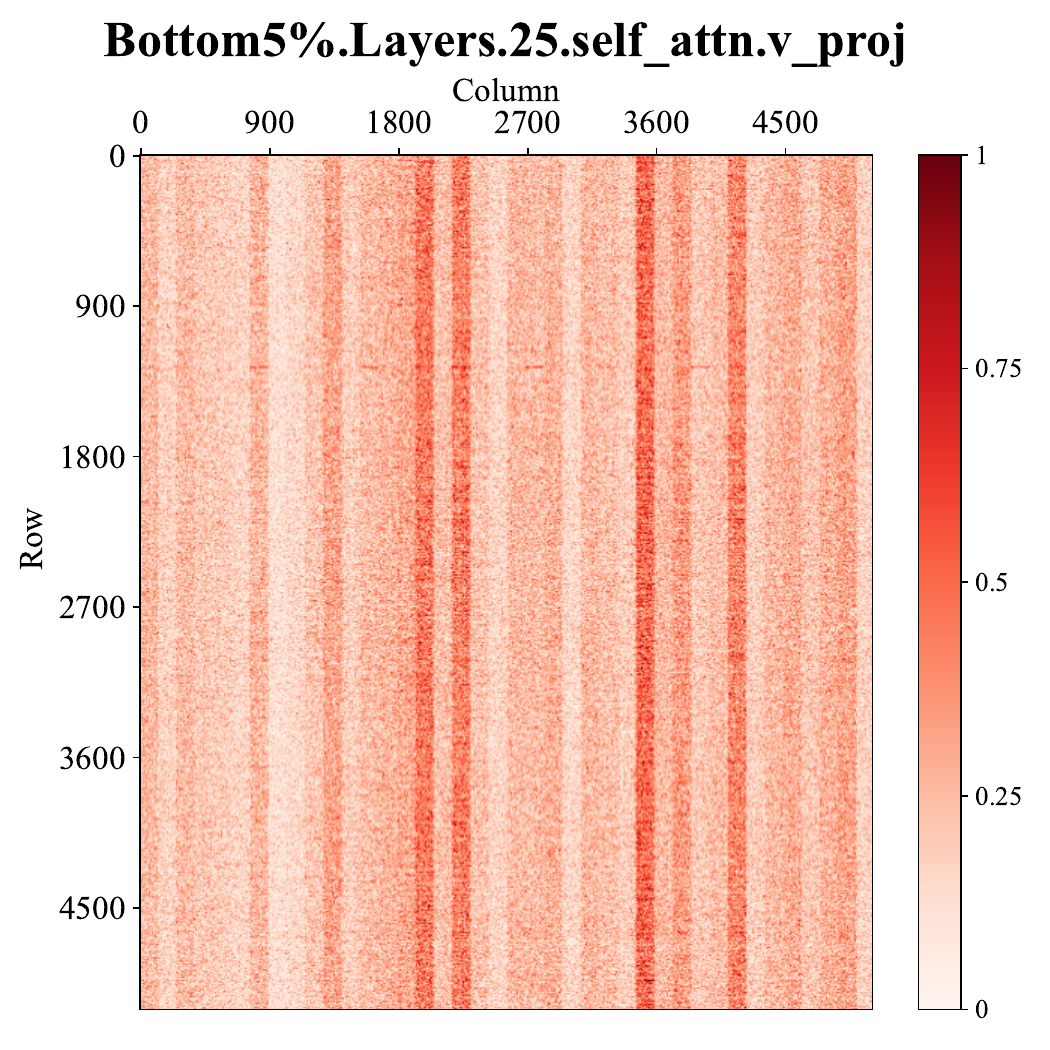}
	\end{minipage}
	}
	\subfigure{
		\centering
	\begin{minipage}[t]{0.23\textwidth}
		\includegraphics[width=3.6cm]{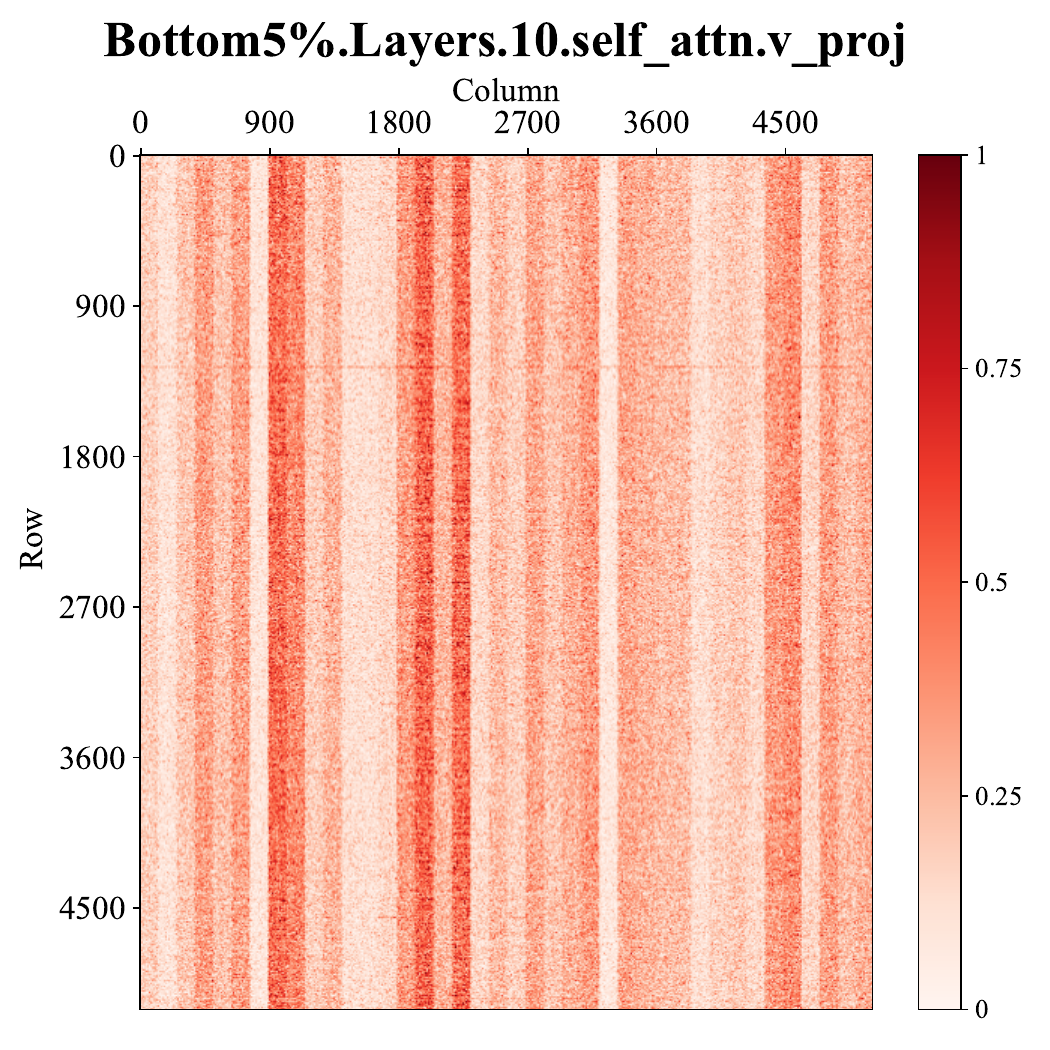}
		\includegraphics[width=3.6cm]{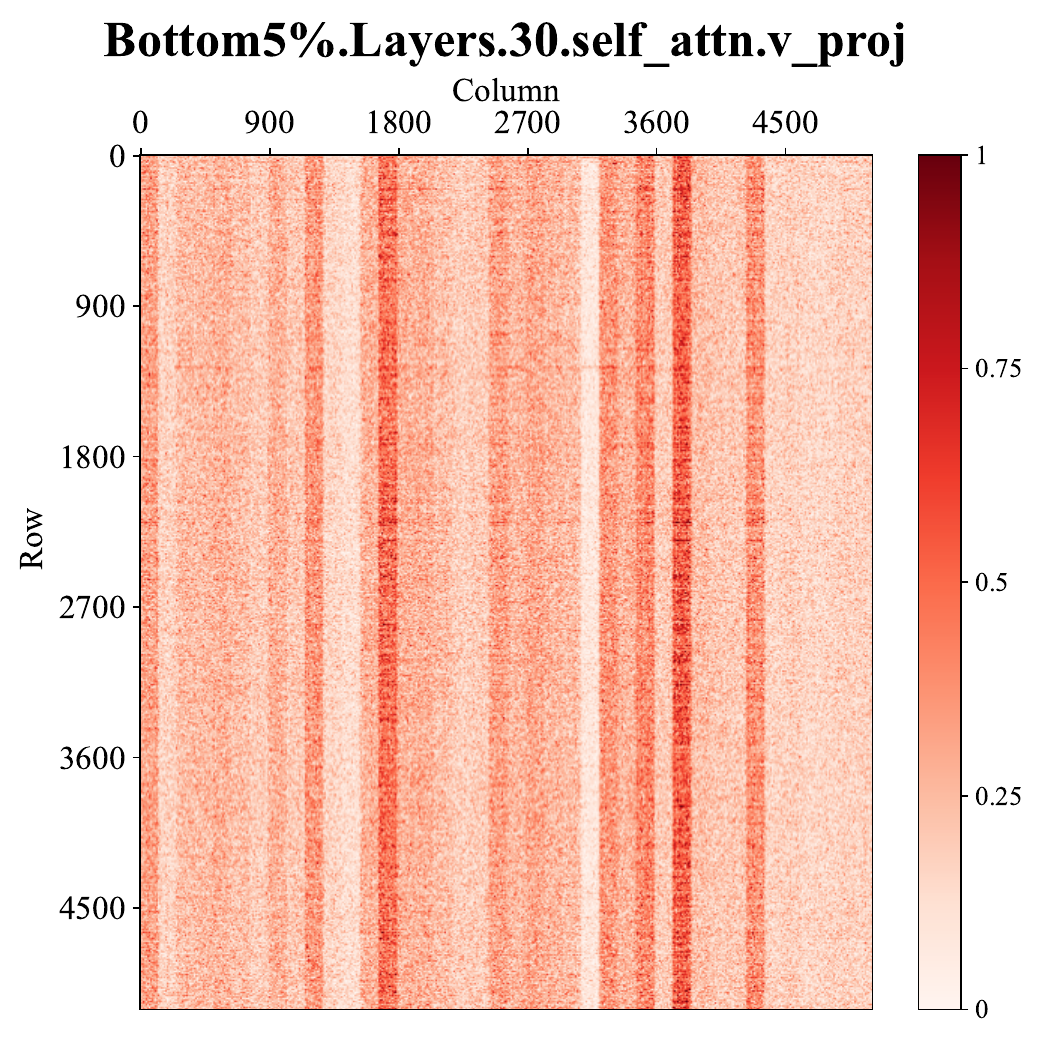}
	\end{minipage}
	}
 \subfigure{
		\centering
	\begin{minipage}[t]{0.23\textwidth}
		\includegraphics[width=3.6cm]{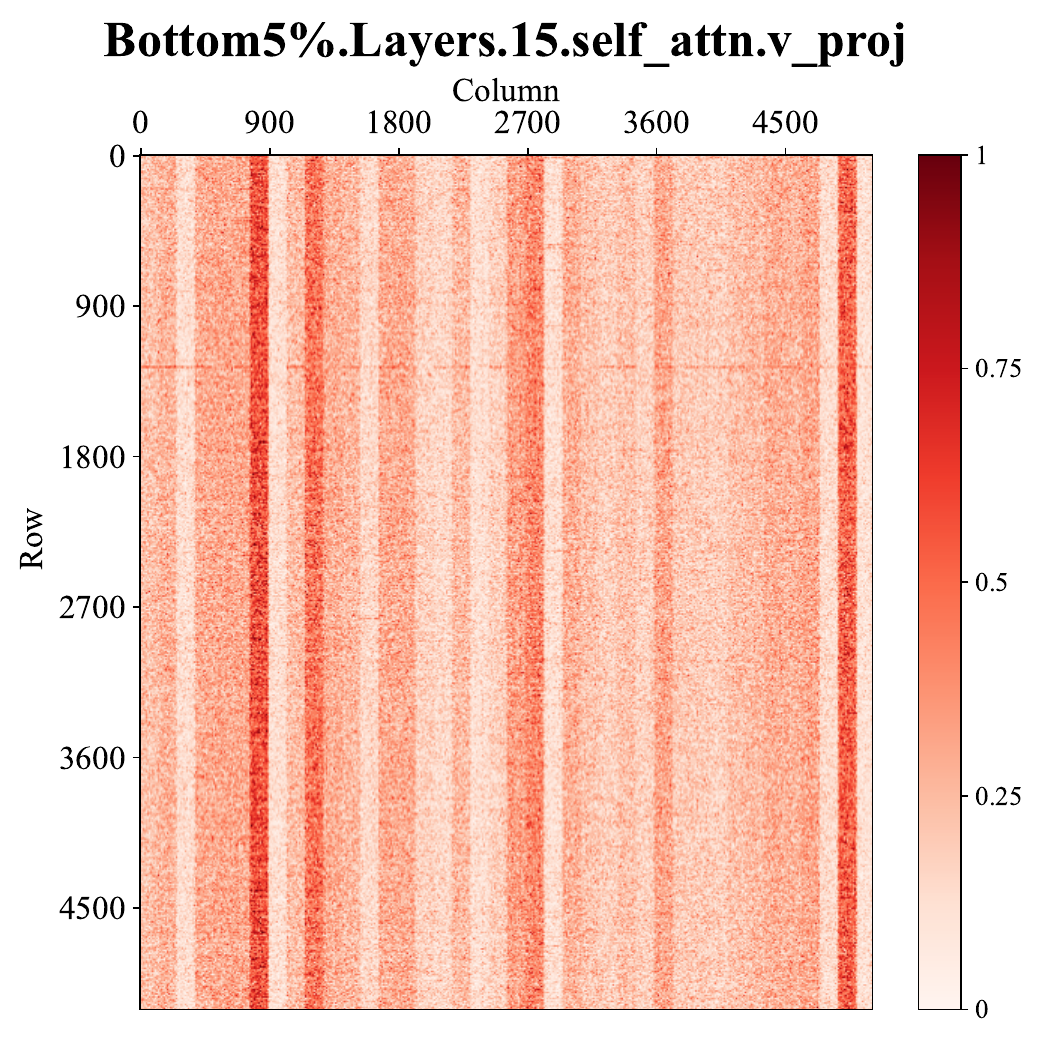}
		\includegraphics[width=3.6cm]{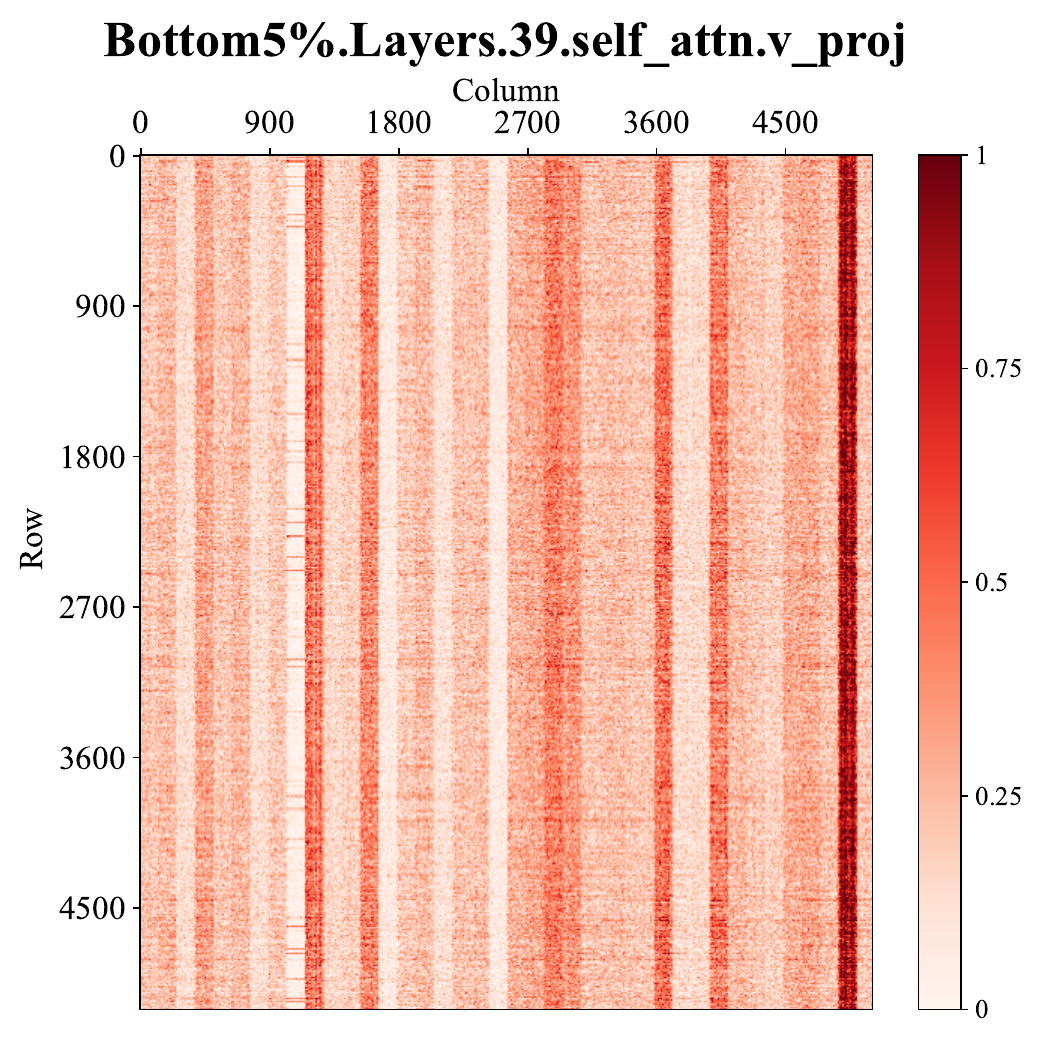}
	\end{minipage}
	}
 \caption{Visualization of Attn.v's `Bottom' region in LLaMA2-13b. The scale from 0 to 1 (after normalization) represent the proportion of parameters within a $3\times3$ vicinity that belong to the Bottom region.}
\label{fig:app_visualize_13b_v_bot}
\end{figure*}

\begin{figure*}[t]
	\centering
	\subfigure{
		\centering
	\begin{minipage}[t]{0.23\textwidth}
        \includegraphics[width=3.6cm]{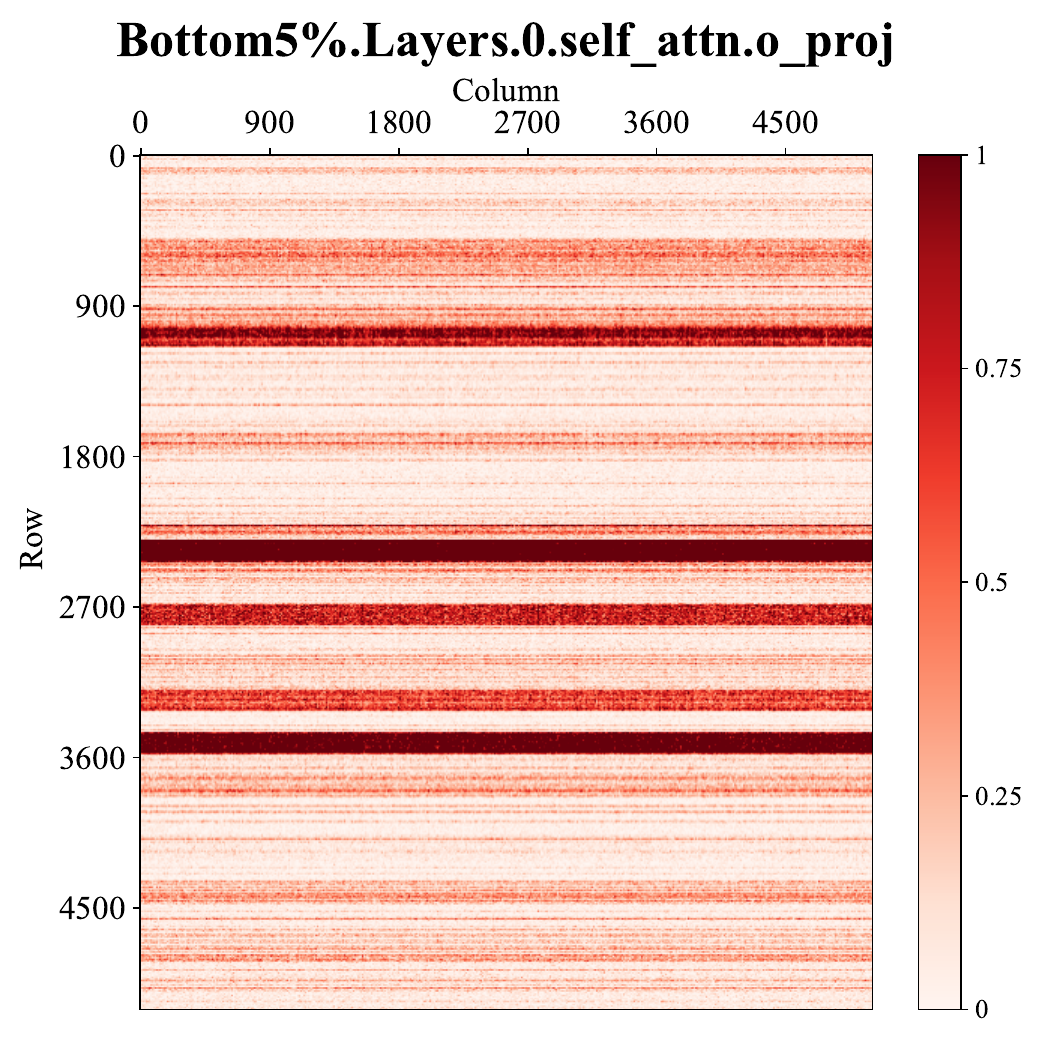}
        \includegraphics[width=3.6cm]{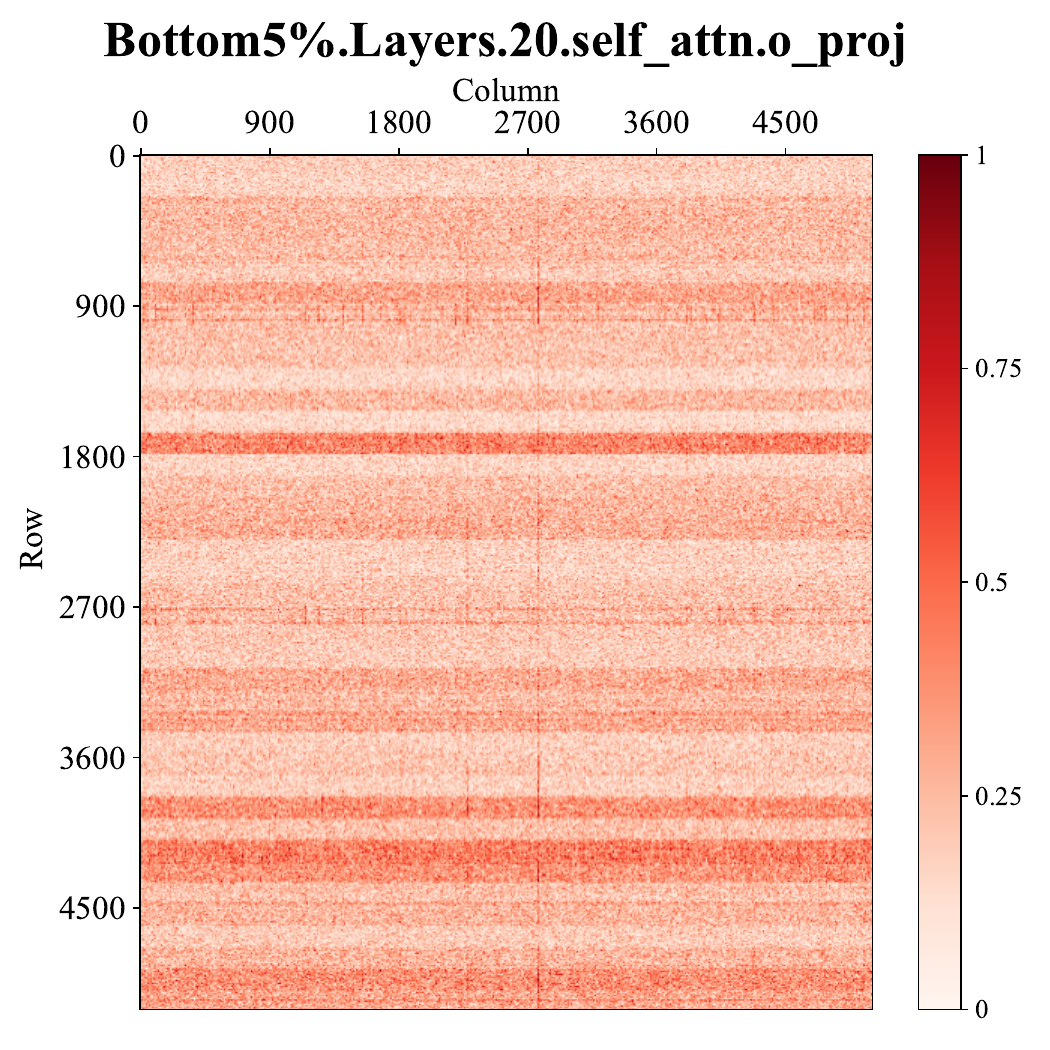}
	\end{minipage}
	}
	\subfigure{
		\centering
	\begin{minipage}[t]{0.23\textwidth}
	       \includegraphics[width=3.6cm]{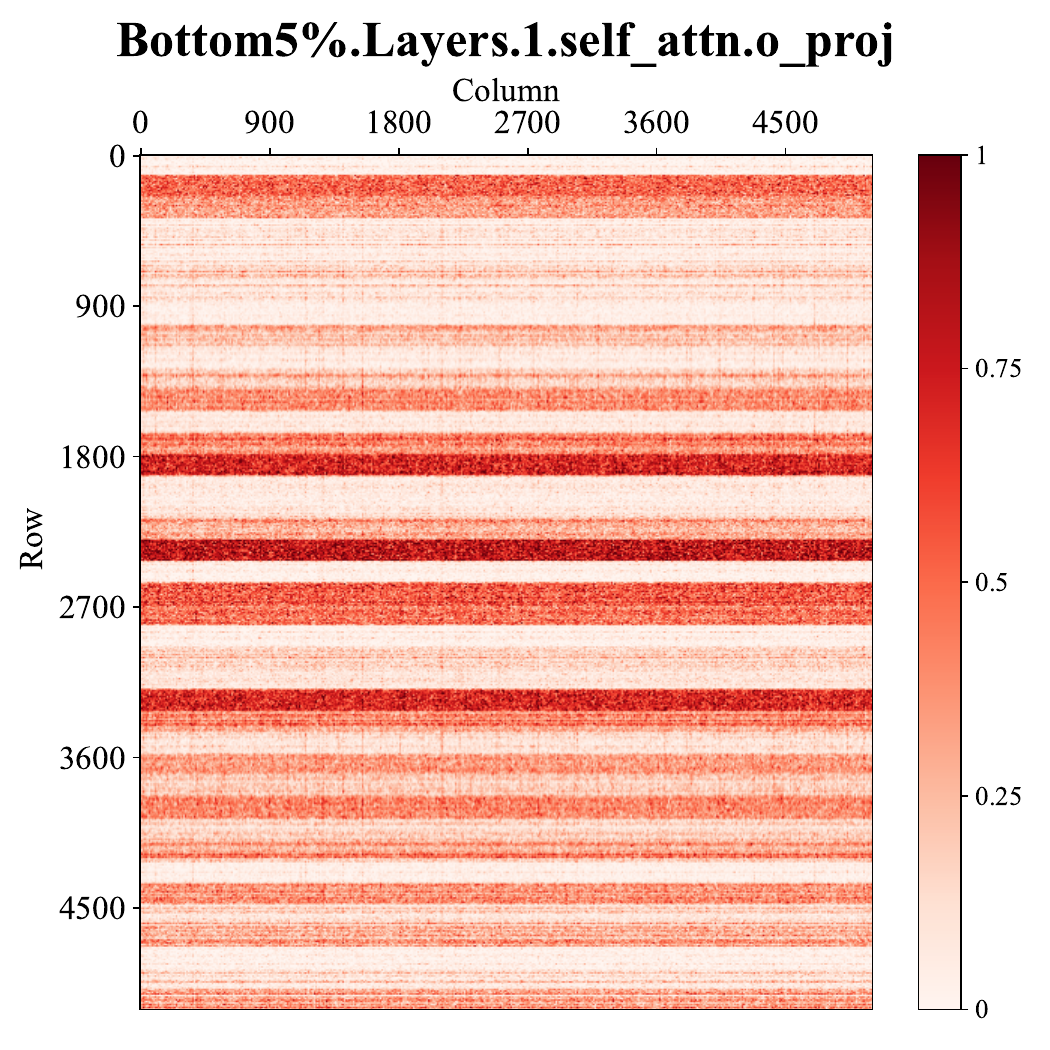}
            \includegraphics[width=3.6cm]{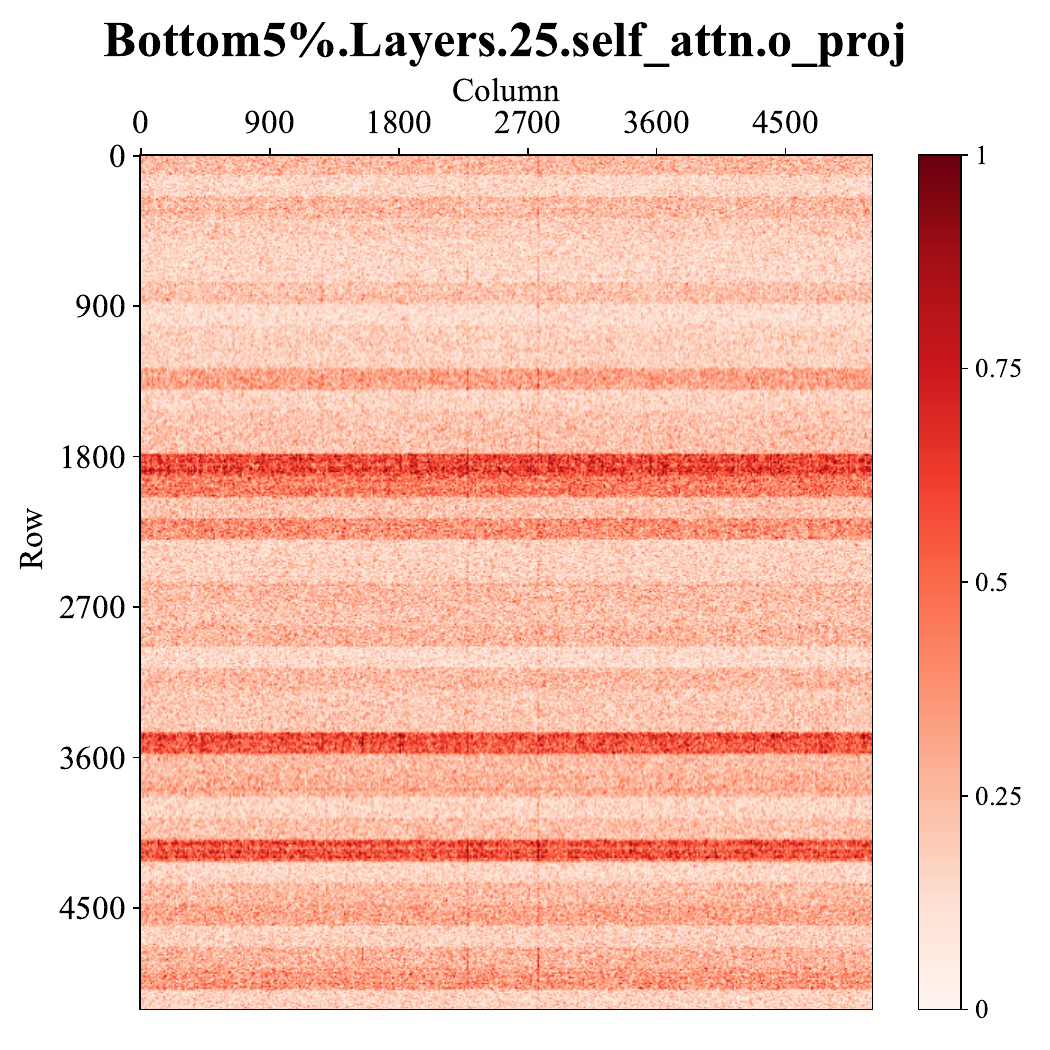}
	\end{minipage}
	}
	\subfigure{
		\centering
	\begin{minipage}[t]{0.23\textwidth}
		\includegraphics[width=3.6cm]{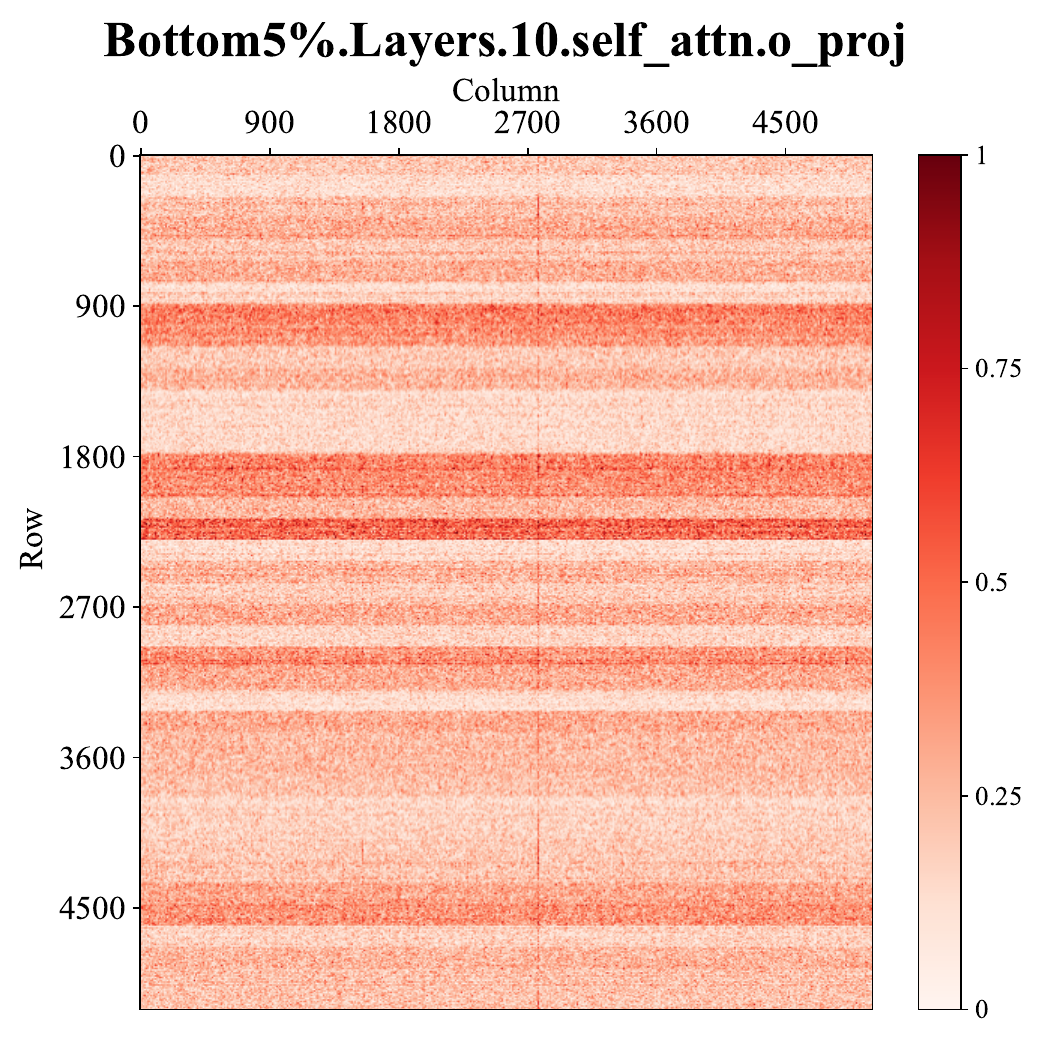}
		\includegraphics[width=3.6cm]{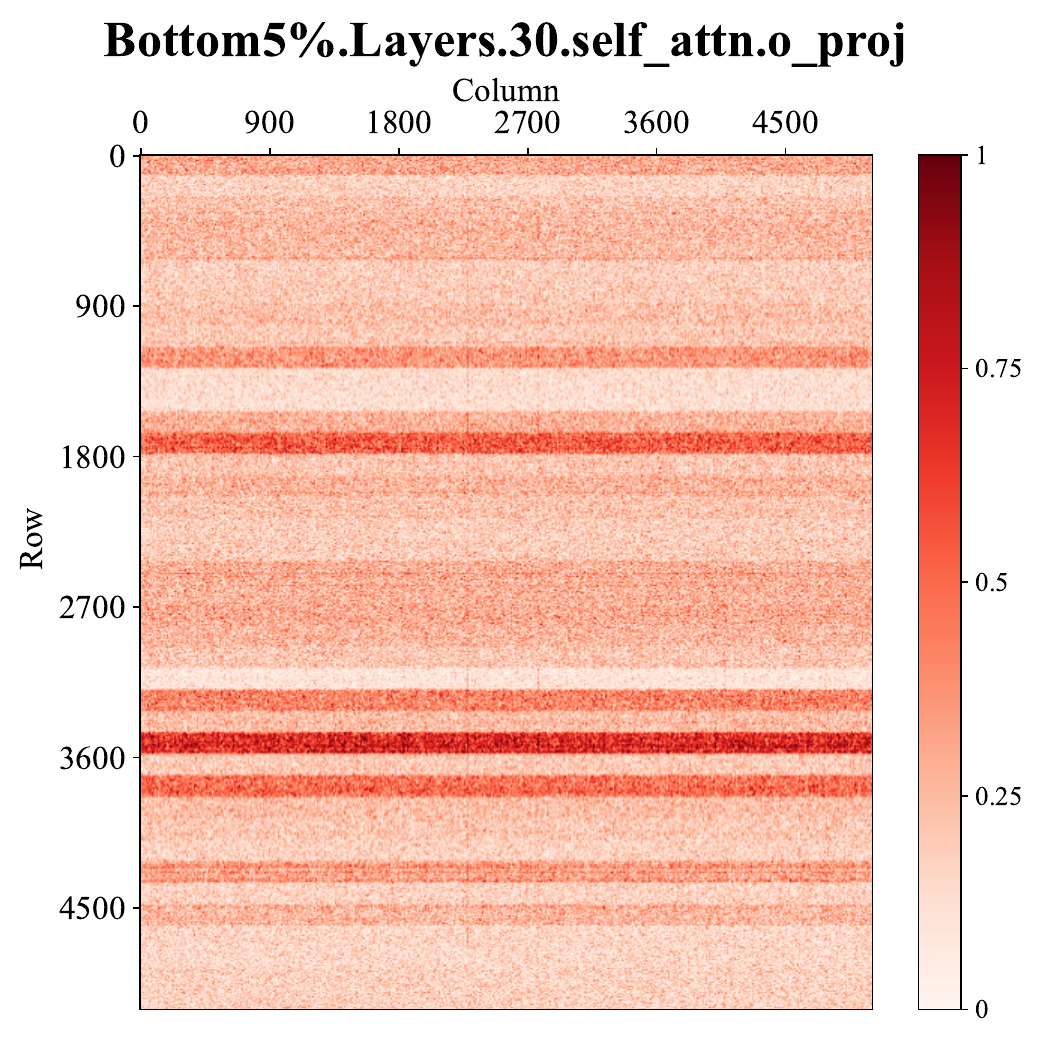}
	\end{minipage}
	}
 \subfigure{
		\centering
	\begin{minipage}[t]{0.23\textwidth}
		\includegraphics[width=3.6cm]{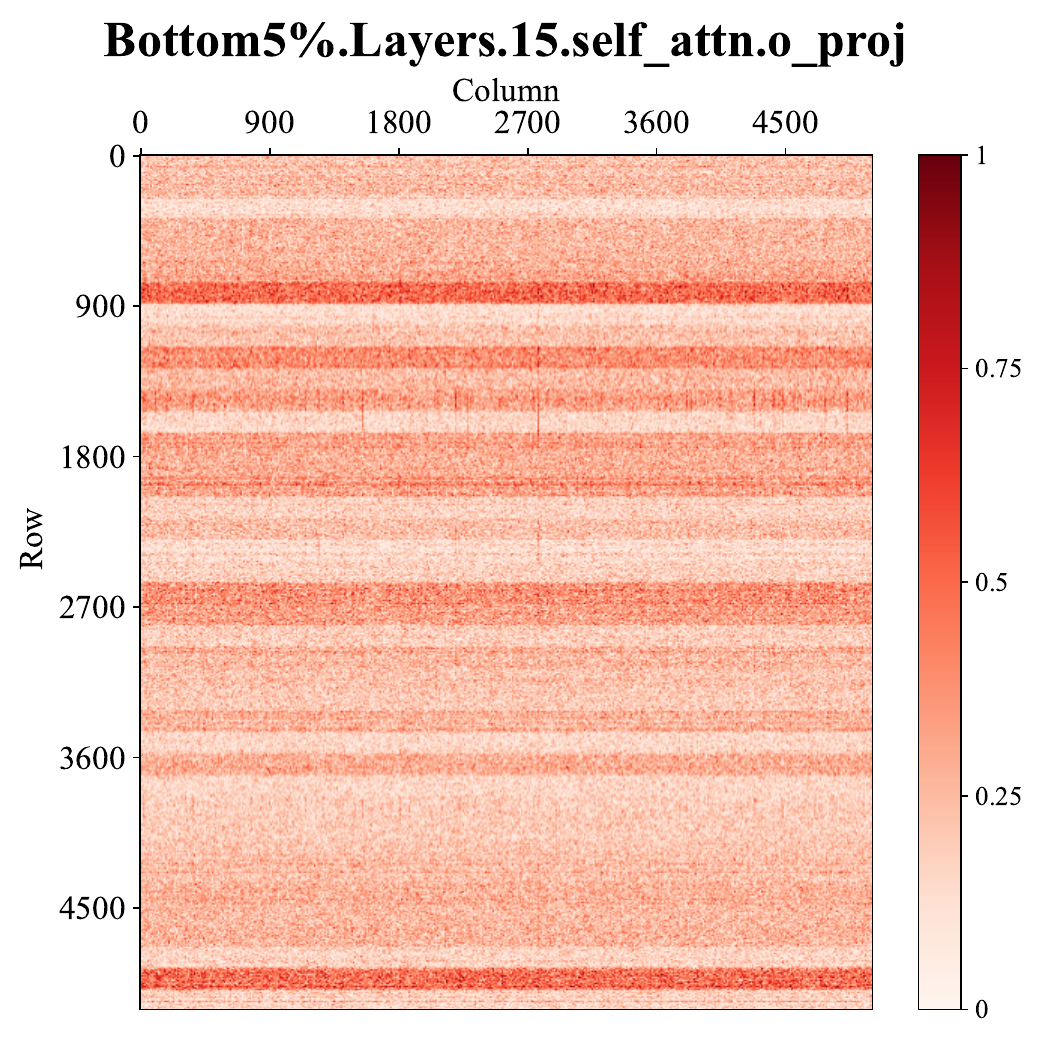}
		\includegraphics[width=3.6cm]{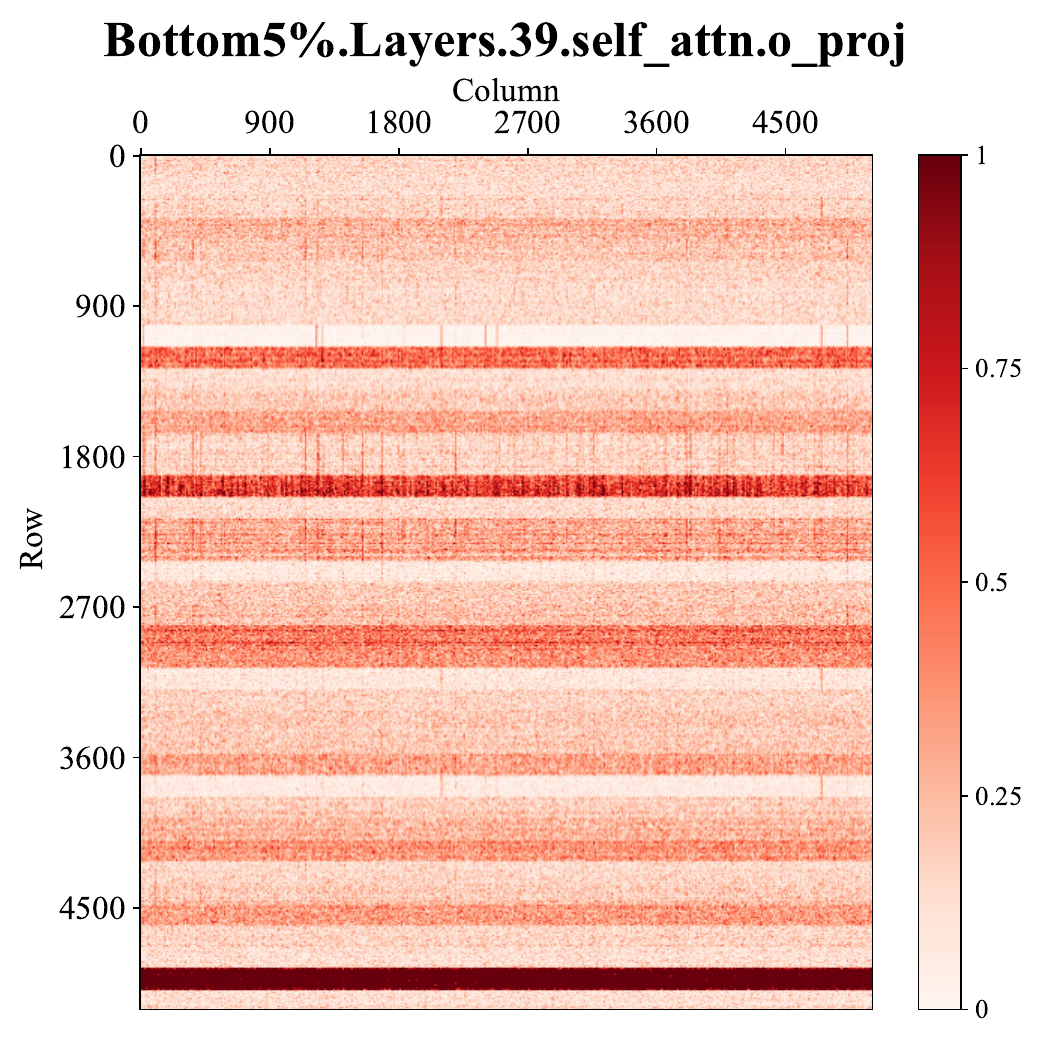}
	\end{minipage}
	}
 \caption{Visualization of Attn.o's `Bottom' region in LLaMA2-13b. The scale from 0 to 1 (after normalization) represent the proportion of parameters within a $3\times3$ vicinity that belong to the Bottom region.}
\label{fig:app_visualize_13b_o_bot}
\end{figure*}

\begin{figure*}[t]
	\centering
	\subfigure{
		\centering
	\begin{minipage}[t]{0.23\textwidth}
        \includegraphics[width=3.6cm]{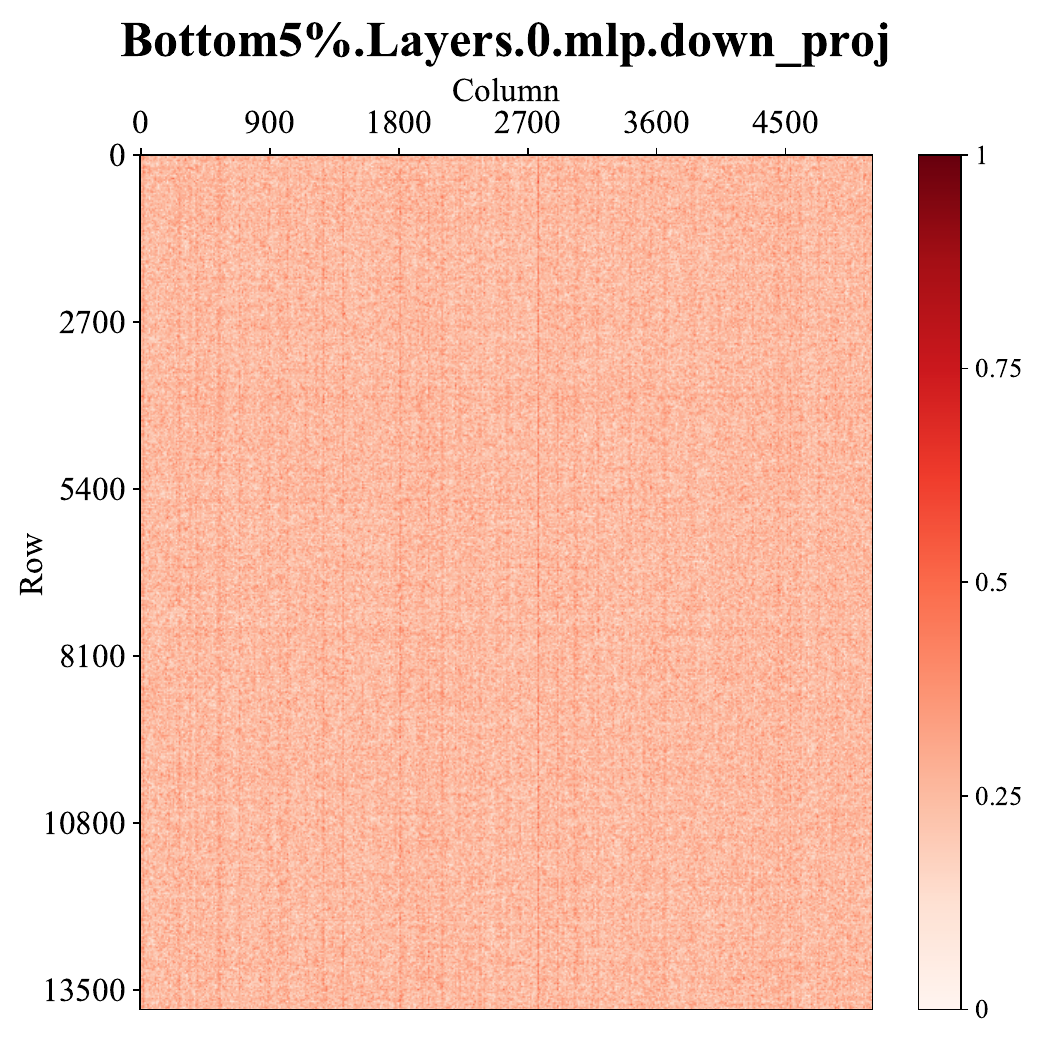}
        \includegraphics[width=3.6cm]{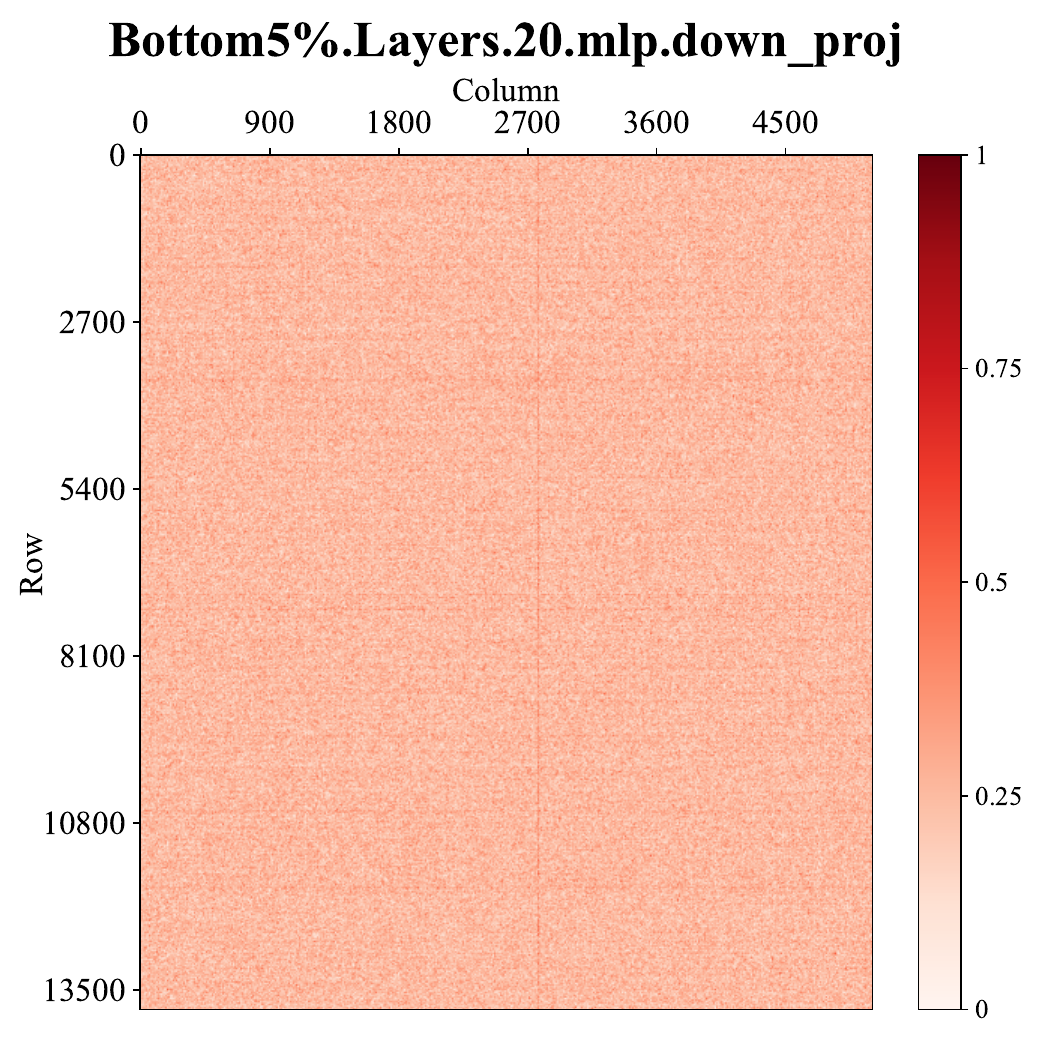}
	\end{minipage}
	}
	\subfigure{
		\centering
	\begin{minipage}[t]{0.23\textwidth}
	       \includegraphics[width=3.6cm]{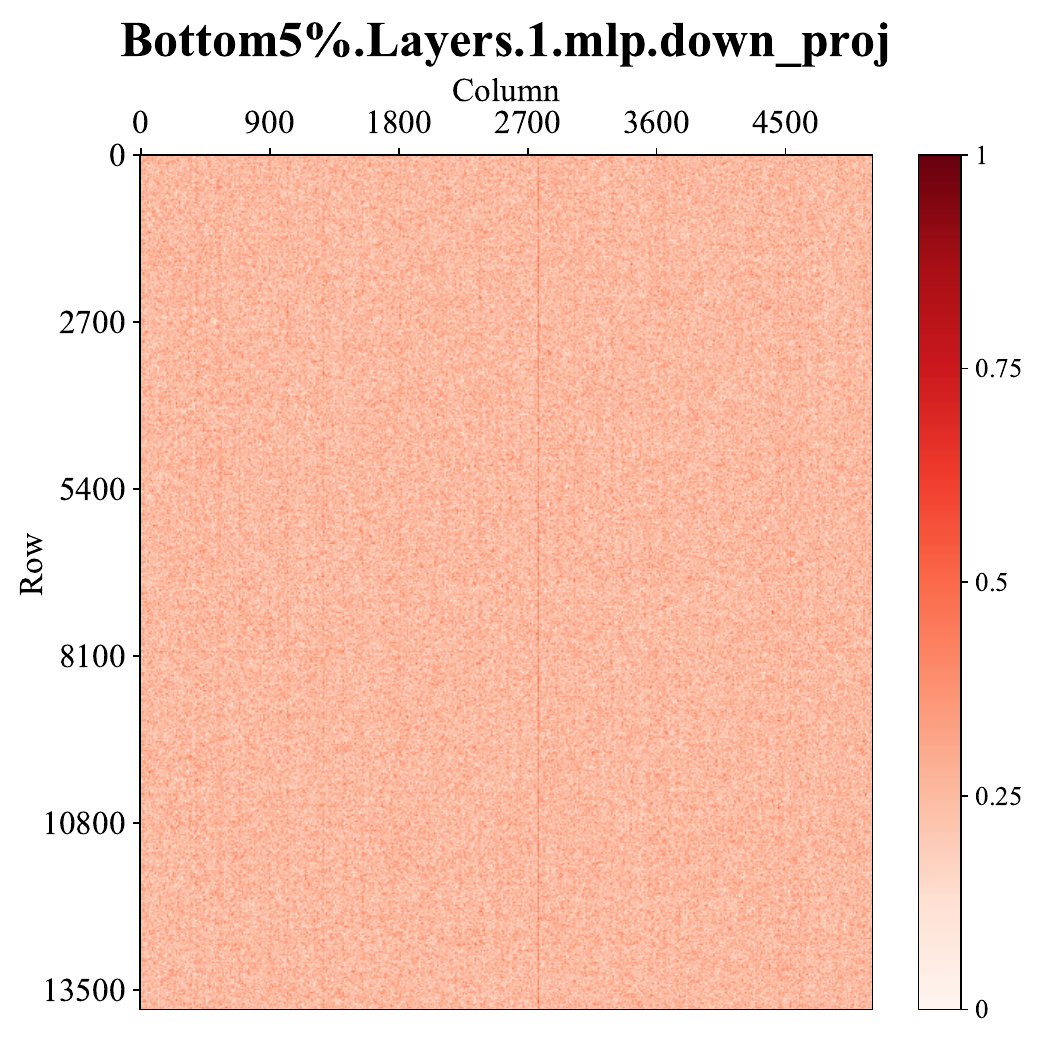}
            \includegraphics[width=3.6cm]{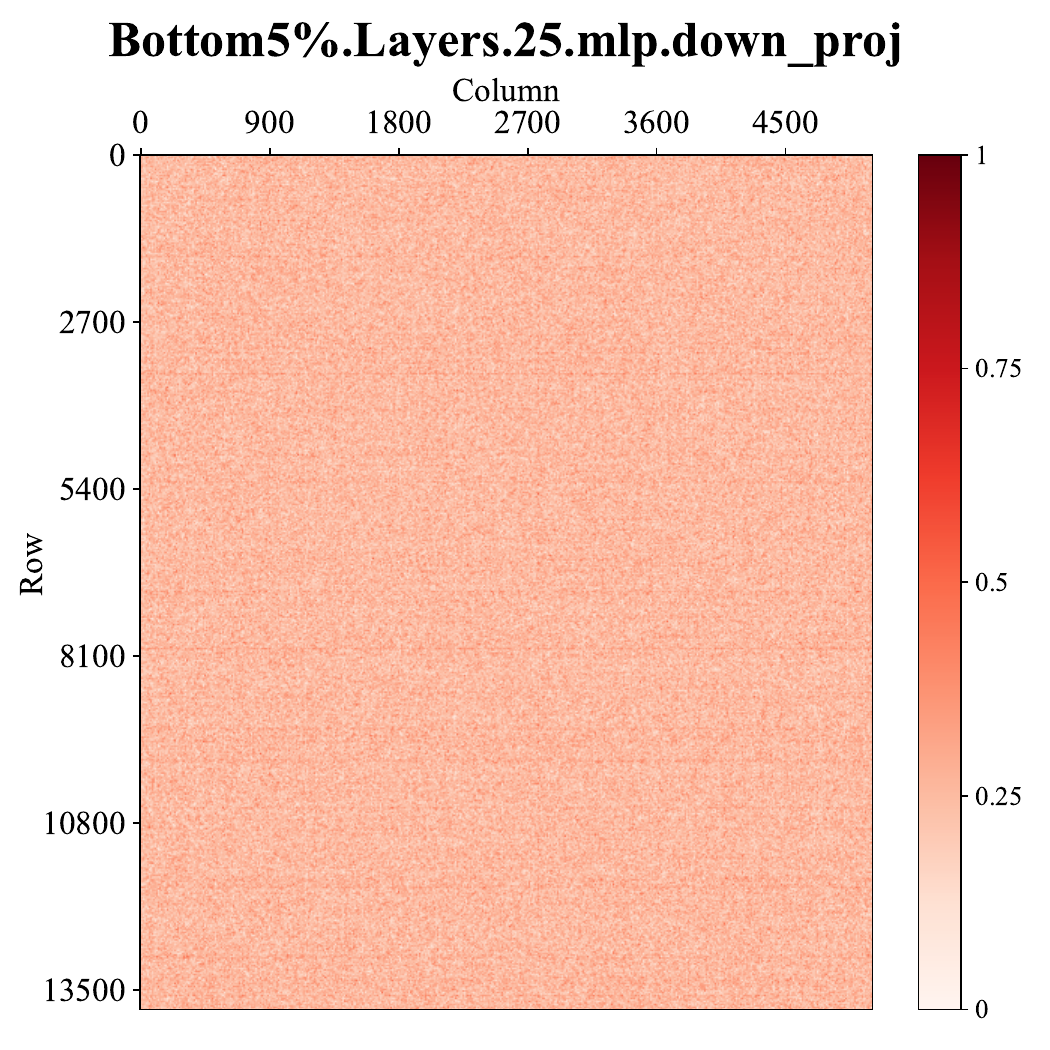}
	\end{minipage}
	}
	\subfigure{
		\centering
	\begin{minipage}[t]{0.23\textwidth}
		\includegraphics[width=3.6cm]{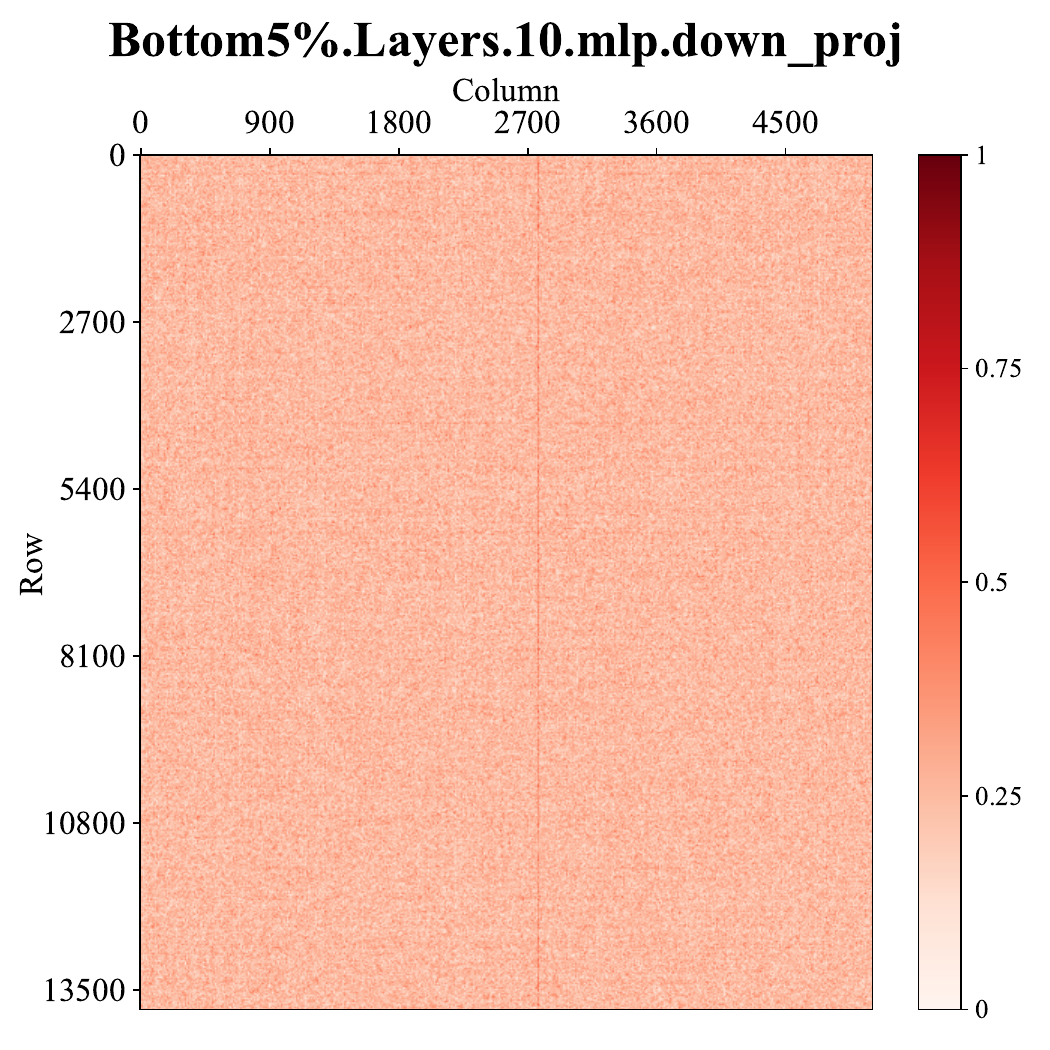}
		\includegraphics[width=3.6cm]{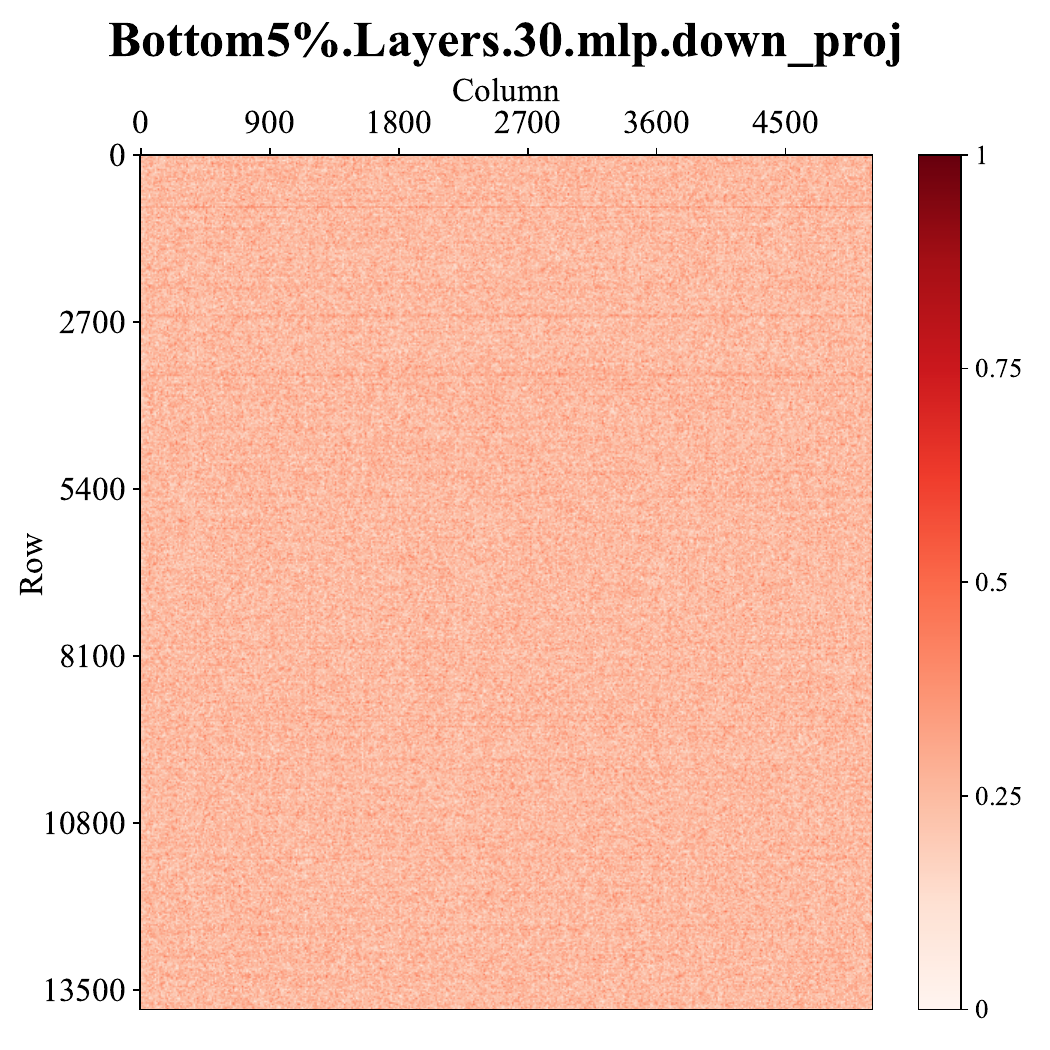}
	\end{minipage}
	}
 \subfigure{
		\centering
	\begin{minipage}[t]{0.23\textwidth}
		\includegraphics[width=3.6cm]{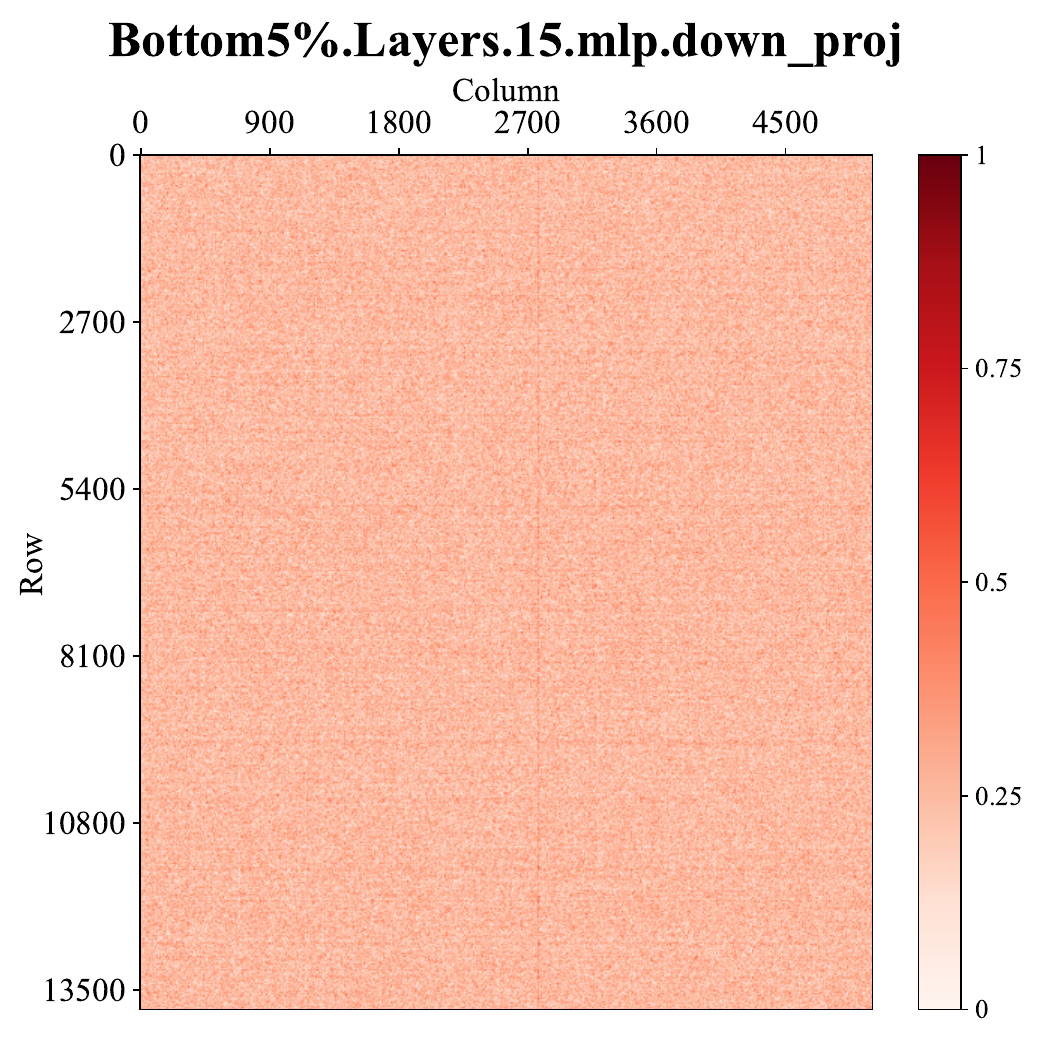}
		\includegraphics[width=3.6cm]{pic/13b-bottom/13b-bottom-mlp-down/39-mlp.down_proj.pdf}
	\end{minipage}
	}
 \caption{Visualization of FFN.down's `Bottom' region in LLaMA2-13b. The scale from 0 to 1 (after normalization) represent the proportion of parameters within a $3\times3$ vicinity that belong to the Bottom region.}
\label{fig:app_visualize_13b_down_bot}
\end{figure*}

\end{document}